\documentclass{article}

\usepackage{microtype}
\usepackage{graphicx}
\usepackage{subfigure}
\usepackage{booktabs} 

\usepackage{hyperref}

\usepackage[accepted]{mlsys2025}

\usepackage{tabularray}
\usepackage{subfigure}

\usepackage{amsmath}

\usepackage{amssymb}
\usepackage{enumitem}
\usepackage{adjustbox}
\usepackage{threeparttable}

\newenvironment{denseitemize}{
	\begin{itemize}[topsep=2pt, partopsep=0pt, leftmargin=1.5em]
		\setlength{\itemsep}{3pt}
		\setlength{\parskip}{0pt}
		\setlength{\parsep}{0pt}
	}{\end{itemize}}

\usepackage{url}
\usepackage[most]{tcolorbox}
\usepackage{multirow}
\newcounter{observation}
\newcounter{surveyobservation}

\usepackage{xcolor}
\usepackage{listings}

\definecolor{codebg}{RGB}{240, 240, 240} 
\definecolor{keyword}{RGB}{0, 0, 255}   
\definecolor{comment}{RGB}{0, 128, 0}   
\definecolor{string}{RGB}{255, 0, 0}    

\mlsystitlerunning{Rethinking KV cache Compression Techniques for LLMs}

\begin{document}

\twocolumn[
\mlsystitle{Rethinking Key-Value Cache Compression Techniques for Large Language Model Serving}



\mlsyssetsymbol{equal}{*}

\begin{mlsysauthorlist}
\mlsysauthor{Wei Gao}{equal,ntu,slab,shai}
\mlsysauthor{Xinyu Zhou}{equal,ntu}
\mlsysauthor{Peng Sun}{shai,st}
\mlsysauthor{Tianwei Zhang}{ntu}
\mlsysauthor{Yonggang Wen}{ntu}
\end{mlsysauthorlist}

\mlsysaffiliation{ntu}{Nanyang Technological University}
\mlsysaffiliation{slab}{S-Lab, Nanyang Technological University}
\mlsysaffiliation{shai}{Shanghai AI Laboratory}
\mlsysaffiliation{st}{SenseTime}

\mlsyscorrespondingauthor{Wei Gao}{gaow0007@e.ntu.edu.sg}

\mlsyskeywords{LLM, KV cache Compression}

\vskip 0.3in

\begin{abstract}
Key-Value cache (\texttt{KV} \texttt{cache}) compression has emerged as a promising technique to optimize Large Language Model (LLM) serving. It primarily decreases the memory consumption of \texttt{KV} \texttt{cache} to reduce the computation cost. Despite the development of many compression algorithms, their applications in production environments are still not prevalent. In this paper, we revisit mainstream \texttt{KV} \texttt{cache} compression solutions from a practical perspective. Our contributions are three-fold. First, we comprehensively review existing algorithmic designs and benchmark studies for \texttt{KV} \texttt{cache} compression and identify missing pieces in their performance measurement, which could hinder their adoption in practice. Second, we empirically evaluate representative \texttt{KV} \texttt{cache} compression methods to uncover two key issues that affect the computational efficiency: (1) while compressing \texttt{KV} \texttt{cache} can reduce memory consumption, current implementations (e.g., FlashAttention, PagedAttention) do not optimize for production-level LLM serving, resulting in suboptimal throughput performance; (2) compressing \texttt{KV} \texttt{cache} may lead to longer outputs, resulting in increased end-to-end latency. We further investigate the accuracy performance of individual samples rather than the overall performance, revealing the intrinsic limitations in \texttt{KV} \texttt{cache} compression when handling specific LLM tasks. Third, we provide tools to shed light on future \texttt{KV} \texttt{cache} compression studies and facilitate their practical deployment in production. They are open-sourced in \href{https://github.com/LLMkvsys/rethink-kv-compression}{https://github.com/LLMkvsys/rethink-kv-compression}. 
\end{abstract}
]




\printAffiliationsAndNotice{\mlsysEqualContribution} 

\section{Introduction}
\label{sec:introduction}
The groundbreaking success of Large Language Models (LLMs), e.g., ChatGPT~\cite{openai2023gpt4}, Gemini~\cite{gemini2023}, and Cluade~\cite{claudeai}, is reshaping the global technological landscape. The exponential growth in LLM service requests is creating an unprecedented demand for inference optimization algorithms, which can reduce exorbitant hardware costs and attain superior efficiency. In particular, many research efforts~\cite{sheng2023flexgen, H2O} pinpoint that the Key-Value cache (\texttt{KV} \texttt{cache}) acts as a major performance bottleneck in LLM serving. For instance, to serve a LLaMA3-70B model with FP16 format, a batch size of 512, and a prompt length of 2048, it requires 130GB of storage space for model weights and an additional 512 GB for the \texttt{KV} \texttt{cache}. This high resource requirement underscores the urgent need to mitigate the \texttt{KV} \texttt{cache} memory overhead.


There are two prominent strategies to compress \texttt{KV} \texttt{cache} and alleviate its exorbitant memory consumption. The first is quantization-based methods~\cite{KIVI, KVQuant, GEAR}, which convert the \texttt{KV} \texttt{cache} into low-precision representations to reduce the GPU memory usage, albeit with the potential of affecting the accuracy. The effectiveness of these methods depends on the design of representations, which need to strike a balance between memory reduction and maintaining model accuracy. The second is sparsity-based methods~\cite{StreamingLLM, Scissorhands, H2O, SnapKV}, which either move less critical portions of the \texttt{KV} \texttt{cache} from fast to slow memory or remove them directly. Their success hinges on accurately assessing the importance of the \texttt{KV} \texttt{cache} to ensure that the removal of selected entries does not compromise model performance.

Although researchers have demonstrated the effectiveness of \texttt{KV} \texttt{cache} compression algorithms, whether they can be deployed in the production environment is still unknown. To bridge this gap, this paper provides an in-depth investigation of existing compression algorithms. We make the following contributions. First, we provide a comprehensive survey of various \texttt{KV} \texttt{cache} compression algorithms (Table~\ref{tab:summary-kv-cache}), and summarize relevant benchmark studies (Table~\ref{tab:summary-kv-cache-benchmark}). Based on the literature review, we identify three key dimensions that have been overlooked in current research evaluation on LLM \texttt{KV} \texttt{cache} compression but are crucial for practical deployment. Beyond the model accuracy and GPU memory reduction, assessing the \textbf{throughput}, \textbf{length distribution}, and \textbf{negative samples}\footnote{This refers to the samples that are benign in the original LLMs, but turn to malign when \texttt{KV} \texttt{cache} compression is applied.} can heavily affect the adoption of these compression algorithms in real-world production environments.

Second, we conduct a full-around experimental analysis on representative \texttt{KV} \texttt{cache} compression algorithms, focusing on our identified three dimensions. This brings several novel and interesting observations. (1) \texttt{KV} \texttt{cache} compression algorithms can bring throughput improvement in the decoding stage. However, our evaluation presents poor performance under certain batch sizes and prompt lengths. Besides, when we integrate \texttt{KV} \texttt{cache} compression algorithms with \texttt{FlashAttention}~\cite{dao2022flashattention,dao2023flashattention2} and \texttt{PagedAttention}~\cite{kwon2023efficient}, the performance gains from \texttt{KV} \texttt{cache} compression diminish further. (2) The evaluation practice for the computational efficiency of \texttt{KV} \texttt{cache} compression is length controlled. By measuring the length distribution shift induced by \texttt{KV} \texttt{cache} compression, we observe that the lengthier responses produced by these methods can outweigh the throughput speedup benefits, ultimately leading to extended end-to-end latency. (3) We analyze the negative samples in the performance evaluation of \texttt{KV} \texttt{cache} compression methods. We observe their existence across many compression methods and uncover the fragility of \texttt{KV} \texttt{cache} compression towards specific task types. 

Third, driven by the limitations of existing \texttt{KV} \texttt{cache}  compression algorithms for practical deployment, we provide a set of tools to facilitate their adoption. This includes (1) a throughput analysis tool to decide under which ranges of batch sizes and prompt lengths \texttt{KV} \texttt{cache} compression can present advantageous performance; (2) a length predictor to inform the serving system of requests that could produce longer responses when applying \texttt{KV} \texttt{cache} compression algorithms; (3) a benchmark dataset consisting of our discovered failure cases to help researchers more fairly and accurately assess their compression solutions.

\section{Background and Motivation}
\label{sec:preliminary}
\subsection{Large Language Model (LLM)}
\label{subsec:background-llm}
A conventional LLM inference process is composed of the following two stages: 

\noindent\textbf{Prefill Stage.}
The prompts are employed to produce \texttt{KV} \texttt{cache} for each transformer layer. Formally, given the input tensor $X \in \mathcal{R}^{b \times l \times d}$, of batch size $b$, input prompt length $l$, and hidden dimension $d$, The key and value tensors are calculated as follows:
\begin{equation}
\begin{aligned}
    X_K &= X \cdot W_K, \\
    X_V &= X \cdot W_V,
\end{aligned}
\label{eq: prefill}
\end{equation}
where $W_K, W_V \in \mathcal{R}^{d \times d}$ are the weights of the key and value layers, respectively. $X_K$ and $X_V$ are typically cached in the memory, forming \texttt{KV} \texttt{cache} to avoid the re-computation cost in the decoding stage.

\noindent\textbf{Decoding Stage.}
The model leverages and updates the \texttt{KV} \texttt{cache} dynamically to decode and generate tokens sequentially.
We denote the current input token embedding as $x \in \mathcal{R}^{b \times 1 \times d}$. The key, value, and query layer outputs are calculated as $x_K = x \cdot W_K, x_V = x \cdot W_V, x_Q = x \cdot W_Q$. Subsequently, we can update the \texttt{KV} \texttt{cache} as follows:

\begin{equation}
\begin{aligned}
    X_K & \leftarrow [X_K, x_K], \\
    X_V & \leftarrow [X_V, x_V],
\end{aligned}
\label{eq: kv update}
\end{equation}
where $[ \cdot ]$ denotes the concatenation operation along with the token dimension.


\subsection{LLM Serving Acceleration}
LLM practitioners commonly adopt two methods to accelerate the LLM serving in production deployment.

\noindent\textbf{FlashAttention}~\cite{dao2022flashattention}. 
This is an IO-aware technique designed to reduce the memory overhead in the attention operation. It employs tiling and re-computation strategies alongside an online softmax operation, facilitating tile-based matrix multiplication. The tile size is carefully determined to fit submatrices within the on-chip memory, thereby minimizing the need for data loading to the memory. 

\noindent\textbf{PagedAttention}.
As introduced in vLLM~\cite{vllm}, this technique emulates the virtual memory and paging mechanisms in operating systems for managing the \texttt{KV} \texttt{cache}. It dynamically allocates small GPU memory blocks for the \texttt{KV} \texttt{cache} on demand, rather than pre-allocating memory to the maximum length. This effectively reduces the GPU memory overhead and fragmentation. Usually, PagedAttention is implemented with FlashAttention together to yield the attention output.

\subsection{\texttt{KV} \texttt{Cache} Compression}
\label{subsec:compression}
Past works have introduced different techniques to compress the \texttt{KV} \texttt{cache} for LLM serving optimization. They can be roughly classified into the following two categories. 

\noindent\textbf{Quantization.} These approaches reduce the size of \texttt{KV} \texttt{cache} by using low bits (e.g., INT8) to represent the original full precision (e.g., FP16) without degrading the model accuracy. The quantization and de-quantization processes can be illustrated as follows: 
\begin{equation}
\begin{aligned}
    \text{quantize:} & \quad X_{\text{quant}} = \lfloor \frac{X - \ell}{\Delta} \rceil, \quad \Delta = \frac{u - \ell}{2^b - 1}, \\
    \text{de-quantize:} & \quad \hat{X} = X_{\text{quant}} \cdot \Delta + \ell, 
\end{aligned}
\label{eq:quantization-operation}
\end{equation}
where $X$ is the original \texttt{KV} \texttt{cache}, $X_{\text{quant}}$ is the quantized result from $X$, and $\hat{X}$ is the de-quantized result from $X_{\text{quant}}$. These quantization-based methods aim to minimize the difference between $X$ and $\hat{X}$. As the quantized \texttt{KV} \texttt{cache} does not participate in the prefill stage, quantization typically happens in the decoding stage to reduce the memory consumption of token generation.


\noindent\textbf{Sparsity.} These approaches are inspired by the sparsity of attention scores~\cite{StreamingLLM, Scissorhands, H2O, SnapKV, PyramidInfer, PyramidKV}. They dramatically reduce the memory footprint of the \texttt{KV} \texttt{cache} by evicting the KV pairs of less important tokens while retaining the ones of more important tokens to preserve the model's accuracy. The key to these solutions is to determine when and how many tokens should be evicted for effective compression.

\subsection{Concerns in LLM Production Deployment}
\label{subsec:llm-production-practice}
Despite the memory reduction benefits, the main concerns that prevent LLM practitioners from deploying \texttt{KV} \texttt{cache} compression methods in production environments are accuracy and computational efficiency. 

For accuracy, many compression studies emphasize minor overall accuracy loss while concealing the specific accuracy performance defects of \texttt{KV} \texttt{cache} compression methods. Furthermore, many techniques possess abundant configurations to balance the accuracy performance and memory compression ratio, increasing the burden of tuning the complex configuration parameters.

For computational efficiency, LLM practitioners focus on the key metrics of Time-To-First-Token (TTFT) and Time-Between-Output-Token (TBOT). However, they are typically measured using the naive transformers library (TRL)~\cite{transformers-library}, which falls short of the performance expected in production environments. Besides, the speedup benefits from \texttt{KV} \texttt{cache} compression are not always guaranteed. In particular, during the decoding stage, compressing the \texttt{KV} \texttt{cache} primarily influences the attention operation. The time required for the attention operation to generate one token is comprised of two steps: (1) loading data into on-chip memory and (2) performing the computation. The reduction of memory consumption brought by \texttt{KV} \texttt{cache} compression can lead to decreased time for the first step. However, for quantization-based methods, the operation described in Eqn.~\ref{eq:quantization-operation} incurs additional overhead for the second step. For sparsity-based methods, although the reduced \texttt{KV} \texttt{cache} size can lower the time cost spent on the second step, an extra computation step is required for the \texttt{KV} \texttt{cache} eviction, with additional time overhead.

\section{A Comprehensive Survey}
\label{sec:surveying}

\begin{table*}[!htp]\centering
\caption{Summary of existing \texttt{KV} \texttt{cache} compression algorithms. }\label{tab:summary-kv-cache}
\begin{adjustbox}{width=0.95\textwidth} 
\begin{tabular}{clcrrrrrrr}\toprule
\textbf{Date (YY.MM)} &\textbf{Algorithm} &\textbf{Quant / Sparse} &\textbf{Algorithm Features} &\textbf{Model} &\textbf{Heavy Eval} & \textbf{Mem}	& \textbf{Prf Thr}	& \textbf{Dec Thr}	& \textbf{Frw} \\\midrule
24.02 &KVQuant~\cite{KVQuant} &Q &Per-channel key quantization &L &65B / 1 / 32k &8.0 $\times$ &- &- &T \\
24.02 &WKVQuant~\cite{WKVQuant} &Q &Loss design for quant parameter optimization &L &13B / 16 / 18k &4.0 $\times$ &- &- &T \\
24.02 &KIVI~\cite{KIVI} &Q &Per-channel key quantization & L, M, F &13B / ~380 / 18k &2.6 $\times$ &2.3 $\times$ &3.4 $\times$ &T \\
24.02 &MiKV~\cite{MiKV} &Q &Mixed-precision quantization &L, M &70B / 8 / 4k &5.0 $\times$ &- &- &T \\
24.03 &IntactKV~\cite{IntactKV} &Q &Keep full-precision caches for outlier tokens & L, M &70B / 1 / - &4.0 $\times$ &- &- &T \\
24.03 &QAQ~\cite{QAQ} &Q &Quality-adaptive quantization &L &13B / 1 / - &10.0 $\times$ &- &- &T \\
24.03 &GEAR~\cite{GEAR} &Q &Approximate the quant error with low-rank matrix &L, M &13B / 18 / 7k &3.8 $\times$ &- &5.0 $\times$ &T \\
24.03 &QuaRot~\cite{QuaRot} &Q &Eliminate KV outliers with Hardmard matrix &L &70B / 64 / 2k &3.7 $\times$ &2.1 $\times$ & &T \\
24.05 &SKVQ~\cite{SKVQ} &Q &Clipped dynamic quant with channel reorder & L, M &13B / 128 / 200k &7.9 $\times$ & &7.0 $\times$ &T \\
24.05 &ZipCache~\cite{ZipCache} &Q &Channel-separable tokenwise quantization &L, M &13B / 8 / 4k &4.9 $\times$ &1.6 $\times$ &2.3 $\times$ &T \\
24.07 &QJL~\cite{QJL} &Q &Elimiate quant constants storage overheads with JL transform &L &8B / 1 / 18k &5.2 $\times$ &- &- &T \\
24.07 &Palu~\cite{Palu} &Q &KV cache compression with low-rank projection &L, M &13B / 1 / 64k &11.4 $\times$ &- &1.6 $\times$ &T \\
24.08 &ZDC~\cite{ZDC} &Q & Eliminate compression overhead & O, L &175B / 1 / 20k &10.0 $\times$ &- &2.8 $\times$ &T / D / V \\
\midrule 
\midrule

23.08 &Scissorhands~\cite{Scissorhands} &S &Window-based eviction with a counter-based token score &O &175B / 128 / 2k &5.0 $\times$ &- &- &T \\
23.12 &StreamingLLM~\cite{StreamingLLM} &S &Retain KV cache of initial tokens &L, F, M &70B / 1 / 18k &5.0 $\times$ &- &- &T \\
23.12 &H2O~\cite{H2O} &S &Accumulate attention scores as token score &L, O, G &66B / 64 / 7k &5.0 $\times$ &- &29.0 $\times$ &T / D / F \\
24.01 &FastGen~\cite{FastGen} &S &Head-adaptive eviction policy &L &65B / 16 / 4k &1.6 $\times$ &- &1.2 $\times$ &T / D / F \\
24.02 &LESS~\cite{LESS} &S &Merge to-be-evicted caches into low-rank matrix &L, F &13B / 64 / 5k &50.0 $\times$ &- &1.7 $\times$ &T \\
24.02 &ROCO~\cite{ROCO} &S &Standard deviation of attention score as token score &L, W &7B / 1 / - &3.3 $\times$ &- &- &T \\
24.04 &Keyformer~\cite{Keyformer} &S &Add gumbel-based regularization in token score &G &7B / 2 / 4k &2.0 $\times$ &- &2.4 $\times$ &T \\
24.04 &SqueezeAttention~\cite{SqueezeAttention} &S &Reallocate KV cache budget across layers &L, M, F, G, O &70B / 224 / 18k &3.3 $\times$ &- &2.2 $\times$ &T \\
24.04 &SnapKV~\cite{SnapKV} &S &Select clustered important KV cache across heads &L, M &35B / 8 / 26k &8.2 $\times$ &- &3.6 $\times$ &T \\
24.04 &CORM~\cite{CORM} &S &Budget-unrestricted KV cache eviction &L &7B / 1 / 18k &3.3 $\times$ &- &- &T \\
24.05 &CaM~\cite{CaM} &S &Merge to-be-evicted caches into recent KV cache &L, O, G &13B / 1 / - &3.3 $\times$ &- &- &T \\
24.05 &PyramidInfer~\cite{PyramidInfer} &S &Drop KV cache during KV cache computation process &L &70B / 88 / 2k &2.1 $\times$ &- &2.2 $\times$ &T / D \\
24.05 &MiniCache~\cite{MiniCache} &S &Multiple layers sharing the same retained KV cache &L, M &70B / 300 / 18k &1.7 $\times$ &- &5.0 $\times$ &T \\
24.05 &InfLLM~\cite{InfLLM} &S &Store evicted tokens as context memory for furthur lookups &L, M &8B / 1 / 100k &2.9 $\times$ &- &1.5 $\times$ &T \\
24.05 &Q-Hitter~\cite{Q-Hitter} &Q + S &Keep quantization-friendly and important tokens &L, O &30B / 1 / 4M &20 $\times$ &- &33.0 $\times$ &T \\
24.06 &Quest~\cite{Quest} &S &Query-aware cache eviction policy &L &7B / 1 / 64k &8.0 $\times$ &- &2.2 $\times$ &F \\
24.06 &PyramidKV~\cite{PyramidKV} &S &Adjust KV cache budget across layers &L, M &8B / 1 / 18k &8.3 $\times$ &- &- &T \\
24.06 &SampleAttention~\cite{SampleAttention} &S &Adaptive structured sparse attention &C, I &6B / 1 / 200k & 12.5$\times$ &2.2 $\times$ &- &T \\
24.07 &TOVA~\cite{TOVA} &S &Enable recent KV cache evictable &L &7B / 139 / 70k & - &- &4.8 $\times$ &T \\
24.07 &LazyLLM~\cite{LazyLLM} &S &Revive previously evicted KV cache &L &7B / 1 / 18k & - &2.3 $\times$ &- &T \\
24.07 &Ada-KV~\cite{Ada-KV} &S &Allocate KV cache budget across different heads &L, M &7B / 1 / 18k & 3.3 $\times$ &- &- &T \\
24.07 &RazorAttention~\cite{RazorAttention} &S &Disable KV cache eviction for retrieval heads &L, Q, B &72B / 1 / 18k &3.3 $\times$ &- &- &T \\
24.07 &ThinK~\cite{ThinK} &S &Evict KV cache in channel dimension &L, M &8B / 1 / 18k &1.25 $\times$ &- &- &T \\
24.08 &NACL~\cite{NACL} &S &General KV cache eviction framework &L &7B / 4 / 32k &5.0 $\times$ &- &- &T \\
24.08 &DoubleSparse~\cite{DoubleSparse} &S &Prefetch tokens with token and channel sparsity & L, M &70B / 32 / 256k &16 $\times$ &- &16.3 $\times$ &T \\
24.09 &GemFilter~\cite{GemFilter} &S &Use early layes of LLM to filter and compress tokens &L, M &12B / 1 / 120k &1.43 $\times$ &- &2.4 $\times$ &T \\
24.09 &RetrievalAttention~\cite{RetrievalAttention} &S &Leverage vector search for dynamic sparse attention &L &8B / 1 / 1M &- &- &4.9 $\times$ &T \\
24.10 &DuoAttention~\cite{DuoAttention} &S &Identify streaming heads to accelerate attention &L, M &8B / 1 / 3.3M &2.55 $\times$ &1.73 $\times$ &2.18 $\times$ &F \\

\bottomrule
\end{tabular}
\end{adjustbox}
\vspace{-2pt}
    \begin{flushleft}
    \begin{tablenotes}[para,flushleft]
    \scriptsize
        Notations: In column \textbf{Model}, L, M, F, O, C, and I represent Llama, Mistral, Falcon, OPT, ChatGLM, and the InternLM LLM family. In column \textbf{Heavy Eval}, we list the heaviest evaluation setting in a format of model size/ batch size/ prompt length. We list the maximum memory reduction, prefill throughput speedup, and decoding throughput speedup in column \textbf{Mem}, \textbf{Prf Thr}, and \textbf{Dec Thr}. For evaluation frameworks shown in column \textbf{Frw}, T, D, F, and V represent the transformer library~\cite{transformers-library}, DeepSpeed~\cite{deepspeed-inference}, FlashInfer~\cite{flashinfer2024}, and vLLM~\cite{vllm}.
        \end{tablenotes}
    \end{flushleft}  
\vspace{-20pt}
\end{table*}

\subsection{\textbf{KV Cache} Compression Algorithms}
\label{subsec:survey-methods}
Table~\ref{tab:summary-kv-cache} summarizes existing quantization- and sparsity-based \texttt{KV} \texttt{cache} compression algorithms. We focus on the design components of these algorithms that could adversely affect computational efficiency. 


\subsubsection{Quantization}

We first review some relevant \texttt{KV} \texttt{cache} quantization algorithms, focusing on the quantization granularity and quantization error. First, a few studies observe that the sensitivity to quantization operations varies with the granularity of \texttt{KV} \texttt{cache}. ZipCache~\cite{ZipCache} and WKVQuant~\cite{WKVQuant} adopt channel-separable token-wise quantization for \texttt{KV} \texttt{cache}. KVQuant~\cite{KVQuant}, and KIVI~\cite{KIVI} utilize per-channel quantization for the key tensor and per-token quantization for the value tensor. QJL~\cite{QJL} introduces the quantized JL transform to key tensor and per-token quantization for value tensor to reduce the memory consumption of the quantized tensor. MiKV~\cite{MiKV}, QAQ~\cite{QAQ}, and SKVQ~\cite{SKVQ} allow varying bits to represent \texttt{KV} \texttt{cache} to attain accuracy and memory reduction balance. Coupled Quantization~\cite{CoupledQuantization} integrates multiple channels and jointly quantizes them. Additionally, it leverages Fisher information to prioritize important tokens when quantizing. From an accuracy standpoint, researchers are advancing dedicated quantization operations toward finer \texttt{KV} \texttt{cache} granularity. \textbf{While finer quantization granularity may preserve the accuracy performance, it introduces irregular computational patterns, thereby limiting the effective utilization of GPU resources.} Consequently, these approaches may not lead to improved computational performance. 

Second, some research efforts aim to rectify‌ the quantization error to sustain the LLM response quality. For example, Gear~\cite{GEAR} approximates the quantization error with a low-rank matrix. Quantization outliers, which manifest as quantization errors with extreme values, can greatly impair model performance. To counter this, IntactKV~\cite{IntactKV} retains the full precision of the outlier to maintain the model accuracy, while Gear employs a sparse matrix to mitigate the error caused by the outliers. Additionally, QuaRot~\cite{QuaRot} innovatively transforms each weight matrix with Hadamard orthogonal matrices to eliminate the quantization outlier without changing the LLM output. Addressing the quantization error remains a promising area to improve the accuracy of quantization-based methods. However, error correction necessitates another step to compute the quantization error. \textbf{Mitigating the quantization error not only compromises the memory efficiency but also incurs additional computational costs during inference, which can offset the computational benefits from memory reduction.}

Note that most quantization-based algorithms keep a window of recent historical \texttt{KV} \texttt{cache} as full precision for accuracy considerations, which could burden the compatibility with PagedAttention. Specifically, PagedAttention maintains a number of tensors with a fixed page size and tensor type. The window-based quantization demands two types of paged tensors to manage full-precision and quantized \texttt{KV} \texttt{cache}, respectively. This introduces unstructured computation patterns when computing the attention output. \textbf{The window-based design choice in quantization-based methods increases the deployment complexity, potentially negating the computational efficiency gains. }

\subsubsection{Sparsity} 
We further review sparsity-based \texttt{KV} \texttt{cache} compression algorithms from two perspectives. First, we investigate the granularity of \texttt{KV} \texttt{cache} compression, encompassing token-level, layer-level, head-level, and channel-level. Early works including Scissorhands~\cite{Scissorhands}, StreamingLLM~\cite{StreamingLLM}, H2O~\cite{H2O}, ROCO~\cite{ROCO}, Keyformer~\cite{Keyformer} propose discarding the \texttt{KV} \texttt{cache} of unimportant tokens. Layer-level \texttt{KV} \texttt{cache} eviction methods, e.g., SqueezeAttention~\cite{SqueezeAttention}, PyramidKV~\cite{PyramidKV}, and MiniCache~\cite{MiniCache}, enable the selective removal of \texttt{KV} \texttt{cache} entries corresponding to different token positions across layers. FastGen~\cite{FastGen}, SnapKV~\cite{SnapKV}, CORM~\cite{CORM}, Ada-KV~\cite{Ada-KV}, RazorAttention~\cite{RazorAttention}, and NACL~\cite{NACL} evict varying token positions across different heads. Notably, ThinK~\cite{ThinK} deviates from these approaches by targeting partial channel dimensions, thereby achieving a consistent reduction in \texttt{KV} \texttt{cache} size, irrespective of the sequence length. PQCache~\cite{PQCache} utilizes Product Quantization (PQ) to manage the KV cache. It employs Maximum Inner-Product Search (MIPS) to identify relevant tokens for attention computations during the decoding stage. \textbf{Similar to quantization-based techniques, pursuing finer granularity in sparsity-based methods can yield improved accuracy, at the cost of high GPU utilization owing to the resulting irregular computational patterns.}

Second, we summarize the eviction policies, focusing on the importance metric, eviction scope, and budget allocation. (1) The \textbf{importance metric} is to determine the relative order by which caches are evicted. Many attention score variants~\cite{Scissorhands,FastGen,ROCO,Keyformer,Quest} are used as the importance metric. The accumulated attention score is particularly popular due to its robust accuracy performance over diverse LLM tasks. (2) The \textbf{eviction scope} defines the tokens eligible for eviction. A prevalent approach is to employ a local window to reduce the eviction overhead. Additionally, constraints are often considered to ensure that specific tokens~\cite{StreamingLLM,TOVA}, heads~\cite{RazorAttention}, or layers~\cite{Quest} are excluded from the eviction scope, safeguarding model quality. (3) The \textbf{budget allocation} dictates how the available GPU memory budget is allocated. A straightforward solution is to set a fixed memory budget; however, not all layers and heads are equally important. Therefore, some methods dynamically allocate the memory budget across layers~\cite{SqueezeAttention,PyramidInfer,PyramidKV} and heads~\cite{Ada-KV}. Moreover, CORM~\cite{CORM} introduces a budget-unrestricted \texttt{KV} \texttt{cache} eviction policy. The eviction policy of sparsity-based methods is more flexible and complex than the quantization policy as described in Eqn.~\ref{eq:quantization-operation}.

The pursuit of fine-grained token eviction and intricate eviction policies intensifies the challenges of gaining computation efficiency from sparsity-based methods. These concerns remain consistent across various LLM serving frameworks. Moreover, our analysis reveals that the algorithmic designs of sparsity-based methods are not compatible with FlashAttention and PagedAttention. FlashAttention gets rid of the multi-pass attention operation and exploits softmax and tiling to attain a one-pass operation, thus reducing the number of passes to load data from global memory to on-chip memory. However, the importance metric depends upon the attention scores, which are not saved in the FlashAttention process. As a result, calculating the importance metric necessitates two additional passes to load the data and one more step to compute the attention scores. PagedAttention usually assumes that the length of \texttt{KV} \texttt{cache} for a request monotonically increases. Sparsity-based methods execute token eviction at a fixed interval. Thus, the length of the remaining \texttt{KV} \texttt{cache} fluctuates over time, exacerbating the complexity of \texttt{KV} \texttt{cache} management with PagedAttention. \textbf{The integration of sparsity-based methods with FlashAttention and PagedAttention increases the implementation complexity and compromises potential computational efficiency advantages, underscoring the necessity for targeted systematic optimizations.}

\subsubsection{Evaluation Settings} 
Table~\ref{tab:summary-kv-cache} collects the evaluation settings of various \texttt{KV} \texttt{cache} compression algorithms. First, many research works spare more empirical analysis on accuracy than computational efficiency. The research trends that incorporate undesirable design elements in \texttt{KV} \texttt{cache} compression may preserve the accuracy but fail to facilitate computational gains because these design elements are not sufficient to exploit GPU parallelism. Notably, many studies only report the throughput for assessing the computational efficiency of the TRL framework, yet the design of compression algorithms possesses features that may not align with established optimizations for LLM serving. As a result, the potential for performance speedup becomes obscured when deploying compression algorithms in production environments.

Second, around half of the quantization-based algorithms are evaluated on models with a maximum size of 13 billion parameters and a maximum sequence length of 20 thousand. Statistically, more sparsity-based works evaluate larger model sizes (70B) and longer prompt lengths (200K) than quantization-based ones. This necessitates tensor parallelism to support such extreme scenarios on multiple GPUs.

\begin{tcolorbox}[colback=blue!5!white,colframe=gray!75!black,left=1mm, right=1mm, top=0.5mm, bottom=0.5mm, arc=1mm]
    \refstepcounter{surveyobservation}
    \textbf{Missing Piece \thesurveyobservation}: With less focus on computational efficiency, only a few compression studies measure the throughput performance using the TRL framework, often neglecting serving techniques, including FlashAttention and PagedAttention. Additionally, there is a lack of assessment regarding the throughput on multiple GPUs, which is crucial for supporting large models and long sequences through tensor parallelism. 
\end{tcolorbox}

Third, the throughput performance of different compression algorithms is typically evaluated with a fixed response length. The response length is a crucial factor that impacts the end-to-end latency of LLM serving requests. The response length is known when a termination token (e.g., \texttt{EOS}) is present. Thus, measuring the computational efficiency with a fixed response length is not an appropriate approach. In a realistic macro-benchmark, it is essential to account for the variations in the length distribution difference and assess the end-to-end latency performance of \texttt{KV} \texttt{cache} compression techniques for a fair comparison.

\begin{tcolorbox}[colback=blue!5!white,colframe=gray!75!black,left=1mm, right=1mm, top=0.5mm, bottom=0.5mm, arc=1mm]
    \refstepcounter{surveyobservation}
    \textbf{Missing Piece \thesurveyobservation}: The end-to-end latency variations from compression algorithms hinge not only on the throughput but also on the response length. However, the effect of compression algorithms on the response length has been largely neglected in existing compression studies. 
\end{tcolorbox}

\begin{table*}
\centering
\footnotesize
\caption{Summary of \texttt{KV} \texttt{cache} compression benchmarks. W/A/KV denote weights, activations, and \texttt{KV} \texttt{cache}, respectively.}
\begin{adjustbox}{width=\textwidth} 
\begin{tabular}{lccccc} 
\toprule
\textbf{Benchmarks}                & \textbf{Tasks}                                                                                                                 & \textbf{Models}                                                                         & \textbf{Metrics}                                                 & \textbf{Eval Methods }                                          & \textbf{Benchmark Features  }                                                       \\ 
\hline
QLLM-Eval~\cite{li2024evaluating}                 & \begin{tabular}[l]{@{}l@{}}Basic NLP Tasks\\Emergent Ability\\Trustworthiness\\Dialogue\\Long-context Tasks\end{tabular}    & \begin{tabular}[l]{@{}l@{}}OPT\\LLaMA2\\Falcon\\Mistral \end{tabular}      & \begin{tabular}[l]{@{}l@{}}Acc\end{tabular}        & W/A/KV Quant                                                & Evaluation of quant impacts for efficient deployment     \\ 
\hline
LLM-QBench~\cite{gong2024llm}                & \begin{tabular}[l]{@{}l@{}} WikiText2, C4\\ Exam \& Coding\end{tabular} & \begin{tabular}[l]{@{}l@{}}LLaMA2\\ChatGLM\\CodeLLaMA\\WizardMath\end{tabular} & \begin{tabular}[l]{@{}l@{}}Acc\\Throughput\end{tabular} & W/A/KV Quant                                                & A plug-and-play toolkit to explore quant impacts  \\ 
\hline
LongCTX-Bench~\cite{yuan2024kv}             & Long-context Tasks                                                                                                    & \begin{tabular}[l]{@{}l@{}}Mistral\\LLaMA\end{tabular}               & Acc                                                     & \begin{tabular}[l]{@{}l@{}}KV Quant\\KV Sparse\end{tabular} & KV compression evaluations in a long-context environment     \\ 
\hline
Shi et al. (2024)~\cite{luohe2024keep} & Basic NLP Tasks & LLaMA                                                                          & \begin{tabular}[l]{@{}l@{}}Acc\end{tabular}        & \begin{tabular}[l]{@{}l@{}}KV Quant\\KV Sparse\end{tabular} & A systematic review on KV optimizations without empirical results                                \\
\bottomrule
\end{tabular}
\label{tab:summary-kv-cache-benchmark}
\end{adjustbox}
\vspace{-15pt}
\end{table*}

\subsection{\texttt{KV} \texttt{Cache} Compression Benchmark Studies}
\label{subsec:survey-benchmarks}
The surge of \texttt{KV} \texttt{cache} compression algorithms drives several benchmark studies. We review them in Table~\ref{tab:summary-kv-cache-benchmark} and dissect them from tasks, models, and metrics. We identify several insights and pinpoint a gap in the accuracy performance demonstration within current studies.

First, for evaluation metrics, only LLM-QBench~\cite{gong2024llm} measures the prefill and decoding throughput. These benchmark studies generally suggest that compression algorithms result in only minor overall performance degradation. However, the aggregated numerical value may obscure the biases introduced by \texttt{KV} \texttt{cache} compression toward specific tasks and sample types. \textbf{Existing studies lack an in-depth analysis of how \texttt{KV} \texttt{cache} compression affects the response quality of individual examples.}

Second, for evaluation models, accuracy performance analysis has been primarily conducted using LLM families such as LLaMA \cite{touvron2023llama} and Mistral \cite{jiang2023mistral}. Their guidance on \texttt{KV} \texttt{cache} compression techniques is derived from empirical studies across LLM families. For example, LLM-QBench~\cite{li2024evaluating} recommends adjusting the quantization bit-widths based on the specific LLM family. \textbf{This underscores the importance of investigating accuracy performance within the LLaMA and Mistral families to gain comprehensive insights.}

Third, for evaluation tasks, a benchmark demands a comprehensive understanding of the LLMs' capabilities in a multidimensional manner. Existing studies~\cite{li2024evaluating,gong2024llm,luohe2024keep} assess a wide range of language tasks, including language understanding, modeling and reasoning, emergent abilities, dialogue, and long-context tasks. According to these analyses, long-context tasks exhibit lower tolerance to \texttt{KV} \texttt{cache} compression. \textbf{A detailed response quality analysis to individual samples in long-context tasks is imperative to understand the limitations of \textbf{KV cache} compression.}

\begin{tcolorbox}[colback=blue!5!white,colframe=gray!75!black,left=1mm, right=1mm, top=0.5mm, bottom=0.5mm, arc=1mm]
    \refstepcounter{surveyobservation}
    \textbf{Missing Piece \thesurveyobservation}: Long-context tasks are challenging for \texttt{KV} \texttt{cache} compression algorithms to maintain the accuracy. Many works report the overall performance, but they often overlook the analysis of response quality for individual samples. 
\end{tcolorbox}

\section{Evaluation}
\label{sec:evaluation}
In this section, we evaluate \texttt{KV} \texttt{cache} compression, with the focus on three missing aspects: \textit{throughput analysis}, \textit{response length distribution}, and \textit{negative sample analysis}. 

\begin{figure*}[!h]
    \centering
    \includegraphics[width=.32\textwidth]{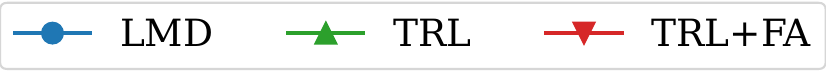}

    \subfigure[Decode, KV Length 256]{
        \includegraphics[width=0.22\textwidth]{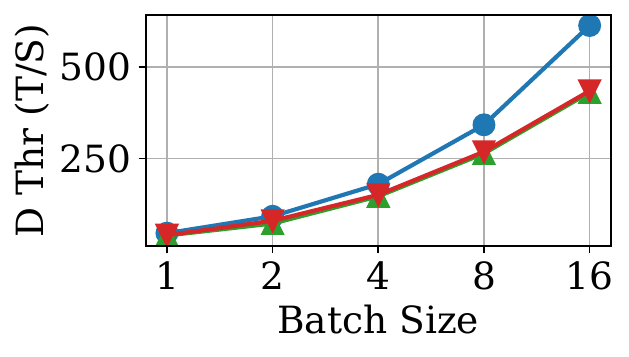}
        \label{fig:frw_thr_qaunt_16_policy_None_phase_decoding_promptlen_256}
    }
    \subfigure[Decode, KV Length 2048]{
        \includegraphics[width=0.22\textwidth]{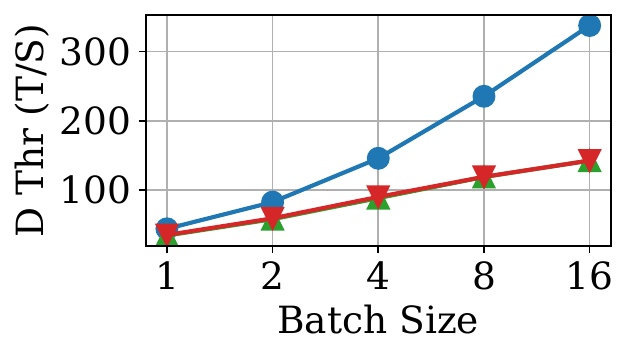}
        \label{fig:frw_thr_qaunt_16_policy_None_phase_decoding_promptlen_2048}
    }
    \subfigure[Decode, KV Length 1024]{
        \includegraphics[width=0.22\textwidth]{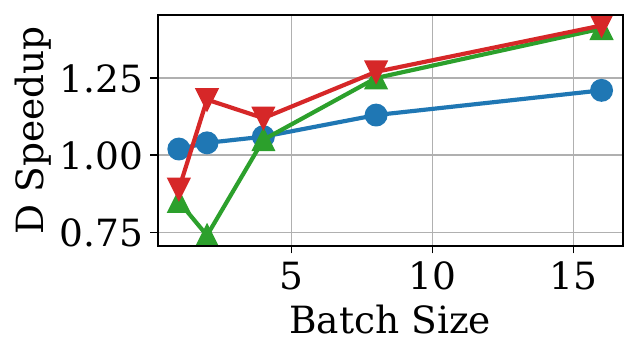}
        \label{fig:frw_speedup_16_StreamingLLM_normal_phase_decoding_promptlen_1024}
    }
    \subfigure[Decode, KV Length 2048]{
        \includegraphics[width=0.22\textwidth]{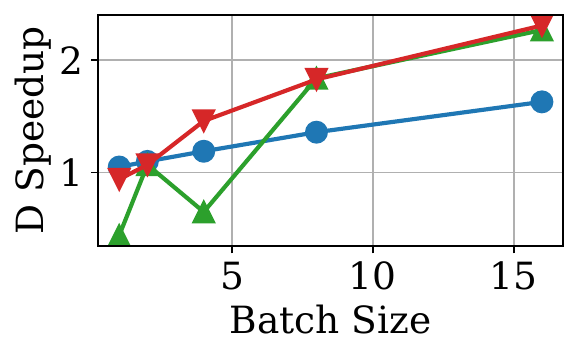}
        \label{fig:frw_speedup_16_StreamingLLM_normal_phase_decoding_promptlen_2048}
    }

    \centering
    \includegraphics[width=.5\textwidth]{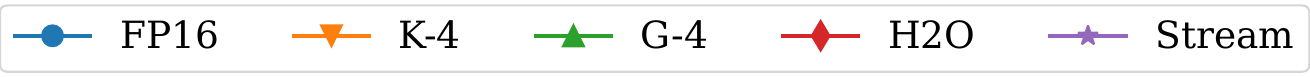}

    \subfigure[Prefill, Prompt 1024]{
        \includegraphics[width=0.22\textwidth]{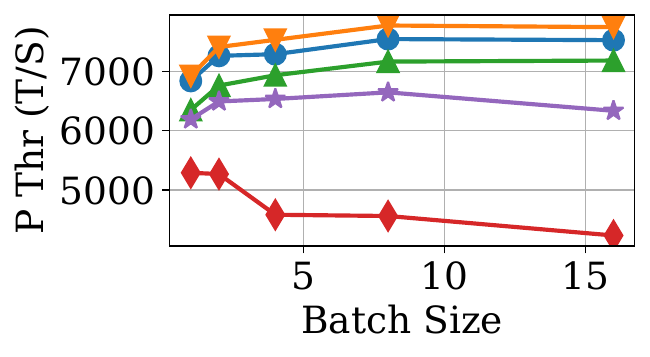}
        \label{fig:normal_phase_prefill_promptlen_1024}
    }
    \subfigure[Prefill, Batch 1]{
        \includegraphics[width=0.22\textwidth]{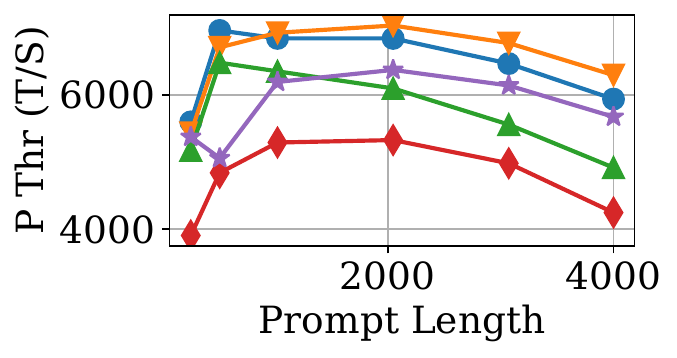}
        \label{fig:normal_phase_prefill_bsz_1}
    }
    \subfigure[Prefill, Prompt 6144]{
        \includegraphics[width=0.22\textwidth]{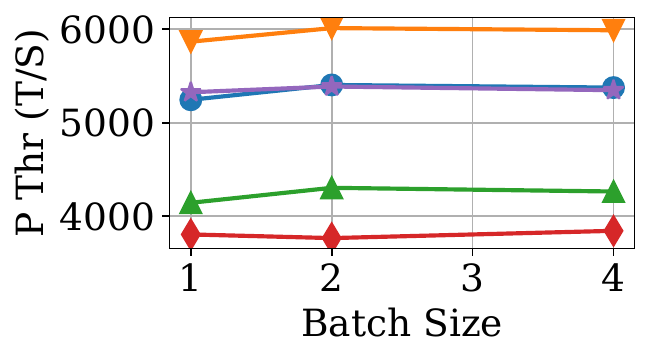}
        \label{fig:long_phase_prefill_promptlen_6144}
    }
    \subfigure[Prefill, Batch 1]{
        \includegraphics[width=0.22\textwidth]{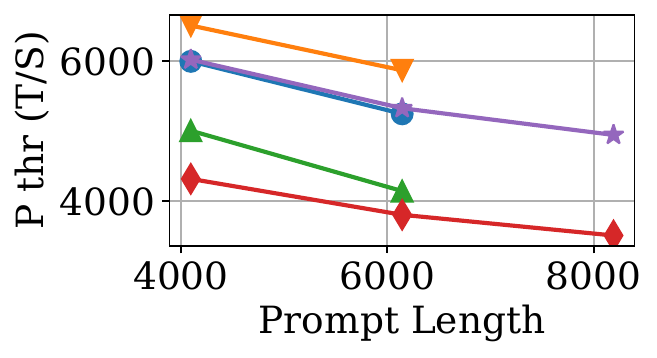}
        \label{fig:long_phase_prefill_bsz_1}
    }
    \centering
    \subfigure[Decode, KV Length 1024]{
        \includegraphics[width=0.22\textwidth]{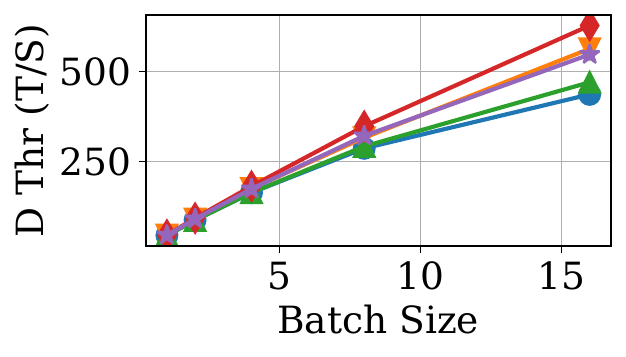}
        \label{fig:normal_phase_decoding_promptlen_1024}
    }
    \subfigure[Decode, Batch 1]{
        \includegraphics[width=0.22\textwidth]{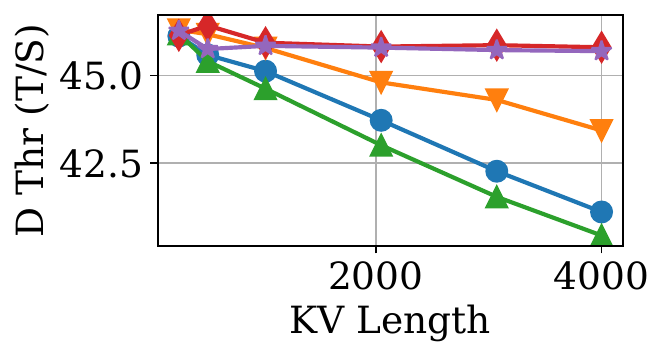}
        \label{fig:normal_phase_decoding_bsz_1}
    }
    \subfigure[Decode, KV Length 6144]{
        \includegraphics[width=0.22\textwidth]{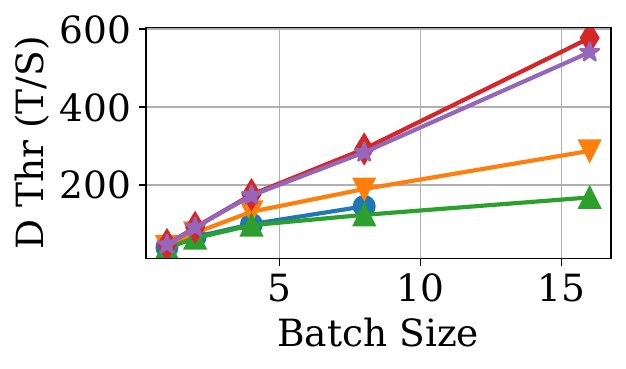}
        \label{fig:long_phase_decoding_promptlen_6144}
    }
    \subfigure[Decode, Batch 1]{
        \includegraphics[width=0.22\textwidth]{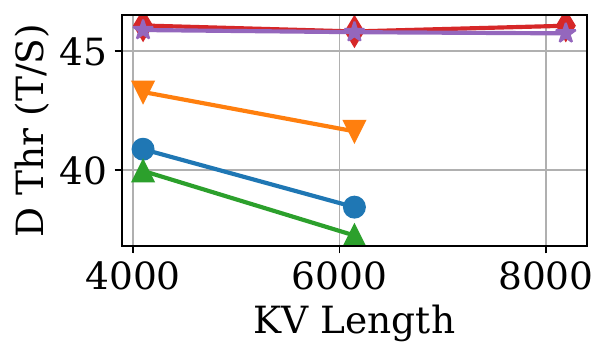}
        \label{fig:long_phase_decoding_bsz_1}
    }
\vspace{-10pt}
\caption{Throughput analysis of LLaMA-7B: (a-b) The FP16 decoding throughput on TRL (with and without FlashAttention) and LMDeploy (LMD). (c-d) The speedup of the StreamingLLM algorithm on TRL and LMD. (e-h) The prefill throughput for various sizes of inputs. (i-l) The decoding throughput for various sizes of inputs.}
\label{fig:thr-analysis}
\vspace{-10pt}
\end{figure*}



\begin{figure}[htbp]
\centering
    \includegraphics[width=.4\textwidth]{Figs/lmd/thr_legend.pdf}
    \subfigure[Prefill, Batch 1]{
        \includegraphics[width=0.2\textwidth]{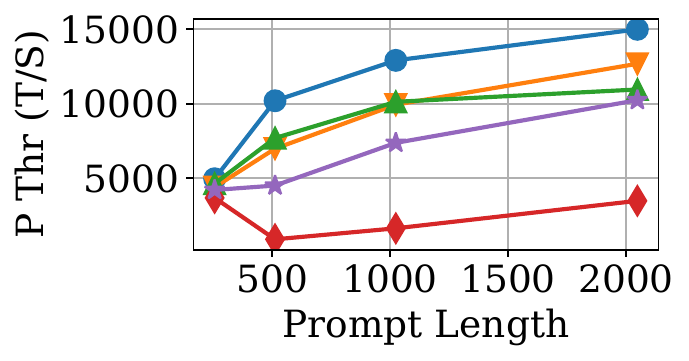}
        \label{fig:H800_normal_phase_prefill_bsz_1_L70}
        }
    \subfigure[Decode, Batch 1]{
        \includegraphics[width=0.18\textwidth]{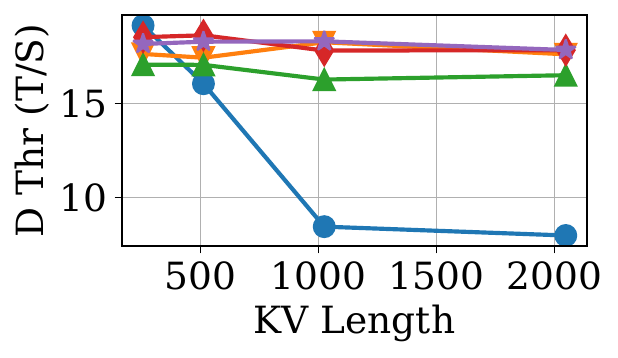}
        \label{fig:H800_normal_phase_decoding_bsz_1_L70}
    }
    \vspace{-10pt}
    \caption{Throughput analysis of LLaMA-70B on H800 GPUs.}
    \vspace{-20pt}
    \label{fig:H800-thr-analysis}
\end{figure}

\subsection{Evaluation Recipes}
\label{subsec:evaluation-recipes}

We single out representative models, datasets, and algorithms based on the above literature analysis. The majority of the experiments are conducted on A6000 GPUs with LLM serving frameworks, including TRL, FlashAttention~\cite{dao2023flashattention2}, and LMDeploy~\cite{2023lmdeploy}. We have also extended the scope to H800 GPU in Figure~\ref{fig:H800-thr-analysis}. We select LMDeploy for its efficient quantization kernels and fast development of KV cache compression algorithms. Appendix~\ref{app:evaluation-details} offers additional explanations and comprehensive details about the evaluation setup, further validating the rationale behind the selection of LMDeploy.


\noindent\textbf{LLMs.} Inspired by previous benchmark studies and implementation complexity, our evaluation only chooses LLaMA and Mistral families. We integrate corresponding compression algorithms and LLMs into TRL and LMDeploy.

\noindent\textbf{Datasets.} We choose ShareGPT~\cite{sharegpt_vicuna_unfiltered} to conduct the throughput analysis. This dataset is commonly used for LLM-serving benchmarks in real-world applications. We choose LongBench~\cite{bai2023longbench} to perform negative sample analysis. This is a task for long-context understanding that covers critical long-text application scenarios.

\noindent\textbf{Metrics.} For computational efficiency, we concentrate on the throughput and end-to-end latency. In particular, we report pre-fill and decoding throughput using synthesized examples and measure the end-to-end latency using samples from ShareGPT. For negative sample analysis, LongBench provides task-specific metrics to evaluate the accuracy of individual samples.

\noindent\textbf{Compression Algorithms.} We choose four compression algorithms for evaluation. For quantization-based ones, KIVI~\cite{KIVI} is a mainstream solution for quantizing key and value tensors. GEAR~\cite{GEAR} is a popular quantization error mitigation algorithm. For sparsity-based ones, StreamingLLM~\cite{StreamingLLM} only keeps first and recent tokens without complex attention score computation and presents a structured computation pattern. H2O~\cite{H2O} dynamically evicts \texttt{KV} \texttt{cache} with complex attention score computations.

\subsection{Throughput Analysis}
\label{subsec:latency-throughput-analysis}
We first utilize synthesized samples and LLaMA-7B to analyze the prefill and decoding throughput of LLM serving acceleration techniques. We measure the decoding throughput of full-precision baseline on TRL, TRL+PagedAttention, and LMDeploy, which possess the functionality of PagedAttention and FlashAttention. As shown in Figure~\ref{fig:thr-analysis} (a-b), PagedAttention and FlashAttention can improve the decoding throughput. Moreover, Figure~\ref{fig:thr-analysis} (c-d) shows the relative speedup of decoding throughput between the FP16 baseline and StreamingLLM across batch sizes by fixing \texttt{KV} lengths. When the batch size exceeds 4 and the sequence length reaches 1024, the relative speedup on TRL does not show a significant advantage when measured against PagedAttention and FlashAttention.



\begin{tcolorbox}[colback=gray!5!white,colframe=gray!75!black,left=1mm, right=1mm, top=0.5mm, bottom=0.5mm, arc=1mm]
    \refstepcounter{observation}
    \textbf{Observation \theobservation}: The computational efficiency results on TRL are unreliable. An appropriate way is to measure the throughput performance on established LLM serving frameworks with prominent LLM serving techniques, including PagedAttention and FlashAttention. 
\end{tcolorbox}

\begin{figure}[htbp]
\centering
    \includegraphics[width=.4\textwidth]{Figs/lmd/thr_legend.pdf}
    \subfigure[Prefill, Batch 1]{
        \includegraphics[width=0.18\textwidth]{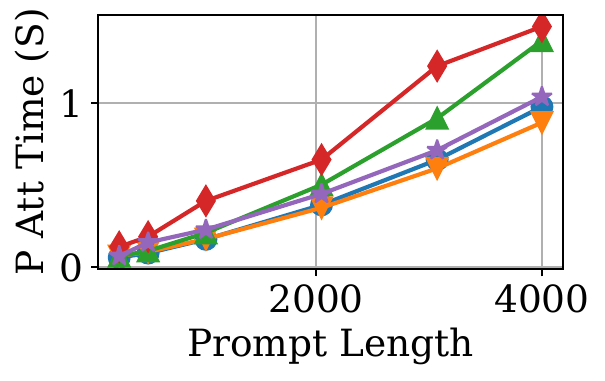}
        \label{fig:att-time-prefill}
        }
    \subfigure[Decode, Batch 1]{
        \includegraphics[width=0.18\textwidth]{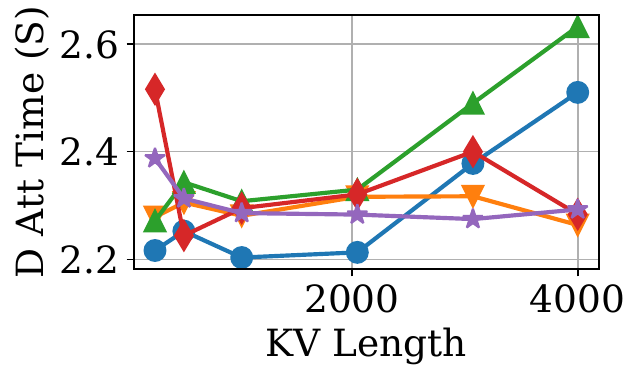}
        \label{fig:att-time-decode}
    }
    \vspace{-10pt}
    \caption{The execution time of the attention layer of various compression algorithms measured across different prompt lengths.}
    \vspace{-10pt}
    \label{fig:att-time-analysis}
\end{figure}

Second, we compare the prefill throughput of different compression algorithms. In particular, we collect the prefill throughput performance across different batch sizes and prompt lengths in Figure~\ref{fig:thr-analysis} (e-h). KIVI and StreamingLLM perform close to and even better than the FP16 baseline.  However, GEAR and H2O consistently lower the prefill throughput, with the gap widening as the prompt length increases. Qualitatively, GEAR introduces extra steps to offset the quantization error, and H2O requires a multi-pass attention operation to compute the attention score as a result of high memory access overhead. The execution time of the attention layer in GEAR and H2O in Figure~\ref{fig:att-time-prefill} further demonstrates that the additional \texttt{KV} \texttt{cache} compression overhead should not be disregarded in the prefill stage.

Third, we measure the decoding throughput for different compression algorithms, shown in Figure~\ref{fig:thr-analysis} (i-l). The throughput difference is insignificant when both batch size and KV length are small among different methods, including the baseline, as illustrated in Figure~\ref{fig:thr-analysis} (i-j). For heavy settings with long KV length and high batch size, sparsity-based methods can maintain their throughput advantages, whereas the benefits of quantization-based methods tend to diminish. We also observe that quantization-based methods even suffer from out-of-memory issues when the KV length reaches up to 8192 in Figure~\ref{fig:thr-analysis} (l). Moreover, Figure~\ref{fig:att-time-decode} depicts the execution time of the attention layer in sparsity-based methods remains more stable during the decoding stage across various KV lengths, as they retain a relatively small \texttt{KV} \texttt{cache}. For larger LLMs, we observe a similar phenomenon in decoding throughput, with results provided in Appendix~\ref{app:experimental-throughput-analysis}.

\begin{table}[!htp]\centering
\vspace{-5pt}
\caption{Relative speedup brought by different compression algorithms in the prefill and decoding.}\label{tab:tp-analysis}
\begin{adjustbox}{width=0.45\textwidth} 
\begin{tabular}{lllcccc}\toprule
\textbf{} &\textbf{TP/Algo} &\textbf{FP16 (Thr: T/S)} &\textbf{K-4} &\textbf{G-4} &\textbf{H2O} &\textbf{Stream} \\\midrule
\multirow{3}{*}{Prefill} &1 &6610.24 &1.06$\times$ &0.86$\times$ &0.58$\times$ &0.95$\times$ \\
&2 &11041.35 &1.09$\times$ &0.80$\times$ &0.58$\times$ &0.96$\times$ \\
&4 &12938.66 &1.03$\times$ &0.90$\times$ &0.51$\times$ &0.92$\times$ \\
\midrule
\multirow{3}{*}{Decode} &1 &129.72 &0.98$\times$ &1.02$\times$ &1.34$\times$ &1.34$\times$ \\
&2 &194.83 &0.88$\times$ &0.97$\times$ &0.69$\times$ &1.01$\times$ \\
&4 &195.02 &0.9$\times$ &0.97$\times$ &0.85$\times$ &0.97$\times$ \\
\bottomrule
\end{tabular}
\end{adjustbox}
\vspace{-5pt}
\end{table}

Fourth, we report the relative speedup in the prefill and decoding throughput of different compression algorithms compared to the FP16 baseline across various tensor parallelism settings in Table~\ref{tab:tp-analysis}. While tensor parallelism can improve throughput, it may also reduce the throughput speedup gained from \texttt{KV} \texttt{cache} compression and, in some cases, even negatively impact overall throughput performance. We ascribe this to that increasing tensor parallelism can alleviate memory bandwidth contention on each GPU, thereby weakening the benefits of reduced memory access overhead achieved through \texttt{KV} \texttt{cache} compression. Note that Appendix~\ref{app:experimental-throughput-analysis} consists of more throughput analyses across different batch sizes, sequence lengths, LLMs (e.g., Mistral~\cite{jiang2023mistral} in Figure~\ref{fig:thr-analysis-M7}), and algorithms (e.g., SnapKV~\cite{SnapKV} in Figure~\ref{fig:thr-analysis-snapkv}), our conclusions remain consistent.

\begin{tcolorbox}[colback=gray!5!white,colframe=gray!75!black,left=1mm, right=1mm, top=0.5mm, bottom=0.5mm, arc=1mm]
    \refstepcounter{observation}
    \textbf{Observation \theobservation}: The \texttt{KV} \texttt{cache} compression methods show negative computational efficiency in certain scenarios of batch size, sequence length, and tensor parallelism in the prefill and decoding stage. We recommend applying compression algorithms for serving requests with heavy \texttt{KV} \texttt{cache}. 
\end{tcolorbox}

\begin{figure*}[htbp]
    \centering
    \subfigure[KIVI]{
        \includegraphics[width=0.22\textwidth]{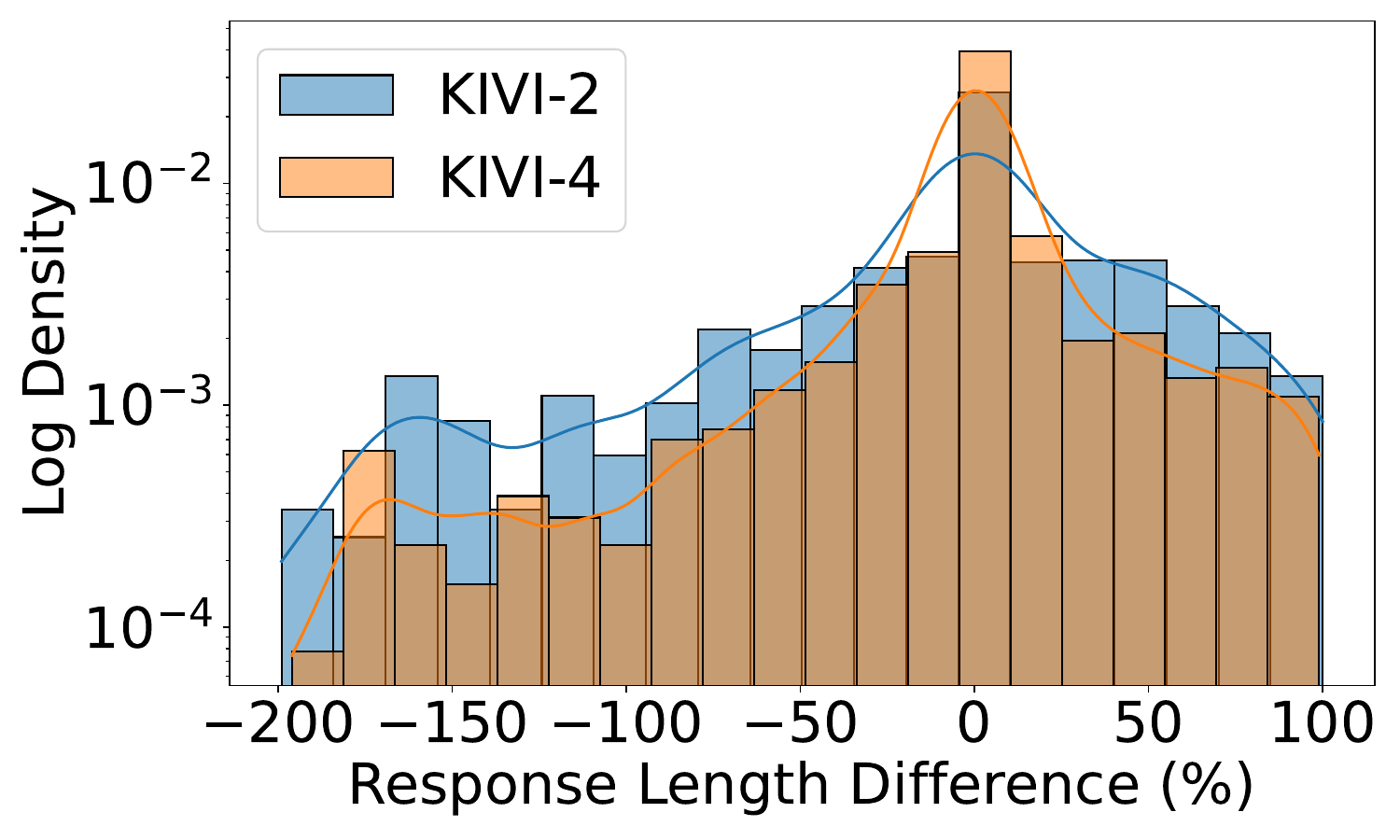}
        \label{fig:kivi-motivation_discrepancy}
        }
    \subfigure[GEAR]{
        \includegraphics[width=0.22\textwidth]{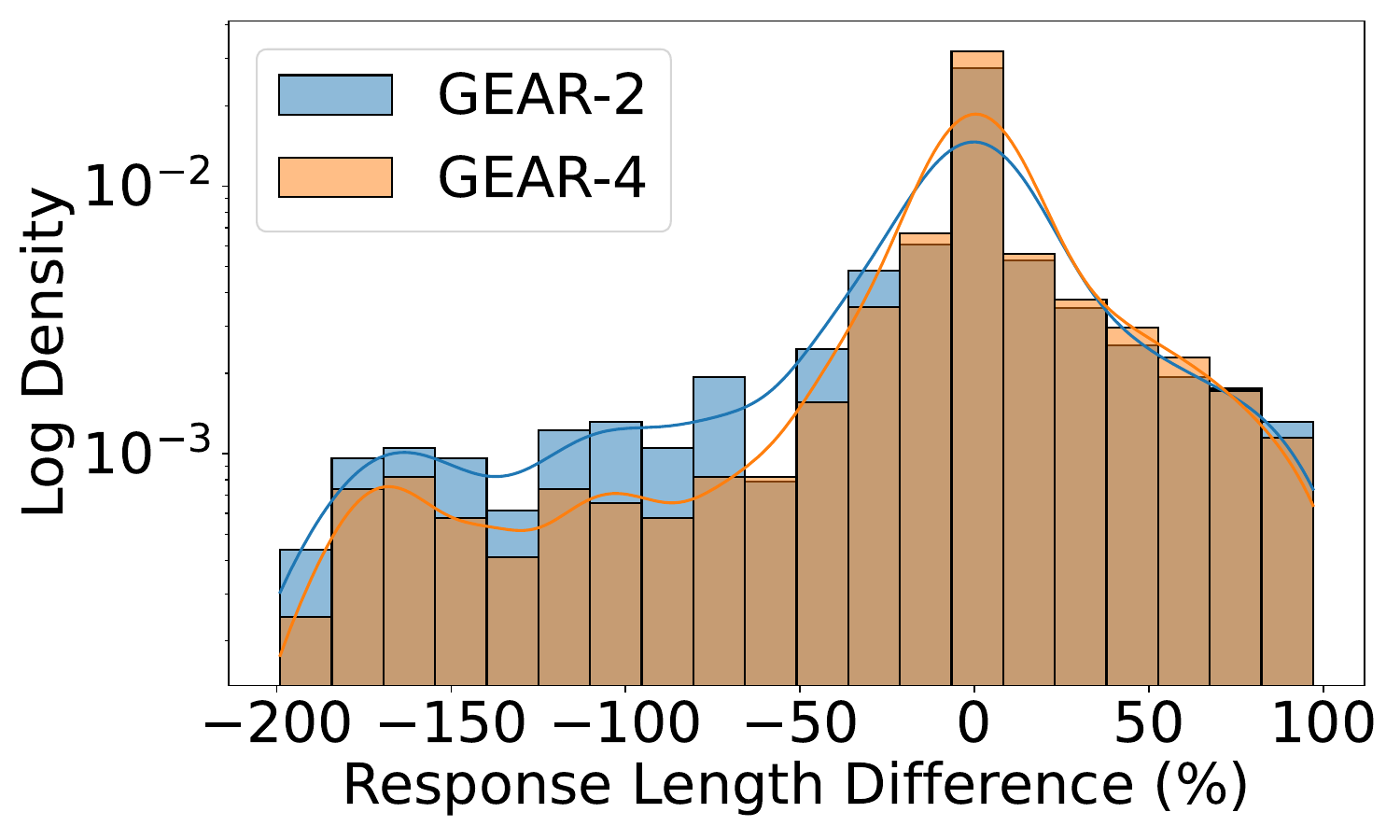}
        \label{fig:gear-motivation_discrepancy}
        }
    \subfigure[H2O]{\includegraphics[width=0.22\textwidth]{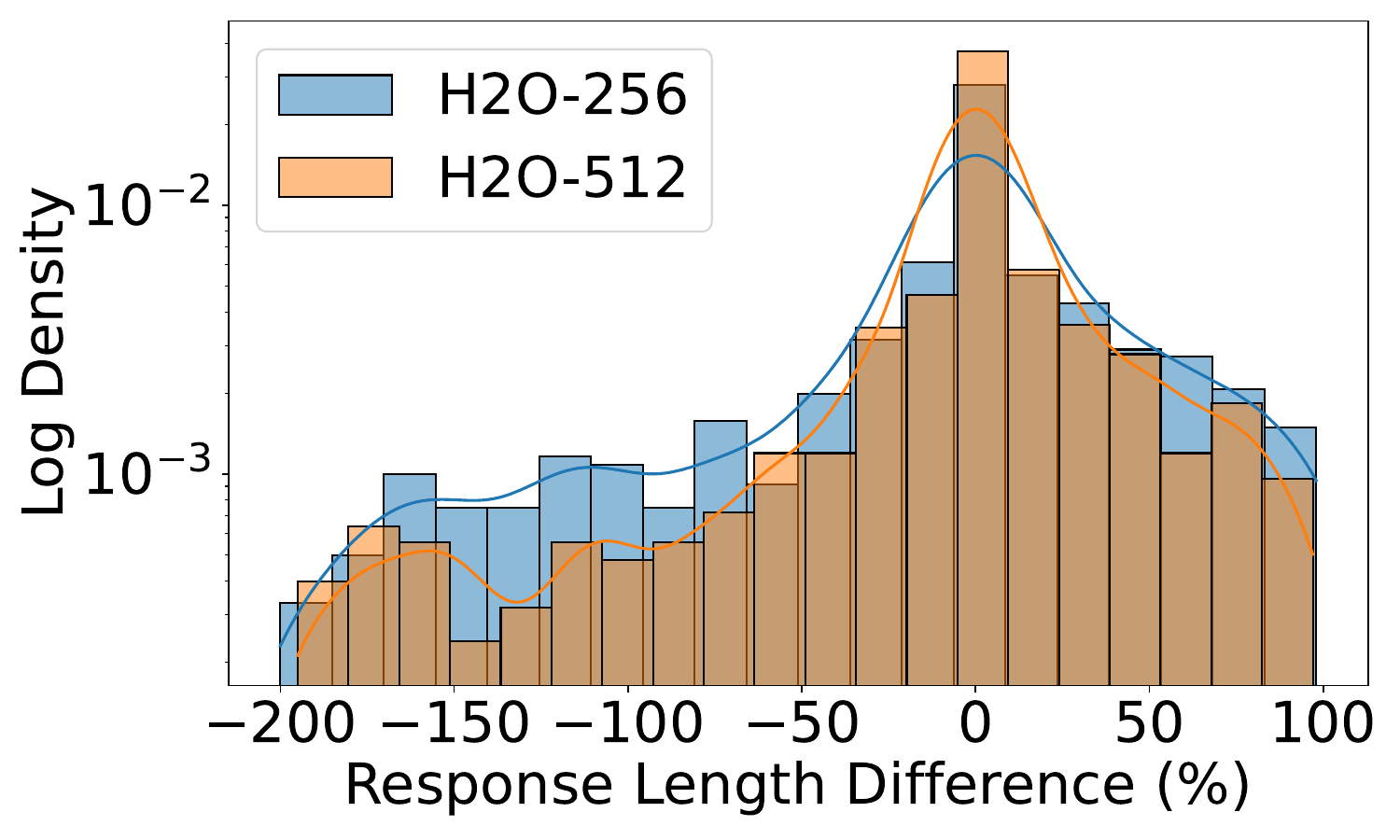}
        \label{fig:h2o-motivation_discrepancy}
    }
    \subfigure[StreamingLLM]{\includegraphics[width=0.22\textwidth]{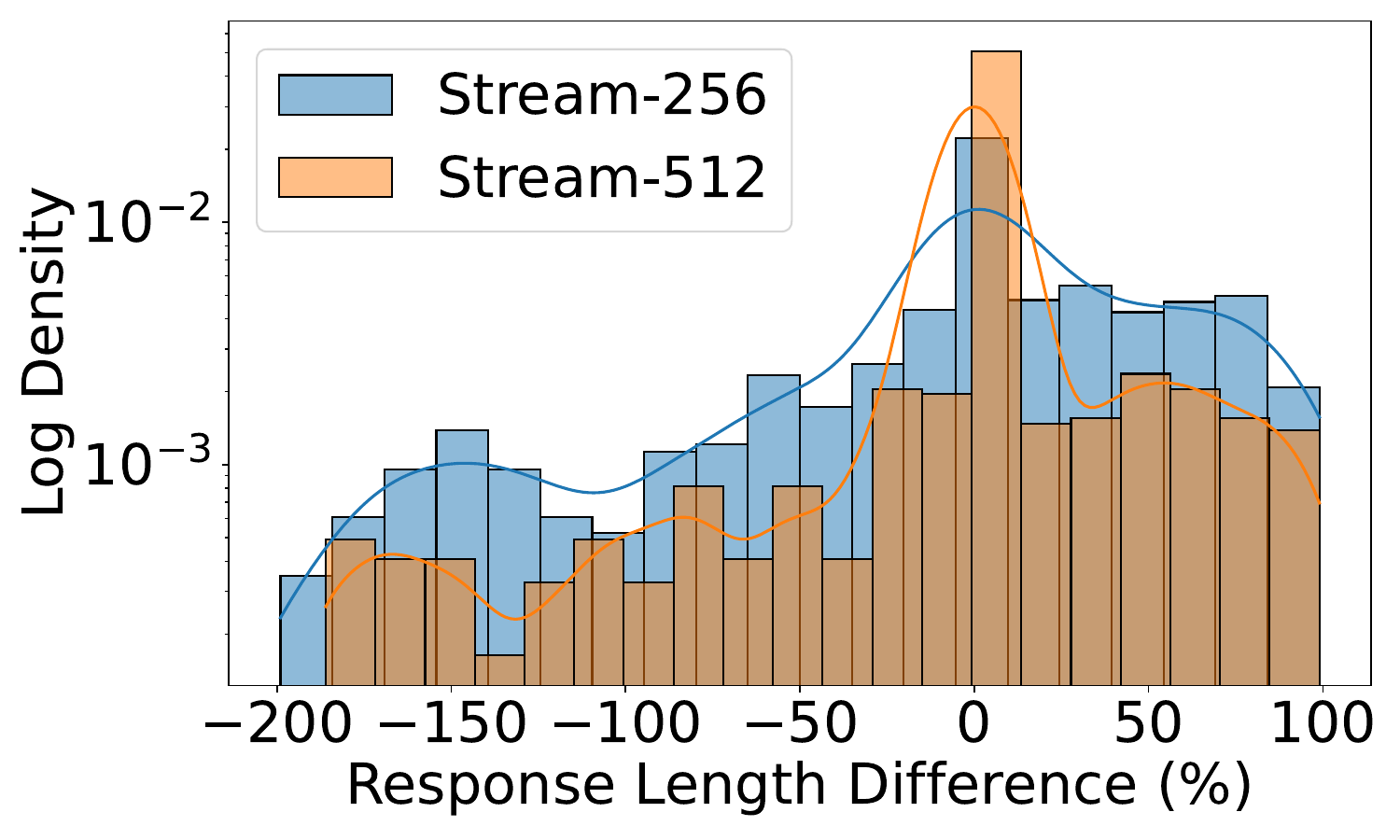}
        \label{fig:stream-motivation_discrepancy}
    }
    \vspace{-10pt}
    \caption{The log-scaled distribution of response length difference over different compression algorithms and configurations.} 
    \label{fig:length-difference-distribution}
    \vspace{-10pt}
\end{figure*}

\subsection{Length Distribution Analysis}
\label{subsec:length-distribution-analysis}
Lossy compression can elicit LLMs to yield different lengths of responses. Despite potential throughput benefits from \texttt{KV} \texttt{cache} compression, the lengthy responses can still prolong the end-to-end latency. We first examine the response length difference induced by compression methods. We define the response length difference as $D = (L^{\text{un}} - L^{\text{cs}})/L^{\text{un}}$, where $L^{\text{un}}$ and $L^{\text{cs}}$ represent the response length without and with compression methods, respectively. A negative $D$ implies that compression methods lead to longer outputs, while a positive $D$ suggests shorter responses due to compression. We use 1,000 samples from ShareGPT and measure $D$ on LLaMA-3.1-8B-instruct  using different \texttt{KV} \texttt{cache} compression algorithms. First, compression algorithms tend to increase the response length. We gather samples whose response length either decreases or increases by 50\%. Table~\ref{tab:response-length-variation} presents the proportion of samples exhibiting substantial variations in response length. Since the temperature in text generation affects response length, we use temperatures of $0.9$ and $1.1$ for a fair comparison while fixing the temperature as $1.0$ for the baseline and other compression algorithms. We observe that the temperature can increase and decrease the response length in roughly equal measure. Differently, \texttt{KV} \texttt{cache} compression leans toward generating lengthy responses. Specifically, more than 20\% of the samples show at least a 1.5 \(\times \) increase in response length. Prior throughput analysis shows that compression methods cannot achieve more than a \( 1.5 \times \) increase in the decoding throughput in many scenarios. This indicates that these samples will suffer from increased end-to-end latency due to their extended response length.

\begin{table}[t]
\centering
\caption{Comparison of semantic scores and length increase for different \texttt{KV} \texttt{cache} compression algorithms.}
\scriptsize
\resizebox{\linewidth}{!}{%
\begin{tabular}{l|ccccc}
\toprule
\textbf{Metric}     & \textbf{FP16}  & \textbf{KIVI-4} & \textbf{GEAR-4} & \textbf{H2O-512} & \textbf{Stream-512} \\ \midrule
Semantic Score      & 49.6           & 50.7                & 46.2                 & 46.2             & 46.3                \\ 
Length Increase ($\times$) & -        & 1.69                & 1.70                 & 1.55             & 1.76                \\ \bottomrule
\end{tabular}
\label{tab:semantic-length-comparison}}
\vspace{-10pt}
\end{table}

Second, we investigate the impact of compression ratio on the length difference distribution.  A higher compression ratio can be achieved with a lower bit in the quantization-based method or a lower length of \texttt{KV} \texttt{cache} in sparsity-based methods. We draw the distribution of the difference in response length (bar) and approximate the kernel density estimation of this distribution (line) across varying compression ratios in Figure~\ref{fig:length-difference-distribution}. We observe that with the increase of the compression ratio, the distribution of response difference flattens, and more samples experience lengthy response. The lengthy responses may serve as an implicit way for \texttt{KV} \texttt{cache} compression to compensate for accuracy loss.

To uncover whether \texttt{KV} \texttt{cache} compression techniques yield verbose output to improve the accuracy, we provide an intuitive experiment to investigate the verbosity of outputs. We define the verbose output as follows: for the output of an LLM with FP 16 baseline, its output quality and length are defined as $Q_{fp16}$ and $L_{fp16}$; for the output of compression baseline, its accuracy and length is defined as $Q_{compress}$ and $L_{fp16}$. We consider the output of compression baseline is verbose when $Q_{compress} \leq Q_{fp16}$ and  $L_{compress} \geq L_{fp16}$. 
We choose 200 requests from ShareGPT, in which KV cache compression techniques yield longer responses than FP 16 baseline when adopting LLaMA-7B. Fortunately, ShareGPT provides real-world conversations between humans and ChatGPT. To evaluate the semantic similarity between the outputs of ChatGPT and LLaMA-7B. We report the average semantic score and relative length increase in Table \ref{tab:semantic-length-comparison}. We find that three KV cache compression approaches produce longer outputs but with relatively minor semantic quality drops. Compared to the output from an LLM with FP 16, the output from an LLM with a KV cache compression approach might be more verbose.

\begin{table}[!t]\centering
\caption{The ratio (\%) of samples experiencing response length variations induced by temperature and \texttt{KV} \texttt{cache} compression. }\label{tab:response-length-variation}
\begin{adjustbox}{width=0.48\textwidth} 
\begin{tabular}{lccccccc}\toprule
\textbf{Metric} & \textbf{T=0.9} & \textbf{T=1.1} & \textbf{KIVI} &\textbf{GEAR} &\textbf{H2O} &\textbf{Stream} \\\midrule
\% of samples $D$ of which $ \geq 50\% $ & 20.8\% & 21.3\%  &10.9\% & 6.8\% &9.5\% & 16.5\% \\ 
\% of samples $D$ of which $ \leq -50\% $ & 27.5\% & 31.4\% &24.5\% &27.1\% &21.3\% & 26.4\% \\
\bottomrule
\end{tabular}
\end{adjustbox}
\vspace{-20pt}
\end{table}

\begin{tcolorbox}[colback=gray!5!white,colframe=gray!75!black,left=1mm, right=1mm, top=0.5mm, bottom=0.5mm, arc=1mm]
    \refstepcounter{observation}
    \textbf{Observation \theobservation}: The lossy compression contributes to a large variation of response length distribution. Compression algorithms produce lengthy responses. Furthermore, a high compression ratio exacerbates this issue. 
\end{tcolorbox}

\begin{figure}[htbp]
    \centering
    \includegraphics[width=0.45\textwidth]{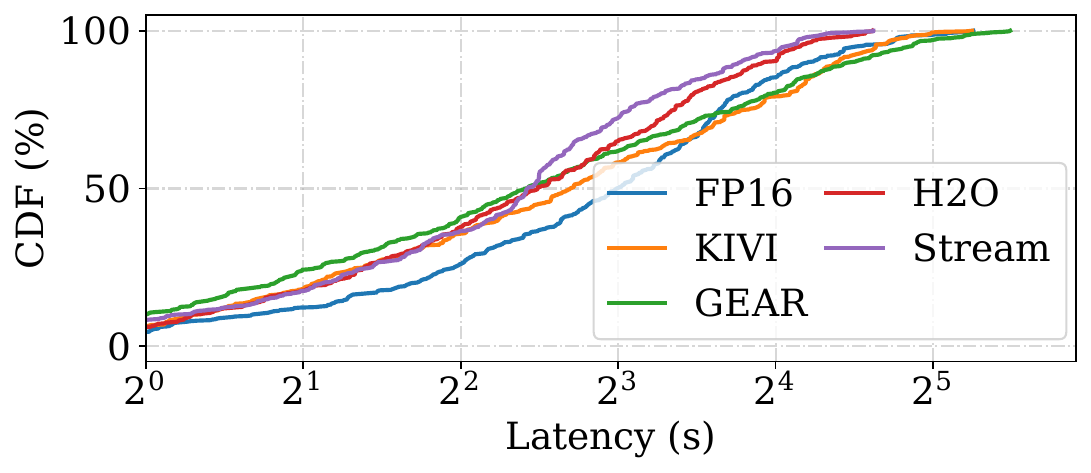}
    \vspace{-12pt}
    \caption{The cumulative distribution function of the end-to-end latency (seconds) of various compression algorithms.} 
    \label{fig:length-latency-difference-distribution}
    \vspace{-10pt}
\end{figure}

Third, beyond the length distribution analysis, we present the distribution of end-to-end latency to gain a vivid understanding of how length difference impacts computational efficiency. Specifically, we measure the end-to-end latency performance for each sample in the ShareGPT dataset with a fixed batch size of one, as shown in Figure~\ref{fig:length-latency-difference-distribution}. When we combine the throughput and response length, the performance gains of compression methods are not significant. We even observe that GEAR leads to longer end-to-end tail latency. In other words, only measuring the throughput performance with a fixed response length cannot convince LLM practitioners to deploy compression algorithms in production environments. We leave more empirical analysis about length distribution in Appendix~\ref{app:experiment-response-length-analysis}.

\begin{tcolorbox}[colback=gray!5!white,colframe=gray!75!black,left=1mm, right=1mm, top=0.5mm, bottom=0.5mm, arc=1mm]
    \refstepcounter{observation}
    \textbf{Observation \theobservation}: Measuring the end-to-end latency uncovers that compression methods still have a long way to go in practice. In addition to the innate algorithm defect of compression that might lower the throughput performance benefits, the lengthy response length is another factor that hinders their adoption. 
\end{tcolorbox}

\subsection{Negative Sample Analysis}
\label{subsec:failure-case-analysis}
Many studies on \texttt{KV} \texttt{cache} compression place emphasis on preserving the quality of LLM responses, supported by empirical evidence from a few benchmark datasets. While most of them report minor/no accuracy performance drops, they often overlook the impact of compression on individual samples. In other words, LLM practitioners lack insightful guidance on when \texttt{KV} \texttt{cache} compression fails to achieve satisfactory performance. To bridge such a gap, we explore the negative performance impact on individual samples. We follow the criterion in QLLM-Eval~\cite{li2024evaluating} to define the \textit{negative sample} as a benign sample where \texttt{KV} \texttt{cache} compression leads to the relative accuracy loss exceeding a given threshold\footnote{In the evaluation, we select samples with accuracy equal to or above the average value as benign.}. Algorithm~\ref{alg:failure-case-collection} describes the process of collecting negative samples given the LLM, dataset, and compression algorithms. The baseline algorithm is the one without using \texttt{KV} \texttt{cache} compression.

\begin{figure}[htbp]
    \centering
    \subfigure[Quantization]{
        \includegraphics[width=0.22\textwidth]{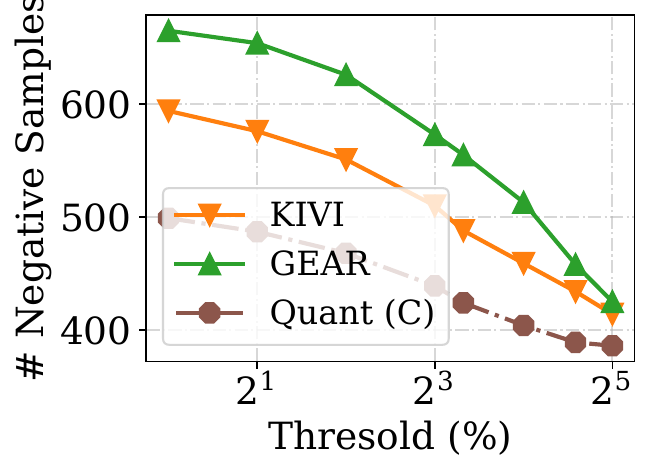}
        \label{fig:quant-failure}
        }
    \subfigure[Sparsity]{
        \includegraphics[width=0.22\textwidth]{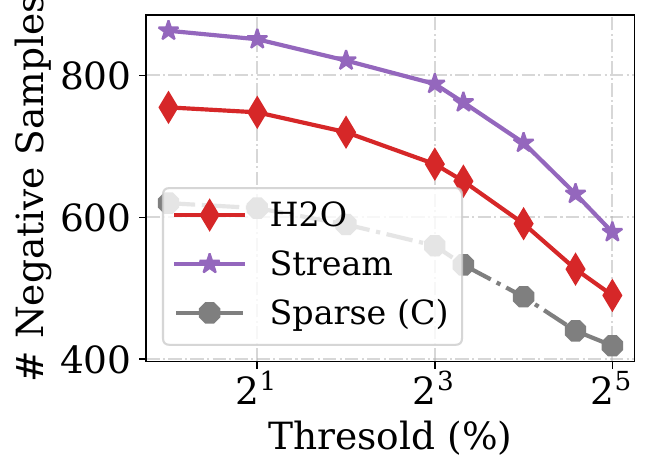}
        \label{fig:sparse-failure}
    }
    \vspace{-10pt}
    \caption{The threshold ($x$-axis) versus the number of negative samples ($y$-axis) for quantization-based (a) and sparsity-based (b) methods. Quant (C) refers to negative samples collected using both KIVI and GEAR together. Sparse (C) refers to negative samples collected using both H2O and StreamingLLM together. }
    \label{fig:compression-negative-analysis}
\end{figure}

First, we unveil the fragility of compression algorithms using negative samples. We use LongBench and LLaMA-3.1-8B-instruct to conduct negative sample analysis. The experimental details can be found in Appendix~\ref{app:experiment-neg-sample-analysis}. We vary the threshold in Algorithm~\ref{alg:failure-case-collection} to collect negative samples shown in Figure~\ref{fig:compression-negative-analysis}. The minor accuracy loss brought by compression algorithms (e.g., KIVI, GEAR) does not mean that each sample suffers from the minor performance loss. Our pinhole observation indicates a high number of negative samples even with a threshold of 10\%, revealing the fragility of compression algorithms. Combining an ensemble of algorithms to construct an algorithm set $\mathcal{A}$ is a feasible approach to reduce the occurrence of negative samples; however, they cannot be totally eliminated.

\begin{tcolorbox}[colback=gray!5!white,colframe=gray!75!black,left=1mm, right=1mm, top=0.5mm, bottom=0.5mm, arc=1mm]
    \refstepcounter{observation}
    \textbf{Observation \theobservation}: \texttt{KV} \texttt{cache} compression algorithms naturally possess negative samples, and the accuracy improvement can reduce the occurrence of negative samples, but it is hard to eradicate. 
\end{tcolorbox}

\begin{figure}[htbp]
 \vspace{-10pt}
    \centering
        \includegraphics[width=0.45\textwidth]{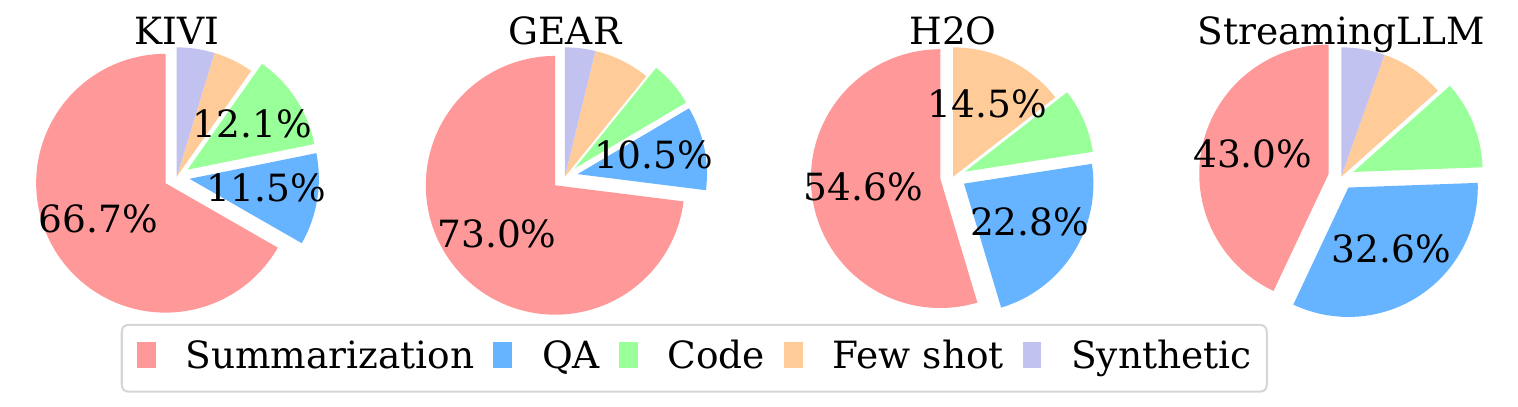}
    \vspace{-5pt}
    \caption{The pie chart details the proportion of negative samples over task types across varying compression algorithms.}
    \label{fig:compression-task-type}
    \vspace{-5pt}
\end{figure}

Second, we explore the sensitivity of task types to \texttt{KV} \texttt{cache} compression using a threshold of 10\%. Figure~\ref{fig:compression-task-type} depicts the breakdown of the number of negative samples across various tasks. Both quantization-based and sparsity-based methods exhibit a similar unbalanced fragility across different task types. Particularly, the summarization tasks depend heavily on context information, and any loss of this information can lead LLMs to produce undesirable responses. Similarly, in the question-answering (QA) task, information lost in the early stages can become significant in later stages, thereby amplifying this issue. Fortunately, a recent work, Quest~\cite{Quest}, proposes a query-aware approach to address this drawback.

\begin{tcolorbox}[colback=gray!5!white,colframe=gray!75!black,left=1mm, right=1mm, top=0.5mm, bottom=0.5mm, arc=1mm]
    \refstepcounter{observation}
    \textbf{Observation \theobservation}: Not all task types are equally treated by \texttt{KV} \texttt{cache} compression algorithms. The compression studies show limitations in maintaining accuracy for summarization and QA tasks.
\end{tcolorbox}

\begin{algorithm}[tb]
\footnotesize
\caption{Negative Sample Collection.} \label{alg:failure-case-collection}
\begin{algorithmic}[1]
   \STATE {\bfseries Input:} Dataset $\mathcal{D}$, Threshold $\theta$, LLM $\mathcal{M}$, Baseline Algorithm $\mathcal{A}_\text{b}$, Compression Algorithm Set $\mathcal{A}$
   \STATE {\bfseries Output:} Negative Dataset $\mathcal{D}_\text{neg}$
   
    \STATE
    \STATE {\bfseries Function} \texttt{AccMetric}$(\mathcal{A}, M, d)$:
        \STATE \quad {\bfseries Input:} Algorithm $\mathcal{A}$, Data point $d$
        \STATE \quad {\bfseries Output:} Accuracy $p$
        \STATE \quad Adopt compression algorithm $\mathcal{A}$ and LLM $M$ to produce the response $r$ for given prompt $d$.

        \STATE \quad Return the accuracy with the response $r$.
   \STATE {\bfseries End Function}
   \STATE
   \STATE Initialize $\mathcal{D}_\text{neg} = \varnothing$.
   \FOR{$d_i$ {\bfseries in} $\mathcal{D}$}
        \STATE $p_\text{base}$ = $\texttt{AccMetric}(\mathcal{A}_\text{base}, d_i)$
        \STATE $negative$ = {\bfseries true}
        \FOR {$\mathcal{A}_j$ {\bfseries in} $\mathcal{A}$}
            \IF {$\texttt{AccMetric}(\mathcal{A}_j, d_i) \geq (1 - \theta) \times p_\text{base} $ }
                \STATE $negative$ = {\bfseries false}
            \ENDIF
        \ENDFOR
    \IF {$negative$}
        \STATE Insert $d_i$ in $\mathcal{D}_\text{neg}$
   \ENDIF
   \ENDFOR
   
\end{algorithmic}
\end{algorithm}

\section{A Suite of Tools}
\label{sec:kv-cache-toolkit}

Given the above limitations of existing studies, we introduce a set of tools to assist \texttt{KV} \texttt{cache} compression. 

\subsection{Throughput Predictor}
First, it is common for online LLM serving systems to handle various sequence lengths and batch sizes in the prefill and decoding stages. In LLM serving, all other operations (e.g., linear product) are independent of \texttt{KV} \texttt{cache} and attention operation. Fortunately, \texttt{KV} \texttt{cache} compression primarily impacts the throughput performance of the attention operations. Hence, we profile the throughput of the attention layer across various sequence lengths and batch sizes in both prefill and decoding stages. We incorporate the offline profiled throughput results into the LLM runtime predictor developed by Vidur~\cite{vidur} to inform the LLM serving systems to make appropriate scheduling decisions. Table~\ref{tab:throughput-accuracy} reports the accuracy of our throughput predictor for LLaMA-7B. Our observation is that it can provide above 85\% prediction accuracy for various compression techniques. More experimental details about the throughput predictor can be found in Appendix~\ref{app:thr-predictor}.


\subsection{Length Predictor}
Recent works~\cite{zheng-length-predictor,qiu-length-predictor} demonstrate the potential for predicting the response length in LLMs. Following this direction, we gather response length from various \texttt{KV} \texttt{cache} compression algorithms and employ a BERT-based classifier to predict the length generated by a given compression algorithm. Table~\ref{tab:throughput-accuracy} reports the prediction results of our length predictor for LLaMA-3.1-8B-instruct, which can attain above 85\% accuracy across four representative \texttt{KV} \texttt{cache} algorithms. More experimental details and results about the length predictor are elaborated in Appendix~\ref{app:length-ratio-predictor}. Our length predictor can inform the online serving system to determine whether to apply compression techniques on incoming requests, thus mitigating the extended end-to-end latency.

\begin{table}[!htp]\centering
\vspace{-10pt}
\caption{The prediction accuracy of our proposed tools.}\label{tab:throughput-accuracy}
\scriptsize
\begin{tabular}{lrrrrrr}\toprule
\textbf{Tools} & \textbf{FP16} & \textbf{KIVI} &\textbf{GEAR} &\textbf{H2O} &\textbf{Stream} \\\midrule
Throughput Predictor   & 88.5\% & 88.4\% &87.7\% &85.8\% &86.6\% \\ 
Length Predictor & 89.3\% & 95.7\% &88.4\% &87.8\% &90.0\% \\
\bottomrule
\end{tabular}
\end{table}

\subsection{Negative Sample Evaluator}
Based on our empirical analysis, we set a 10\% threshold to identify negative samples and compile them into a benchmark dataset. This dataset evaluates both existing and future \texttt{KV} \texttt{cache} compression techniques.
Using LongBench's evaluation score, we measure the performance of four representative compression methods on LLaMA-3.1-8B-instruct, as reported in Table\ref{tab:benchmark-accuracy}.
The baseline (FP16) achieves high scores, but various \texttt{KV} \texttt{cache} compression algorithms show significant drops. More details are provided in Appendix \ref{app:perf-on-negative-bench}. We recommend further research into these negative samples to better understand the impact of \texttt{KV} \texttt{cache} compression algorithms on accuracy.

To mitigate negative samples, we recommend the following solutions to realize task-specific \texttt{KV} \texttt{cache} compression techniques. First, we can adopt a lightweight model to predict the task types of input requests for LLM serving. Second, we can develop task-specific \texttt{KV} \texttt{cache} compression approaches or adopt \texttt{KV} \texttt{cache} with varying compression levels. 




\begin{table}[!htp]\centering
\vspace{-10pt}
\caption{The measured score of various algorithms on the negative sample benchmark dataset using LongBench's provided metric.}\label{tab:benchmark-accuracy}

\scriptsize
\begin{tabular}{lccccc}\toprule
\textbf{Task Type} & \textbf{Baseline} & \textbf{KIVI} &\textbf{GEAR} &\textbf{H2O} &\textbf{Stream} \\\midrule
\textbf{Summarization}  & 31.6 & 24.8 & 23.7 & 24.7 & 24.3 \\ 
\textbf{Question Answering} & 52.0 & 28.8 & 28.7 & 33.8 & 30.4 \\ 
\textbf{Code}           & 97.0 & 30.0 & 30.0 & 57.2 & 61.3 \\ 
\bottomrule
\end{tabular}
\end{table}


\subsection{Usage of Tools: Request Router}
We leverage our throughput and length predictor to explore how both tools can be used to expedite online LLM serving via request routing. Particularly, we run LLaMA-7B on four A6000 GPUs with LMDeploy, and sample 1000 requests from ShareGPT, using Poisson distributions with request per second as 10. \textit{Baseline} refers to running LLaMA-7B with FP 16 or a given \texttt{KV} \texttt{cache} compression approach on four GPUs. It adopts a load-balancing technique to route incoming requests to a GPU with minimum memory usage. In the following three policies, we empirically use one GPU to run FP16 and three GPUs to run a given compression technique. We need to route an incoming request to an appropriate GPU. \textit{w/ Throughput} refers to routing a request to a GPU that can yield the estimated highest decoding throughput by the throughput predictor. \textit{w/ Length} refers to routing to a GPU, which can yield the estimated minimum response length by length predictor. \textit{w/ Both} refers to routing to a GPU that can yield the minimum estimated end-to-end latency. The end-to-end latency is calculated by the prefilling time and a product between the estimated decoding throughput and estimated response length. We report the average end-to-end latency (seconds) for FP16 and compression techniques in Table~\ref{tab:e2e-performance-router}. The throughput predictor speeds up the end-to-end latency by 1.18-1.48\(\times\). In contrast, the length predictor yields a speedup of 0.83-1.03\(\times\), suggesting that relying solely on the length predictor potentially compromises the latency. Combining the throughput predictor and length predictor speeds up the latency by 1.45 to 1.80\(\times\).

\begin{table}[t]\centering
\caption{Average end-to-end latency of different routing methods.}
\scriptsize
\begin{tabular}{l|ccccc}
\toprule
\textbf{Average E2E} & \textbf{FP16} & \textbf{KIVI} & \textbf{GEAR} & \textbf{H2O} & \textbf{Stream} \\ \midrule
Baseline         & 11.4          & 9.1           & 13.4          & 10.6         & 10.3            \\ 
w/ Throughput           & -             & 7.7           & 9.1           & 8.3          & 8.2             \\ 
w/ Length           & -             & 10.9          & 13            & 11.2         & 11.3            \\ 
w/ Both          & -             & 6.3           & 7.4           & 6.9          & 6.6             \\ \bottomrule
\end{tabular}
\label{tab:e2e-performance-router}
\end{table}

\section{Conclusion}
In this paper, we present a retrospective study of \texttt{KV} \texttt{cache} compression for LLM serving. We conduct a comprehensive literature survey and empirical analysis of existing algorithms, identifying several under-explored aspects of their practical usage. Our analysis reveals key difficulties that hinder the real-world deployment of \texttt{KV} \texttt{cache} compression. We recommend dissecting the evaluation of LLM \texttt{KV} \texttt{cache} compression algorithms into three critical dimensions, including throughput, length distribution, and negative samples. We gather insights from both literature and our evaluations to design three tools aimed at facilitating the applications of LLM \texttt{KV} \texttt{cache} compression algorithms in the production environment. 


\section*{Acknowledgement}
We thank the anonymous reviewers for their valuable comments. The research is supported under the RIE2020 Industry Alignment Fund - Industry Collaboration Projects (IAF-ICP) Funding Initiative, as well as cash and in-kind contributions from the industry partner(s).



\nocite{langley00}

\bibliography{0_references}

\begin{thebibliography}{68}
\providecommand{\natexlab}[1]{#1}
\providecommand{\url}[1]{\texttt{#1}}
\expandafter\ifx\csname urlstyle\endcsname\relax
  \providecommand{\doi}[1]{doi: #1}\else
  \providecommand{\doi}{doi: \begingroup \urlstyle{rm}\Url}\fi

\bibitem[kiv()]{kivi-issue-4}
Issue \#4: [integrate kivi into inference frameworks?].
\newblock \url{https://github.com/jy-yuan/KIVI/issues/4}.
\newblock Accessed: 2025.04.

\bibitem[Adnan et~al.(2024)Adnan, Arunkumar, Jain, Nair, Soloveychik, and Kamath]{Keyformer}
Adnan, M., Arunkumar, A., Jain, G., Nair, P., Soloveychik, I., and Kamath, P.
\newblock Keyformer: Kv cache reduction through key tokens selection for efficient generative inference.
\newblock \emph{Proceedings of Machine Learning and Systems}, 6:\penalty0 114--127, 2024.

\bibitem[Agrawal et~al.(2024)Agrawal, Kedia, Mohan, Panwar, Kwatra, Gulavani, Ramjee, and Tumanov]{vidur}
Agrawal, A., Kedia, N., Mohan, J., Panwar, A., Kwatra, N., Gulavani, B., Ramjee, R., and Tumanov, A.
\newblock Vidur: A large-scale simulation framework for llm inference, 2024.
\newblock URL \url{https://arxiv.org/abs/2405.05465}.

\bibitem[Aminabadi et~al.(2022)Aminabadi, Rajbhandari, Awan, Li, Li, Zheng, Ruwase, Smith, Zhang, Rasley, et~al.]{deepspeed-inference}
Aminabadi, R.~Y., Rajbhandari, S., Awan, A.~A., Li, C., Li, D., Zheng, E., Ruwase, O., Smith, S., Zhang, M., Rasley, J., et~al.
\newblock Deepspeed-inference: enabling efficient inference of transformer models at unprecedented scale.
\newblock In \emph{SC22: International Conference for High Performance Computing, Networking, Storage and Analysis}, pp.\  1--15. IEEE, 2022.

\bibitem[Anon(2024)]{sharegpt_vicuna_unfiltered}
Anon.
\newblock Sharegpt vicuna unfiltered dataset.
\newblock \url{https://huggingface.co/datasets/anon8231489123/ShareGPT_Vicuna_unfiltered}, 2024.
\newblock Accessed: 2024-06-13.

\bibitem[Anthropic(2024)]{claudeai}
Anthropic.
\newblock Claude ai, 2024.
\newblock URL \url{https://claude.ai/}.
\newblock Accessed: 2024-09.

\bibitem[Ashkboos et~al.(2024)Ashkboos, Mohtashami, Croci, Li, Jaggi, Alistarh, Hoefler, and Hensman]{QuaRot}
Ashkboos, S., Mohtashami, A., Croci, M.~L., Li, B., Jaggi, M., Alistarh, D., Hoefler, T., and Hensman, J.
\newblock Quarot: Outlier-free 4-bit inference in rotated llms.
\newblock \emph{arXiv preprint arXiv:2404.00456}, 2024.

\bibitem[Bai et~al.(2023)Bai, Lv, Zhang, Lyu, Tang, Huang, Du, Liu, Zeng, Hou, et~al.]{bai2023longbench}
Bai, Y., Lv, X., Zhang, J., Lyu, H., Tang, J., Huang, Z., Du, Z., Liu, X., Zeng, A., Hou, L., et~al.
\newblock Longbench: A bilingual, multitask benchmark for long context understanding.
\newblock \emph{arXiv preprint arXiv:2308.14508}, 2023.

\bibitem[{BentoML}()]{bentoml}
{BentoML}.
\newblock Benchmarking llm inference backends.
\newblock \url{https://bentoml.com/blog/benchmarking-llm-inference-backends}.
\newblock Accessed: 2025.03.

\bibitem[Chang et~al.(2024)Chang, Lin, Lin, Chen, Hu, Wang, Huang, Ceze, and Wu]{Palu}
Chang, C.-C., Lin, W.-C., Lin, C.-Y., Chen, C.-Y., Hu, Y.-F., Wang, P.-S., Huang, N.-C., Ceze, L., and Wu, K.-C.
\newblock Palu: Compressing kv-cache with low-rank projection.
\newblock \emph{arXiv preprint arXiv:2407.21118}, 2024.

\bibitem[Chen et~al.(2024)Chen, Wang, Shang, Cui, Zhang, Liu, Wang, Sun, Yu, and Wu]{NACL}
Chen, Y., Wang, G., Shang, J., Cui, S., Zhang, Z., Liu, T., Wang, S., Sun, Y., Yu, D., and Wu, H.
\newblock Nacl: A general and effective kv cache eviction framework for llms at inference time.
\newblock \emph{arXiv preprint arXiv:2408.03675}, 2024.

\bibitem[Contributors(2023)]{2023lmdeploy}
Contributors, L.
\newblock Lmdeploy: A toolkit for compressing, deploying, and serving llm.
\newblock \url{https://github.com/InternLM/lmdeploy}, 2023.

\bibitem[Dai et~al.(2024)Dai, Huang, Jiang, Chen, Cai, Bi, and Shi]{CORM}
Dai, J., Huang, Z., Jiang, H., Chen, C., Cai, D., Bi, W., and Shi, S.
\newblock Sequence can secretly tell you what to discard.
\newblock \emph{arXiv preprint arXiv:2404.15949}, 2024.

\bibitem[Dao(2024)]{dao2023flashattention2}
Dao, T.
\newblock Flash{A}ttention-2: Faster attention with better parallelism and work partitioning.
\newblock In \emph{International Conference on Learning Representations (ICLR)}, 2024.

\bibitem[Dao et~al.(2022)Dao, Fu, Ermon, Rudra, and R{\'e}]{dao2022flashattention}
Dao, T., Fu, D.~Y., Ermon, S., Rudra, A., and R{\'e}, C.
\newblock Flash{A}ttention: Fast and memory-efficient exact attention with {IO}-awareness.
\newblock In \emph{Advances in Neural Information Processing Systems (NeurIPS)}, 2022.

\bibitem[Dong et~al.(2024{\natexlab{a}})Dong, Yang, Zhang, Wang, Chi, and Chen]{LESS}
Dong, H., Yang, X., Zhang, Z., Wang, Z., Chi, Y., and Chen, B.
\newblock Get more with less: Synthesizing recurrence with kv cache compression for efficient llm inference.
\newblock \emph{arXiv preprint arXiv:2402.09398}, 2024{\natexlab{a}}.

\bibitem[Dong et~al.(2024{\natexlab{b}})Dong, Cheng, Qin, and Wang]{QAQ}
Dong, S., Cheng, W., Qin, J., and Wang, W.
\newblock Qaq: Quality adaptive quantization for llm kv cache.
\newblock \emph{arXiv preprint arXiv:2403.04643}, 2024{\natexlab{b}}.

\bibitem[Duanmu et~al.(2024)Duanmu, Yuan, Li, Duan, Zhang, and Lin]{SKVQ}
Duanmu, H., Yuan, Z., Li, X., Duan, J., Zhang, X., and Lin, D.
\newblock Skvq: Sliding-window key and value cache quantization for large language models.
\newblock \emph{arXiv preprint arXiv:2405.06219}, 2024.

\bibitem[Feng et~al.(2024)Feng, Lv, Cao, Xie, and Zhou]{Ada-KV}
Feng, Y., Lv, J., Cao, Y., Xie, X., and Zhou, S.~K.
\newblock Ada-kv: Optimizing kv cache eviction by adaptive budget allocation for efficient llm inference, 2024.
\newblock URL \url{https://arxiv.org/abs/2407.11550}.

\bibitem[Flashinfer(2024)]{flashinfer2024}
Flashinfer.
\newblock Flashinfer: A lightweight framework for inferencing.
\newblock \url{https://github.com/flashinfer-ai/flashinfer}, 2024.
\newblock Accessed: 2024-10.

\bibitem[Fu et~al.(2024)Fu, Cho, Merth, Mehta, Rastegari, and Najibi]{LazyLLM}
Fu, Q., Cho, M., Merth, T., Mehta, S., Rastegari, M., and Najibi, M.
\newblock Lazyllm: Dynamic token pruning for efficient long context llm inference.
\newblock \emph{arXiv preprint arXiv:2407.14057}, 2024.

\bibitem[Ge et~al.(2023)Ge, Zhang, Liu, Zhang, Han, and Gao]{FastGen}
Ge, S., Zhang, Y., Liu, L., Zhang, M., Han, J., and Gao, J.
\newblock Model tells you what to discard: Adaptive kv cache compression for llms.
\newblock \emph{arXiv preprint arXiv:2310.01801}, 2023.

\bibitem[Gong et~al.(2024)Gong, Yong, Gu, Huang, Zhang, Liu, and Tao]{gong2024llm}
Gong, R., Yong, Y., Gu, S., Huang, Y., Zhang, Y., Liu, X., and Tao, D.
\newblock Llm-qbench: A benchmark towards the best practice for post-training quantization of large language models.
\newblock \emph{arXiv preprint arXiv:2405.06001}, 2024.

\bibitem[Hassabis \& the Gemini~Team(2023)Hassabis and the Gemini~Team]{gemini2023}
Hassabis, D. and the Gemini~Team.
\newblock Introducing gemini: our largest and most capable ai model, 2023.
\newblock URL \url{https://blog.google/technology/ai/google-gemini-ai}.
\newblock Accessed: 2024-06-07.

\bibitem[He et~al.(2024)He, Zhang, Wu, Liu, Zhou, and Zhuang]{ZipCache}
He, Y., Zhang, L., Wu, W., Liu, J., Zhou, H., and Zhuang, B.
\newblock Zipcache: Accurate and efficient kv cache quantization with salient token identification.
\newblock \emph{arXiv preprint arXiv:2405.14256}, 2024.

\bibitem[Hooper et~al.(2024)Hooper, Kim, Mohammadzadeh, Mahoney, Shao, Keutzer, and Gholami]{KVQuant}
Hooper, C., Kim, S., Mohammadzadeh, H., Mahoney, M.~W., Shao, Y.~S., Keutzer, K., and Gholami, A.
\newblock Kvquant: Towards 10 million context length llm inference with kv cache quantization.
\newblock \emph{arXiv preprint arXiv:2401.18079}, 2024.

\bibitem[Jiang et~al.(2023)Jiang, Sablayrolles, Mensch, Bamford, Chaplot, Casas, Bressand, Lengyel, Lample, Saulnier, et~al.]{jiang2023mistral}
Jiang, A.~Q., Sablayrolles, A., Mensch, A., Bamford, C., Chaplot, D.~S., Casas, D. d.~l., Bressand, F., Lengyel, G., Lample, G., Saulnier, L., et~al.
\newblock Mistral 7b.
\newblock \emph{arXiv preprint arXiv:2310.06825}, 2023.

\bibitem[Kang et~al.(2024)Kang, Zhang, Kundu, Jeong, Liu, Krishna, and Zhao]{GEAR}
Kang, H., Zhang, Q., Kundu, S., Jeong, G., Liu, Z., Krishna, T., and Zhao, T.
\newblock Gear: An efficient kv cache compression recipefor near-lossless generative inference of llm.
\newblock \emph{arXiv preprint arXiv:2403.05527}, 2024.

\bibitem[Kwon et~al.(2023{\natexlab{a}})Kwon, Li, Zhuang, Sheng, Zheng, Yu, Gonzalez, Zhang, and Stoica]{kwon2023efficient}
Kwon, W., Li, Z., Zhuang, S., Sheng, Y., Zheng, L., Yu, C.~H., Gonzalez, J., Zhang, H., and Stoica, I.
\newblock Efficient memory management for large language model serving with pagedattention.
\newblock In \emph{Proceedings of the 29th Symposium on Operating Systems Principles}, pp.\  611--626, 2023{\natexlab{a}}.

\bibitem[Kwon et~al.(2023{\natexlab{b}})Kwon, Li, Zhuang, Sheng, Zheng, Yu, Gonzalez, Zhang, and Stoica]{vllm}
Kwon, W., Li, Z., Zhuang, S., Sheng, Y., Zheng, L., Yu, C.~H., Gonzalez, J.~E., Zhang, H., and Stoica, I.
\newblock Efficient memory management for large language model serving with pagedattention.
\newblock In \emph{Proceedings of the ACM SIGOPS 29th Symposium on Operating Systems Principles}, 2023{\natexlab{b}}.

\bibitem[Li et~al.(2024{\natexlab{a}})Li, Ning, Wang, Liu, Shi, Yan, Dai, Yang, and Wang]{li2024evaluating}
Li, S., Ning, X., Wang, L., Liu, T., Shi, X., Yan, S., Dai, G., Yang, H., and Wang, Y.
\newblock Evaluating quantized large language models.
\newblock \emph{arXiv preprint arXiv:2402.18158}, 2024{\natexlab{a}}.

\bibitem[Li et~al.(2024{\natexlab{b}})Li, Huang, Yang, Venkitesh, Locatelli, Ye, Cai, Lewis, and Chen]{SnapKV}
Li, Y., Huang, Y., Yang, B., Venkitesh, B., Locatelli, A., Ye, H., Cai, T., Lewis, P., and Chen, D.
\newblock Snapkv: Llm knows what you are looking for before generation.
\newblock \emph{arXiv preprint arXiv:2404.14469}, 2024{\natexlab{b}}.

\bibitem[Liu et~al.(2024{\natexlab{a}})Liu, Liu, Pan, He, Haffari, and Zhuang]{MiniCache}
Liu, A., Liu, J., Pan, Z., He, Y., Haffari, G., and Zhuang, B.
\newblock Minicache: Kv cache compression in depth dimension for large language models.
\newblock \emph{arXiv preprint arXiv:2405.14366}, 2024{\natexlab{a}}.

\bibitem[Liu et~al.(2024{\natexlab{b}})Liu, Chen, Lu, Jiang, Han, Zhang, Chen, Zhang, Ding, Zhang, et~al.]{RetrievalAttention}
Liu, D., Chen, M., Lu, B., Jiang, H., Han, Z., Zhang, Q., Chen, Q., Zhang, C., Ding, B., Zhang, K., et~al.
\newblock Retrievalattention: Accelerating long-context llm inference via vector retrieval.
\newblock \emph{arXiv preprint arXiv:2409.10516}, 2024{\natexlab{b}}.

\bibitem[Liu et~al.(2024{\natexlab{c}})Liu, Bai, Lin, Li, Gao, Xu, Hou, Yao, and Yuan]{IntactKV}
Liu, R., Bai, H., Lin, H., Li, Y., Gao, H., Xu, Z., Hou, L., Yao, J., and Yuan, C.
\newblock Intactkv: Improving large language model quantization by keeping pivot tokens intact.
\newblock \emph{arXiv preprint arXiv:2403.01241}, 2024{\natexlab{c}}.

\bibitem[Liu et~al.(2024{\natexlab{d}})Liu, Desai, Liao, Wang, Xie, Xu, Kyrillidis, and Shrivastava]{Scissorhands}
Liu, Z., Desai, A., Liao, F., Wang, W., Xie, V., Xu, Z., Kyrillidis, A., and Shrivastava, A.
\newblock Scissorhands: Exploiting the persistence of importance hypothesis for llm kv cache compression at test time.
\newblock \emph{Advances in Neural Information Processing Systems}, 36, 2024{\natexlab{d}}.

\bibitem[Liu et~al.(2024{\natexlab{e}})Liu, Yuan, Jin, Zhong, Xu, Braverman, Chen, and Hu]{KIVI}
Liu, Z., Yuan, J., Jin, H., Zhong, S., Xu, Z., Braverman, V., Chen, B., and Hu, X.
\newblock Kivi: A tuning-free asymmetric 2bit quantization for kv cache.
\newblock \emph{arXiv preprint arXiv:2402.02750}, 2024{\natexlab{e}}.

\bibitem[Luohe et~al.(2024)Luohe, Hongyi, Yao, Zuchao, and Hai]{luohe2024keep}
Luohe, S., Hongyi, Z., Yao, Y., Zuchao, L., and Hai, Z.
\newblock Keep the cost down: A review on methods to optimize llm's kv-cache consumption.
\newblock \emph{arXiv preprint arXiv:2407.18003}, 2024.

\bibitem[OpenAI(2023)]{openai2023gpt4}
OpenAI.
\newblock Gpt-4 technical report.
\newblock \emph{arXiv preprint arXiv:2303.08774}, 2023.

\bibitem[Oren et~al.(2024)Oren, Hassid, Adi, and Schwartz]{TOVA}
Oren, M., Hassid, M., Adi, Y., and Schwartz, R.
\newblock Transformers are multi-state rnns.
\newblock \emph{arXiv preprint arXiv:2401.06104}, 2024.

\bibitem[Qiu et~al.(2024)Qiu, Mao, Patke, Cui, Jha, Wang, Franke, Kalbarczyk, Ba\c{s}ar, and Iyer]{qiu-length-predictor}
Qiu, H., Mao, W., Patke, A., Cui, S., Jha, S., Wang, C., Franke, H., Kalbarczyk, Z.~T., Ba\c{s}ar, T., and Iyer, R.~K.
\newblock Efficient interactive llm serving with proxy model-based sequence length prediction.
\newblock In \emph{The 5th International Workshop on Cloud Intelligence / AIOps at ASPLOS 2024}, volume~5, pp.\  1--7, San Diego, CA, USA, 2024. Association for Computing Machinery.

\bibitem[Ren \& Zhu(2024)Ren and Zhu]{ROCO}
Ren, S. and Zhu, K.~Q.
\newblock On the efficacy of eviction policy for key-value constrained generative language model inference.
\newblock \emph{arXiv preprint arXiv:2402.06262}, 2024.

\bibitem[Sheng et~al.(2023)Sheng, Zheng, Yuan, Li, Ryabinin, Chen, Liang, R{\'e}, Stoica, and Zhang]{sheng2023flexgen}
Sheng, Y., Zheng, L., Yuan, B., Li, Z., Ryabinin, M., Chen, B., Liang, P., R{\'e}, C., Stoica, I., and Zhang, C.
\newblock Flexgen: High-throughput generative inference of large language models with a single gpu.
\newblock In \emph{International Conference on Machine Learning}, pp.\  31094--31116. PMLR, 2023.

\bibitem[Shi et~al.(2024)Shi, Ming, Nguyen, Liang, and Joty]{GemFilter}
Shi, Z., Ming, Y., Nguyen, X.-P., Liang, Y., and Joty, S.
\newblock Discovering the gems in early layers: Accelerating long-context llms with 1000x input token reduction.
\newblock \emph{arXiv preprint arXiv:2409.17422}, 2024.

\bibitem[Tang et~al.(2024{\natexlab{a}})Tang, Lin, Lin, Han, Hong, Yao, and Wang]{RazorAttention}
Tang, H., Lin, Y., Lin, J., Han, Q., Hong, S., Yao, Y., and Wang, G.
\newblock Razorattention: Efficient kv cache compression through retrieval heads.
\newblock \emph{arXiv preprint arXiv:2407.15891}, 2024{\natexlab{a}}.

\bibitem[Tang et~al.(2024{\natexlab{b}})Tang, Zhao, Zhu, Xiao, Kasikci, and Han]{Quest}
Tang, J., Zhao, Y., Zhu, K., Xiao, G., Kasikci, B., and Han, S.
\newblock Quest: Query-aware sparsity for efficient long-context llm inference.
\newblock \emph{arXiv preprint arXiv:2406.10774}, 2024{\natexlab{b}}.

\bibitem[Touvron et~al.(2023)Touvron, Lavril, Izacard, Martinet, Lachaux, Lacroix, Rozi{\`e}re, Goyal, Hambro, Azhar, et~al.]{touvron2023llama}
Touvron, H., Lavril, T., Izacard, G., Martinet, X., Lachaux, M.-A., Lacroix, T., Rozi{\`e}re, B., Goyal, N., Hambro, E., Azhar, F., et~al.
\newblock Llama: Open and efficient foundation language models.
\newblock \emph{arXiv preprint arXiv:2302.13971}, 2023.

\bibitem[Wang \& Gan(2024)Wang and Gan]{SqueezeAttention}
Wang, Z. and Gan, S.
\newblock Squeezeattention: 2d management of kv-cache in llm inference via layer-wise optimal budget.
\newblock \emph{arXiv preprint arXiv:2404.04793}, 2024.

\bibitem[Wolf et~al.(2020)Wolf, Debut, Sanh, Chaumond, Delangue, Moi, Cistac, Rault, Louf, Funtowicz, Davison, Shleifer, von Platen, Ma, Jernite, Plu, Xu, Scao, Gugger, Drame, Lhoest, and Rush]{transformers-library}
Wolf, T., Debut, L., Sanh, V., Chaumond, J., Delangue, C., Moi, A., Cistac, P., Rault, T., Louf, R., Funtowicz, M., Davison, J., Shleifer, S., von Platen, P., Ma, C., Jernite, Y., Plu, J., Xu, C., Scao, T.~L., Gugger, S., Drame, M., Lhoest, Q., and Rush, A.~M.
\newblock Transformers: State-of-the-art natural language processing.
\newblock In \emph{Proceedings of the 2020 Conference on Empirical Methods in Natural Language Processing: System Demonstrations}, pp.\  38--45, Online, October 2020. Association for Computational Linguistics.
\newblock URL \url{https://www.aclweb.org/anthology/2020.emnlp-demos.6}.

\bibitem[Xiao et~al.(2024{\natexlab{a}})Xiao, Zhang, Han, Xiao, Lin, Zhang, Liu, and Sun]{InfLLM}
Xiao, C., Zhang, P., Han, X., Xiao, G., Lin, Y., Zhang, Z., Liu, Z., and Sun, M.
\newblock Infllm: Training-free long-context extrapolation for llms with an efficient context memory.
\newblock In \emph{First Workshop on Long-Context Foundation Models@ ICML 2024}, 2024{\natexlab{a}}.

\bibitem[Xiao et~al.(2023)Xiao, Tian, Chen, Han, and Lewis]{StreamingLLM}
Xiao, G., Tian, Y., Chen, B., Han, S., and Lewis, M.
\newblock Efficient streaming language models with attention sinks.
\newblock \emph{arXiv preprint arXiv:2309.17453}, 2023.

\bibitem[Xiao et~al.(2024{\natexlab{b}})Xiao, Tang, Zuo, Guo, Yang, Tang, Fu, and Han]{DuoAttention}
Xiao, G., Tang, J., Zuo, J., Guo, J., Yang, S., Tang, H., Fu, Y., and Han, S.
\newblock Duoattention: Efficient long-context llm inference with retrieval and streaming heads, 2024{\natexlab{b}}.
\newblock URL \url{https://arxiv.org/abs/2410.10819}.

\bibitem[Xu et~al.(2024)Xu, Jie, Dong, Wang, Lu, Zhou, Saha, Xiong, and Sahoo]{ThinK}
Xu, Y., Jie, Z., Dong, H., Wang, L., Lu, X., Zhou, A., Saha, A., Xiong, C., and Sahoo, D.
\newblock Think: Thinner key cache by query-driven pruning.
\newblock \emph{arXiv preprint arXiv:2407.21018}, 2024.

\bibitem[Yang et~al.(2024{\natexlab{a}})Yang, Han, Gao, Hu, Zhang, and Zhao]{PyramidInfer}
Yang, D., Han, X., Gao, Y., Hu, Y., Zhang, S., and Zhao, H.
\newblock Pyramidinfer: Pyramid kv cache compression for high-throughput llm inference.
\newblock \emph{arXiv preprint arXiv:2405.12532}, 2024{\natexlab{a}}.

\bibitem[Yang et~al.(2024{\natexlab{b}})Yang, Kim, Bae, Kwon, Park, Yang, Kwon, and Lee]{MiKV}
Yang, J.~Y., Kim, B., Bae, J., Kwon, B., Park, G., Yang, E., Kwon, S.~J., and Lee, D.
\newblock No token left behind: Reliable kv cache compression via importance-aware mixed precision quantization.
\newblock \emph{arXiv preprint arXiv:2402.18096}, 2024{\natexlab{b}}.

\bibitem[Yang et~al.(2024{\natexlab{c}})Yang, Sheng, Gonzalez, Stoica, and Zheng]{DoubleSparse}
Yang, S., Sheng, Y., Gonzalez, J.~E., Stoica, I., and Zheng, L.
\newblock Post-training sparse attention with double sparsity.
\newblock \emph{arXiv preprint arXiv:2408.07092}, 2024{\natexlab{c}}.

\bibitem[Yuan et~al.(2024)Yuan, Liu, Chuang, Li, Wang, Le, Jin, Chaudhary, Xu, Liu, et~al.]{yuan2024kv}
Yuan, J., Liu, H., Chuang, Y.-N., Li, S., Wang, G., Le, D., Jin, H., Chaudhary, V., Xu, Z., Liu, Z., et~al.
\newblock Kv cache compression, but what must we give in return? a comprehensive benchmark of long context capable approaches.
\newblock \emph{arXiv preprint arXiv:2407.01527}, 2024.

\bibitem[Yue et~al.(2024)Yue, Yuan, Duanmu, Zhou, Wu, and Nie]{WKVQuant}
Yue, Y., Yuan, Z., Duanmu, H., Zhou, S., Wu, J., and Nie, L.
\newblock Wkvquant: Quantizing weight and key/value cache for large language models gains more.
\newblock \emph{arXiv preprint arXiv:2402.12065}, 2024.

\bibitem[Zandieh et~al.(2024)Zandieh, Daliri, and Han]{QJL}
Zandieh, A., Daliri, M., and Han, I.
\newblock Qjl: 1-bit quantized jl transform for kv cache quantization with zero overhead.
\newblock \emph{arXiv preprint arXiv:2406.03482}, 2024.

\bibitem[Zhang et~al.(2024{\natexlab{a}})Zhang, Ji, Chen, Fu, Miao, Nie, Chen, and Cui]{PQCache}
Zhang, H., Ji, X., Chen, Y., Fu, F., Miao, X., Nie, X., Chen, W., and Cui, B.
\newblock Pqcache: Product quantization-based kvcache for long context llm inference.
\newblock \emph{arXiv preprint arXiv:2407.12820}, 2024{\natexlab{a}}.

\bibitem[Zhang et~al.(2024{\natexlab{b}})Zhang, Yi, Xu, and Shrivastava]{CoupledQuantization}
Zhang, T., Yi, J., Xu, Z., and Shrivastava, A.
\newblock Kv cache is 1 bit per channel: Efficient large language model inference with coupled quantization.
\newblock \emph{Advances in Neural Information Processing Systems}, 37:\penalty0 3304--3331, 2024{\natexlab{b}}.

\bibitem[Zhang et~al.(2024{\natexlab{c}})Zhang, Du, Luo, Zhong, Zhang, Liu, and Ji]{CaM}
Zhang, Y., Du, Y., Luo, G., Zhong, Y., Zhang, Z., Liu, S., and Ji, R.
\newblock Cam: Cache merging for memory-efficient llms inference.
\newblock In \emph{Forty-first International Conference on Machine Learning}, 2024{\natexlab{c}}.

\bibitem[Zhang et~al.(2024{\natexlab{d}})Zhang, Gao, Liu, Lu, Xiong, Dong, Chang, Hu, Xiao, et~al.]{PyramidKV}
Zhang, Y., Gao, B., Liu, T., Lu, K., Xiong, W., Dong, Y., Chang, B., Hu, J., Xiao, W., et~al.
\newblock Pyramidkv: Dynamic kv cache compression based on pyramidal information funneling.
\newblock \emph{arXiv preprint arXiv:2406.02069}, 2024{\natexlab{d}}.

\bibitem[Zhang \& Shen(2024)Zhang and Shen]{ZDC}
Zhang, Z. and Shen, H.
\newblock Zero-delay qkv compression for mitigating kv cache and network bottlenecks in llm inference.
\newblock \emph{arXiv preprint arXiv:2408.04107}, 2024.

\bibitem[Zhang et~al.(2024{\natexlab{e}})Zhang, Liu, Chen, Kailkhura, Chen, and Wang]{Q-Hitter}
Zhang, Z., Liu, S., Chen, R., Kailkhura, B., Chen, B., and Wang, A.
\newblock Q-hitter: A better token oracle for efficient llm inference via sparse-quantized kv cache.
\newblock \emph{Proceedings of Machine Learning and Systems}, 6:\penalty0 381--394, 2024{\natexlab{e}}.

\bibitem[Zhang et~al.(2024{\natexlab{f}})Zhang, Sheng, Zhou, Chen, Zheng, Cai, Song, Tian, R{\'e}, Barrett, et~al.]{H2O}
Zhang, Z., Sheng, Y., Zhou, T., Chen, T., Zheng, L., Cai, R., Song, Z., Tian, Y., R{\'e}, C., Barrett, C., et~al.
\newblock H2o: Heavy-hitter oracle for efficient generative inference of large language models.
\newblock \emph{Advances in Neural Information Processing Systems}, 36, 2024{\natexlab{f}}.

\bibitem[Zheng et~al.(2023)Zheng, Ren, Xue, Luo, Jiang, and You]{zheng-length-predictor}
Zheng, Z., Ren, X., Xue, F., Luo, Y., Jiang, X., and You, Y.
\newblock Response length perception and sequence scheduling: An llm-empowered llm inference pipeline.
\newblock \emph{arXiv preprint arXiv:2305.13144}, 2023.

\bibitem[Zhu et~al.(2024)Zhu, Duan, Chen, Liu, Li, Feng, Lv, Cao, Chuanfu, Zhang, Lin, and Yang]{SampleAttention}
Zhu, Q., Duan, J., Chen, C., Liu, S., Li, X., Feng, G., Lv, X., Cao, H., Chuanfu, X., Zhang, X., Lin, D., and Yang, C.
\newblock Sampleattention: Near-lossless acceleration of long context llm inference with adaptive structured sparse attention, 2024.
\newblock URL \url{https://arxiv.org/abs/2406.15486}.

\end{thebibliography}
\bibliographystyle{mlsys2025}

\newpage
\appendix

\section{Evaluation Details}
\label{app:evaluation-details}
\subsection{Dataset}
\noindent \textbf{ShareGPT.} We select a subset of requests from ShareGPT~\cite{sharegpt_vicuna_unfiltered} to conduct the experiments of response length difference distribution. We refer to the benchmark code in vLLM\footnote{\url{https://github.com/vllm-project/vllm/blob/main/benchmarks/benchmark_serving.py}} to sample 1, 000 requests. Due to time and resource constraints, we set the maximum number of generation tokens as 1024 in the evaluation. We also truncate contexts for input prompts that exceed the model's maximum length allowance to ensure they fit within the model's capacity.

\noindent\textbf{LongBench.} LongBench~\cite{bai2023longbench} is a task for long context understanding that covers key long-text application scenarios, including multi-document QA, single-document QA, summarization, few-shot learning, code completion, and synthetic tasks. We keep strictly the evaluation metrics and settings in their released codebase~\footnote{\url{https://github.com/THUDM/LongBench}} to ensure fair assessments. 

\subsection{Models}
\noindent\textbf{LLaMA Family.} The LLaMA family, developed by Meta using a high-quality corpus, is widely favored by researchers working on \texttt{KV cache} compression algorithms. Many choose LLaMA models to evaluate the effectiveness of their methods. In our performance evaluation, we cover LLaMA-2-7B, LLaMA-2-13B, and LLaMA-2-70B to emphasize the advantages of \texttt{KV} \texttt{cache} due to their exorbitant GPU memory consumption of \texttt{KV} \texttt{cache}. Additionally, LLaMA-3.1-8B, known for generating high-quality responses and excelling in long-context tasks, is used in our length distribution and negative sample analysis.

\noindent\textbf{Mistral Family.} Similarly to the LLaMA family, many models from the Mistral family are used to demonstrate the benefits of \texttt{KV} \texttt{cache} algorithms. Mistral models incorporate grouped-query attention (GQA) for faster inference and are renowned for their exceptional performance. In our length difference and negative sample analysis, we utilize Mistral-7B-v0.1 to obtain relevant experimental results.

\subsection{Algorithms}
\label{app:eval-alg-details}
\noindent\textbf{KIVI.} KIVI~\cite{KIVI} is a notable quantization algorithm for \texttt{KV} \texttt{cache} compression, specializing in per-channel quantization for key tensors and per-token quantization for value tensors. We utilize their official implementation~\footnote{\url{https://github.com/jy-yuan/KIVI}}. The critical hyperparameters in KIVI are group size $G$ and the residual length $R$. $G$ refers to the number of channels that are grouped for quantization in the key cache, while $R$ controls the number of most recent tokens that are kept in full precision. Following the paper's recommendations for achieving optimal performance, we have set them to $G=32, R=128$.

\noindent\textbf{GEAR.} GEAR~\cite{GEAR} is a typical quantization error mitigation algorithm. We took their open-source code\footnote{\url{https://github.com/opengear-project/GEAR}}. The key parameters of GEAR are sparsity ratio $s$ and rank $r$. Specifically, $s$ specifies the number of retained full-precision outlier values. $r$ controls the richness of the low-rank approximation matrix, which recovers the model's ability from quantization errors. In line with the default settings in the official codebase, we set $s=2\%, r=2\%$.

\noindent \textbf{StreamingLLM.} StreamingLLM~\cite{StreamingLLM} is an attention sparsity-based cache eviction algorithm. It retains only a limited number of initial and most recent tokens. The key parameters for controlling the sizes of the initial and recent tokens are set to 64 and 448, respectively, resulting in a total cache size of 512.

\noindent \textbf{H2O.} H2O~\cite{H2O} is another widely-used cache eviction algorithm that dynamically calculates and refreshes the \texttt{KV} \texttt{cache}. The parameters for the heavy hitter oracle token size and the recent size are configured to 64 and 448, respectively, with a total cache size of 512.

\subsection{LLM Serving Engine.}
\noindent\textbf{Transformers Library.} We directly use Torch 2.1.2 and Transformers 4.43.1 to measure the throughput performance. 

\noindent\textbf{FlashAttention.} FlashAttention\footnote{\url{https://github.com/Dao-AILab/flash-attention}} can fully exploit the GPU resources to realize fast and memory-efficient attention operation. In our throughput evaluation, we measure the throughput performance of TRL+FA by enabling FlashAttention 2.5.6 in the transformers library. 

\noindent\textbf{LMDeploy.} LMDeploy\footnote{\url{https://github.com/InternLM/lmdeploy/tree/main}} allows LLM developers to compress, deploy, and serve various LLMs. It naturally supports the functionality of PagedAttention and FlashAttention. We implement various \texttt{KV} \texttt{cache} algorithms based on LMDeploy v6.0.1. We chose LMDeploy for three reasons.

First, LMDeploy stands out by implementing more efficient quantization kernels than vLLM. This results in superior performance in KV cache compression approaches compared to vLLM, despite the significant attention vLLM has garnered. Note that the primary focus of our paper is on KV cache compression approaches, with a particular emphasis on quantization. A prior benchmark study conducted by BentoML~\cite{bentoml} uncovers that LMDeploy obtains the best throughput performance with 4-bit quantization.

Second, LMDeploy offers a better way for faster development of KV cache compression algorithms than vLLM. The author of KVI has stated the challenges of integrating the KIVI algorithm into vLLM as early as April 2024~\cite{kivi-issue-4}. As of now, there has been no significant progress on this front.

Third, our conclusions, except for Observation 2, do not pertain to any specific serving features of the inference engines. Consequently, they remain independent of the LLM inference engine used. For Observation 2, our objective is to explore the impact of KV cache compression methods like sparsity and quantization on popular serving features (e.g., Page Attention, Flash Attention) rather than focusing on any particular serving engine. As long as the selected inference engines support the efficient implementation of the necessary serving features (Page Attention, FlashAttention), it will not affect Observation 2.

\subsection{Hardware Environment.}
Our evaluation experiments are conducted on a GPU node with four NVIDIA A6000 GPUs interconnected via NVLink and powered by an Intel Xeon Gold 6326 CPU at 2.90 GHz.

\begin{figure*}[!t]
    \centering
    \includegraphics[width=.32\textwidth]{Figs/frw/frw_legend.pdf}

    \subfigure[Decode, KV Length 256]{
        \includegraphics[width=0.22\textwidth]{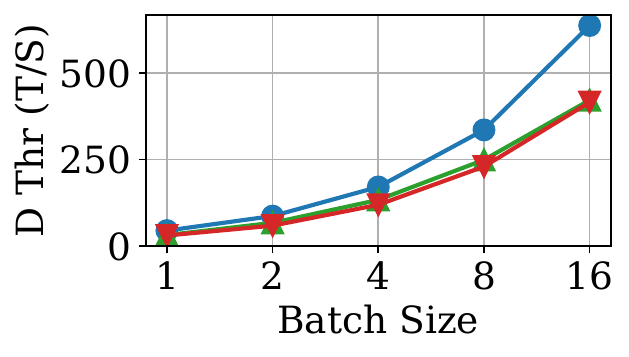}
        \label{fig:frw_thr_qaunt_16_policy_None_phase_decoding_promptlen_256_M7}
    }
    \subfigure[Decode, KV Length 2048]{
        \includegraphics[width=0.22\textwidth]{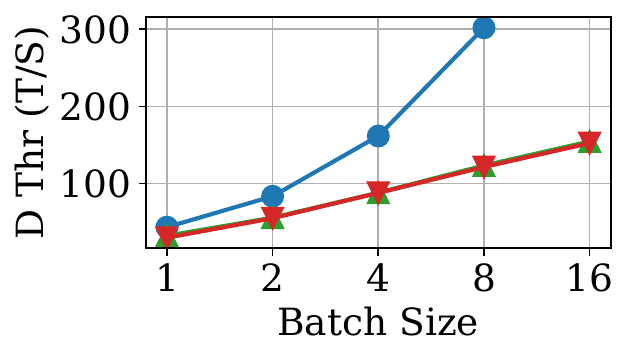}
        \label{fig:frw_thr_qaunt_16_policy_None_phase_decoding_promptlen_2048_M7}
    }
    \subfigure[Decode, KV Length 256]{
        \includegraphics[width=0.22\textwidth]{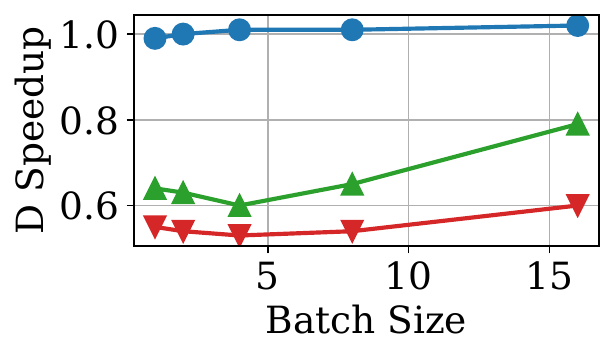}
        \label{fig:frw_speedup_4_KIVI_normal_phase_decoding_promptlen_256_M7}
    }
    \subfigure[Decode, KV Length 2048]{
        \includegraphics[width=0.22\textwidth]{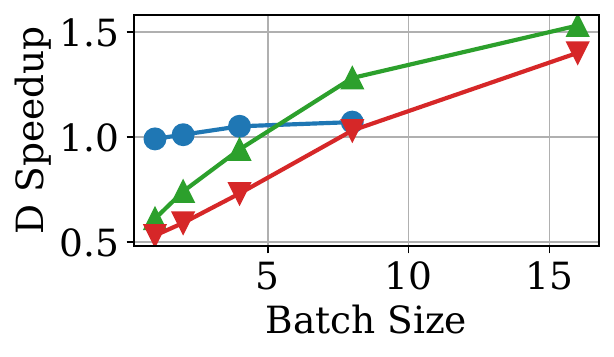}
        \label{fig:frw_speedup_4_KIVI_normal_phase_decoding_promptlen_2048_M7}
    }
    
    \centering
    \includegraphics[width=.5\textwidth]{Figs/lmd/thr_legend.pdf}

    \subfigure[Prefill, Prompt 1024]{
        \includegraphics[width=0.22\textwidth]{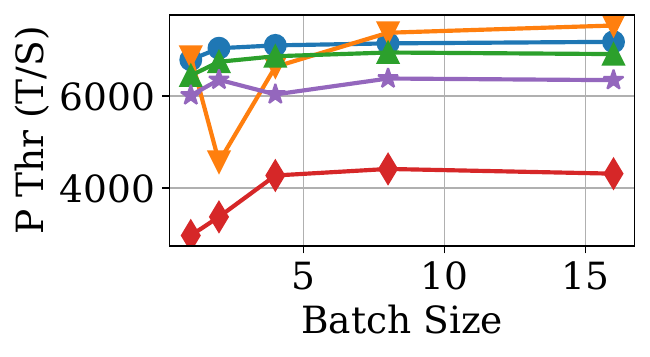}
        \label{fig:normal_phase_prefill_promptlen_1024_M7}
    }
    \subfigure[Prefill, Batch 1]{
        \includegraphics[width=0.22\textwidth]{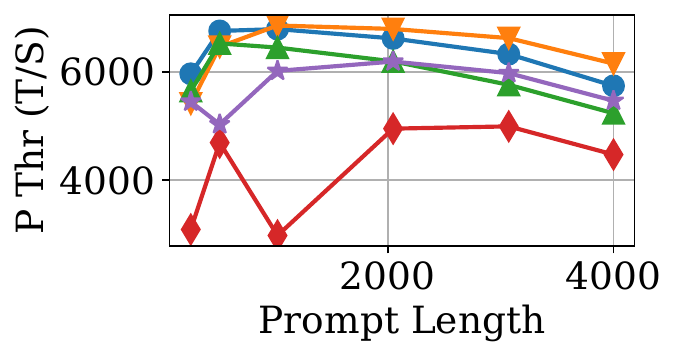}
        \label{fig:normal_phase_prefill_bsz_1_M7}
    }
    \subfigure[Decode, KV Length 1024]{
        \includegraphics[width=0.22\textwidth]{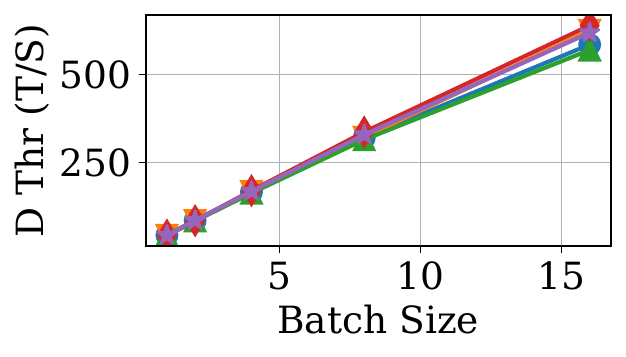}
        \label{fig:normal_phase_decoding_promptlen_1024_M7}
    }
    \subfigure[Decode, Batch 1]{
        \includegraphics[width=0.22\textwidth]{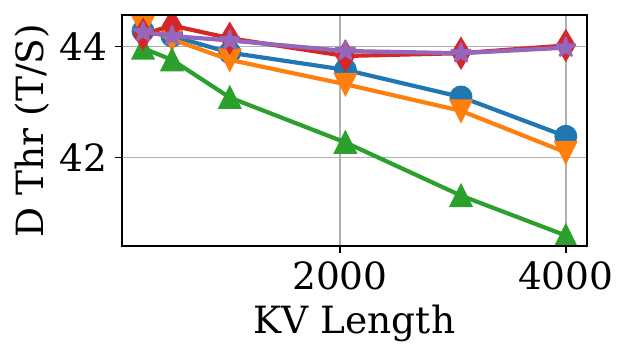}
        \label{fig:normal_phase_decoding_bsz_1_M7}
    }
\caption{
Throughput analysis of \textbf{Mistral-7B} (a-b) The FP16 decoding throughput on TRL (with and without FlashAttention) and LMDeploy (LMD). (c-d) The speedup of the KIVI-4bit algorithm on TRL and LMD. (e-h) The prefill and decoding throughput for inputs of moderate size.
}
\label{fig:thr-analysis-M7}
\end{figure*}

\begin{figure*}[!h]    
    \centering
    \includegraphics[width=.55\textwidth]{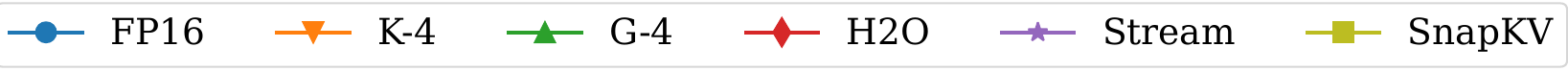}

    \subfigure[Prefill, Prompt 1024]{
        \includegraphics[width=0.22\textwidth]{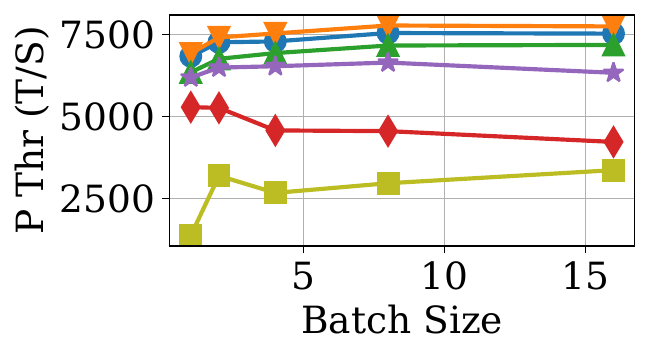}
        \label{fig:snapkv_normal_phase_prefill_promptlen_1024}
    }
    \subfigure[Prefill, Batch 1]{
        \includegraphics[width=0.22\textwidth]{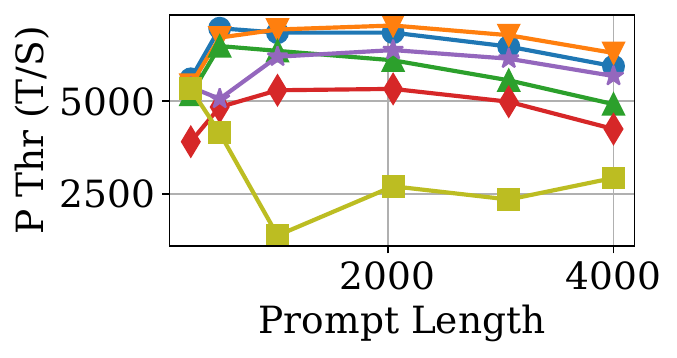}
        \label{fig:snapkv_normal_phase_prefill_bsz_1}
    }
    \subfigure[Decode, KV Length 1024]{
        \includegraphics[width=0.22\textwidth]{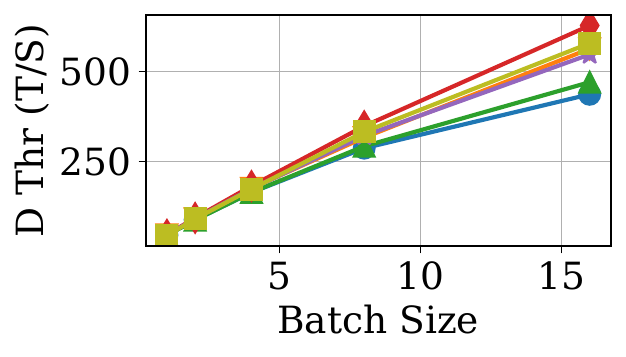}
        \label{fig:snapkv_normal_phase_decoding_promptlen_1024}
    }
    \subfigure[Decode, Batch 1]{
        \includegraphics[width=0.22\textwidth]{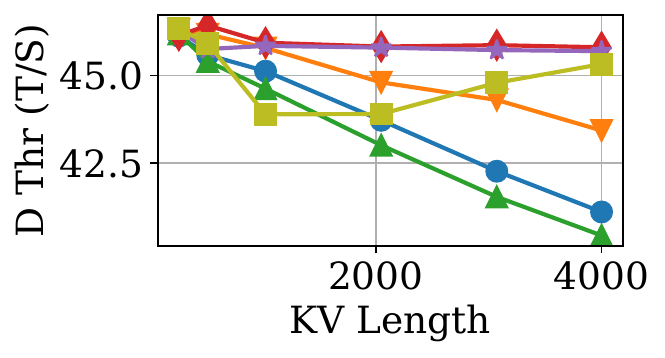}
        \label{fig:snapkv_normal_phase_decoding_bsz_1}
    }
\vspace{-10pt}
\caption{Throughput analysis of \textbf{LLaMA-7B}, with \texttt{KV cache} compression algorithm SnapKV~\cite{SnapKV} integrated.}
\label{fig:thr-analysis-snapkv}
\vspace{-10pt}
\end{figure*}

\begin{figure*}[!h]
    \centering
    \includegraphics[width=.4\textwidth]{Figs/frw/frw_legend.pdf}

    \subfigure[Decode, KV Length 256]{
        \includegraphics[width=0.22\textwidth]{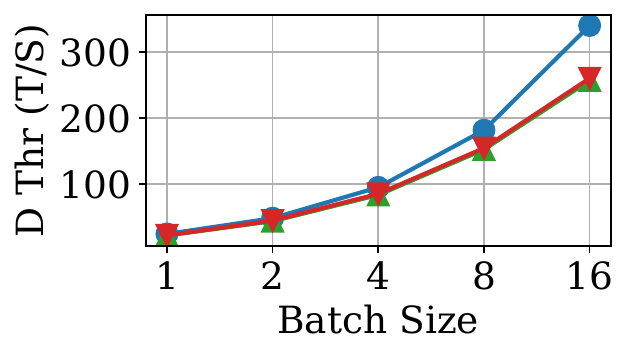}\label{fig:frw_thr_qaunt_16_policy_None_phase_decoding_promptlen_256_L13}
    }
    \subfigure[Decode, KV Length 2048]{
        \includegraphics[width=0.22\textwidth]{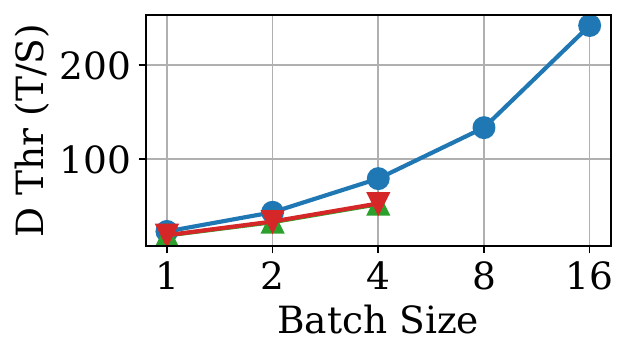}\label{fig:frw_thr_qaunt_16_policy_None_phase_decoding_promptlen_2048_L13}
    }
    \subfigure[Decode, KV Length 1024]{
        \includegraphics[width=0.22\textwidth]{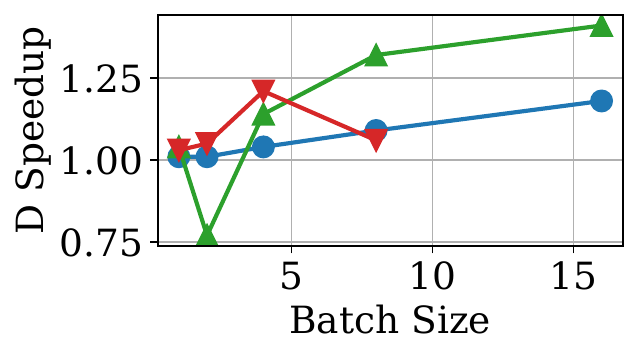}
        \label{fig:frw_speedup_16_StreamingLLM_normal_phase_decoding_promptlen_1024_L13}
    }
    \subfigure[Decode, KV Length 2048]{
        \includegraphics[width=0.22\textwidth]{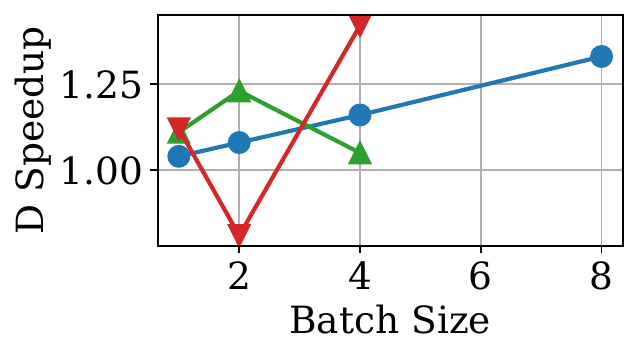}
        \label{fig:frw_speedup_16_StreamingLLM_normal_phase_decoding_promptlen_2048_L13}
    }
    
    \centering
    \includegraphics[width=.6\textwidth]{Figs/lmd/thr_legend.pdf}

    \centering
    \subfigure[Prefill, Prompt 1024]{
        \includegraphics[width=0.22\textwidth]{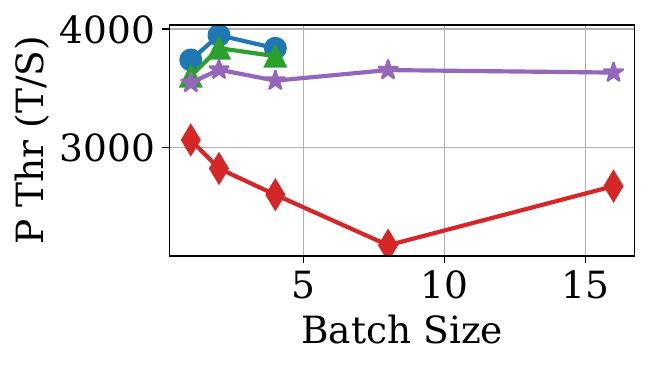}
        \label{fig:normal_phase_prefill_promptlen_1024_L13}
    }
    \subfigure[Prefill, Batch 1]{
        \includegraphics[width=0.22\textwidth]{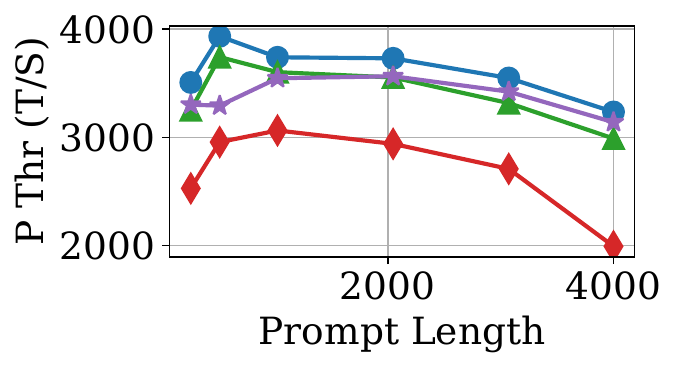}
        \label{fig:normal_phase_prefill_bsz_1_L13}
    }
    \subfigure[Prefill, Prompt 6144]{
        \includegraphics[width=0.22\textwidth]{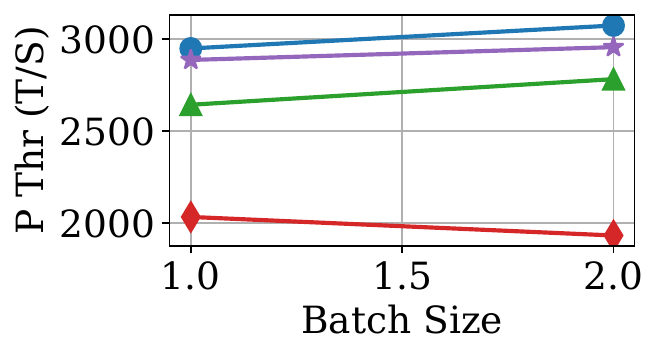}
        \label{fig:long_phase_prefill_promptlen_6144_L13}
    }
    \subfigure[Prefill, Batch 1]{
        \includegraphics[width=0.22\textwidth]{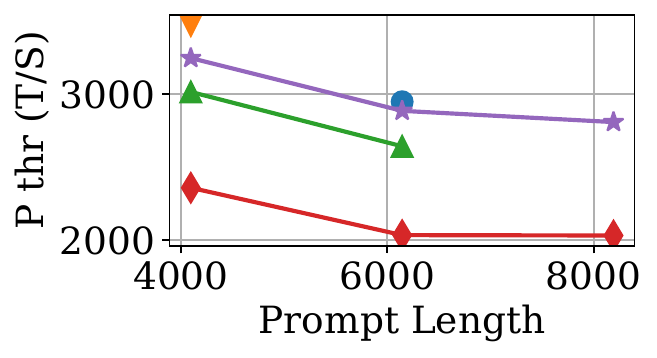}
        \label{fig:long_phase_prefill_bsz_1_L13}
    }

    \centering
    \subfigure[Decode, KV Length 1024]{
        \includegraphics[width=0.22\textwidth]{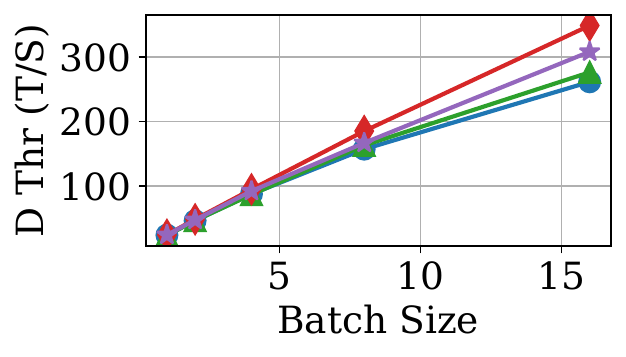}
        \label{fig:normal_phase_decoding_promptlen_1024_L13}
    }
    \subfigure[Decode, Batch 1]{
        \includegraphics[width=0.22\textwidth]{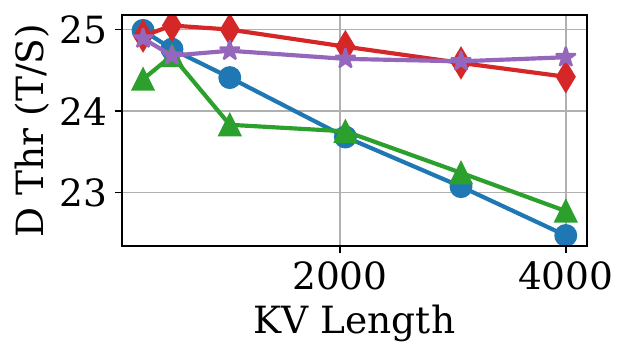}
        \label{fig:normal_phase_decoding_bsz_1_L13}
    }
    \subfigure[Decode, KV Length 6144]{
        \includegraphics[width=0.22\textwidth]{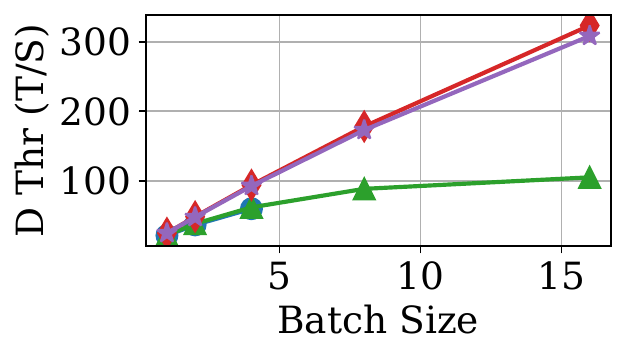}
        \label{fig:long_phase_decoding_promptlen_6144_L13}
    }
    \subfigure[Decode, Batch 1]{
        \includegraphics[width=0.22\textwidth]{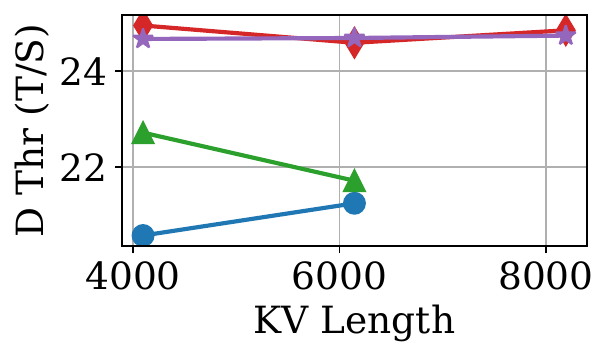}
        \label{fig:long_phase_decoding_bsz_1_L13}
    }
\caption{
Throughput analysis of \textbf{LLaMA-13B}: (a-b) The FP16 decoding throughput on TRL (with and without FlashAttention) and LMDeploy (LMD). (c-d) The speedup of the StreamingLLM algorithm on TRL and LMD. (e-h) The prefill throughput for various sizes of inputs. (i-l) The decoding throughput for various sizes of inputs.
}
\label{fig:thr-analysis-L13}
\vspace{-10pt}
\end{figure*}

\begin{figure*}[!h]    
    \centering
    \includegraphics[width=.32\textwidth]{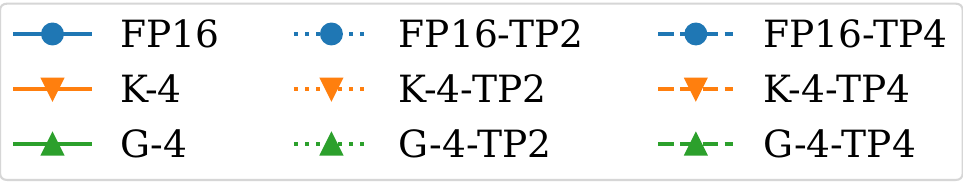}

    \subfigure[Prefill, Prompt 1024]{
        \includegraphics[width=0.22\textwidth]{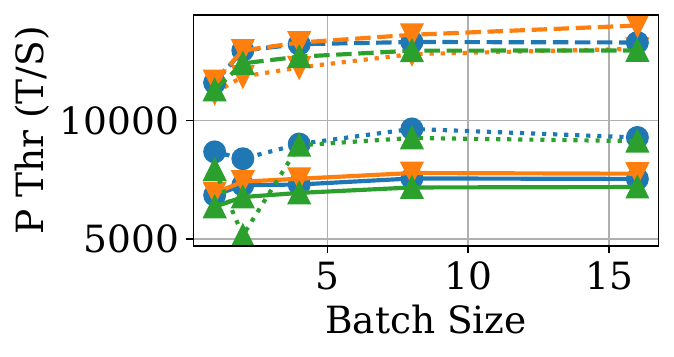}
        \label{fig:normal_phase_prefill_promptlen_1024_quanttp}
    }
    \subfigure[Prefill, Batch 1]{
        \includegraphics[width=0.22\textwidth]{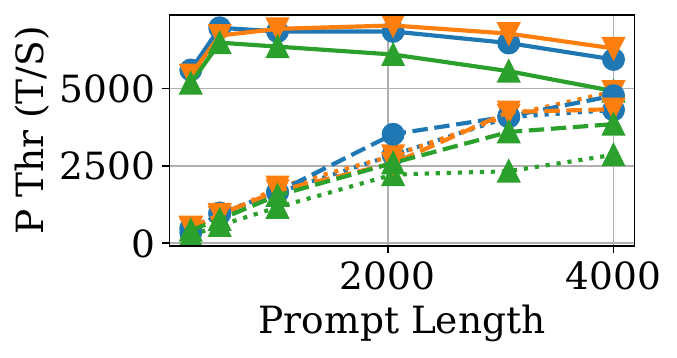}
        \label{fig:normal_phase_prefill_bsz_1_quanttp}
    }
    \subfigure[Decode, KV Length 1024]{
        \includegraphics[width=0.22\textwidth]{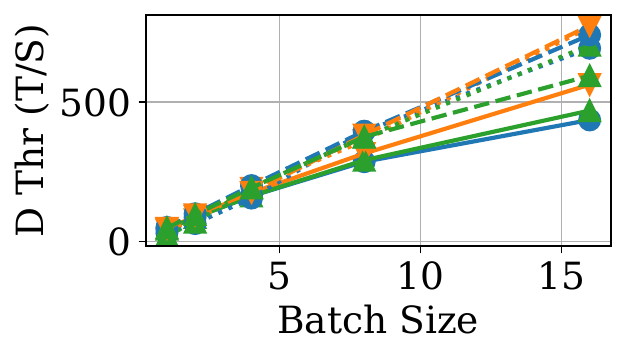}
        \label{fig:normal_phase_decoding_promptlen_1024_quanttp}
    }
    \subfigure[Decode, Batch 1]{
        \includegraphics[width=0.22\textwidth]{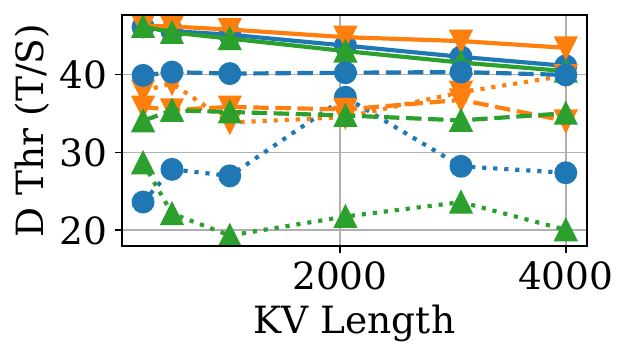}
        \label{fig:normal_phase_decoding_bsz_1_quanttp}
    }

    \centering
    \includegraphics[width=.35\textwidth]{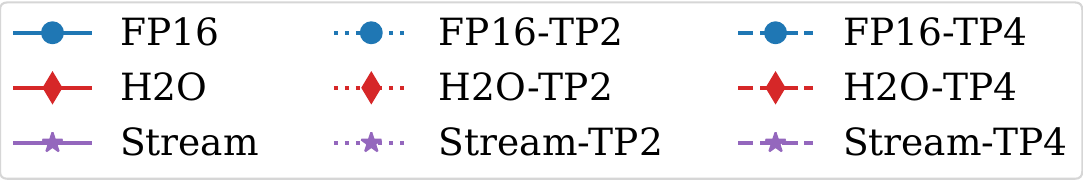}

    \subfigure[Prefill, Prompt 1024]{
        \includegraphics[width=0.22\textwidth]{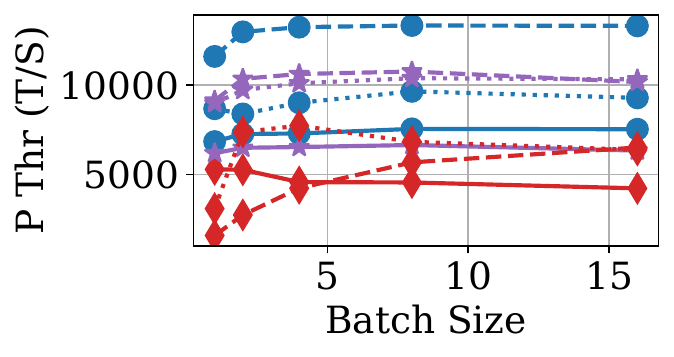}
        \label{fig:normal_phase_prefill_promptlen_1024_sparsetp}
    }
    \subfigure[Prefill, Batch 1]{
        \includegraphics[width=0.22\textwidth]{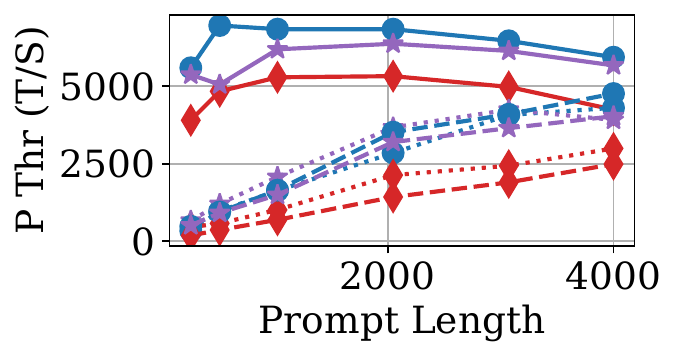}
        \label{fig:normal_phase_prefill_bsz_1_sparsetp}
    }
    \subfigure[Decode, KV Length 1024]{
        \includegraphics[width=0.22\textwidth]{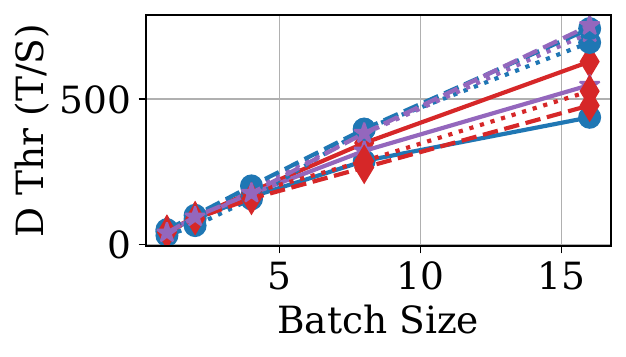}
        \label{fig:normal_phase_decoding_promptlen_1024_sparsetp}
    }
    \subfigure[Decode, Batch 1]{
        \includegraphics[width=0.22\textwidth]{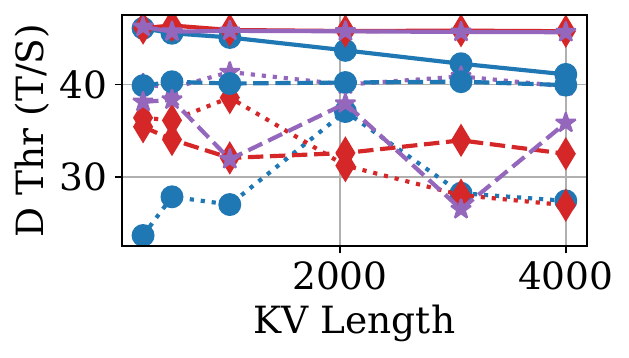}
        \label{fig:normal_phase_decoding_bsz_1_sparsetp}
    }
\caption{
Throughput analysis of \textbf{LLaMA-7B}, with different tensor parallelism configurations. (a-d) The throughput of quantization-based methods. (e-h) The throughput of sparsity-based methods.
}
\label{fig:tp-analysis}
\end{figure*}

\begin{figure*}[!h]    
    \centering
    \includegraphics[width=.32\textwidth]{Figs/lmd/quanttp_legend.pdf}

    \subfigure[Prefill, Prompt 1024]{
        \includegraphics[width=0.22\textwidth]{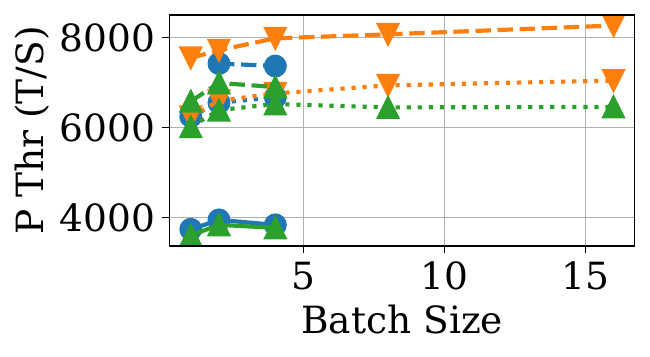}
        \label{fig:normal_phase_prefill_promptlen_1024_quanttp_L13}
    }
    \subfigure[Prefill, Batch 1]{
        \includegraphics[width=0.22\textwidth]{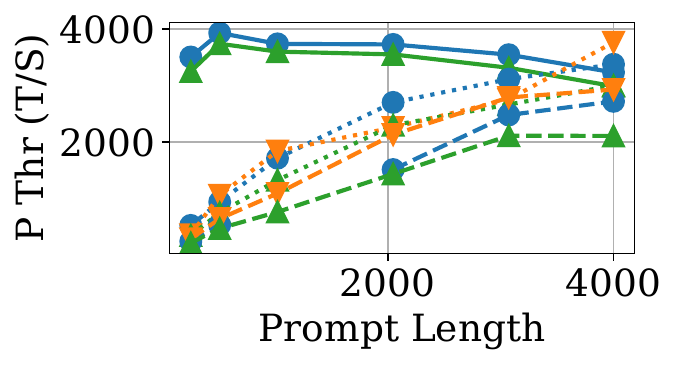}
        \label{fig:normal_phase_prefill_bsz_1_quanttp_L13}
    }
    \subfigure[Decode, KV Length 1024]{
        \includegraphics[width=0.22\textwidth]{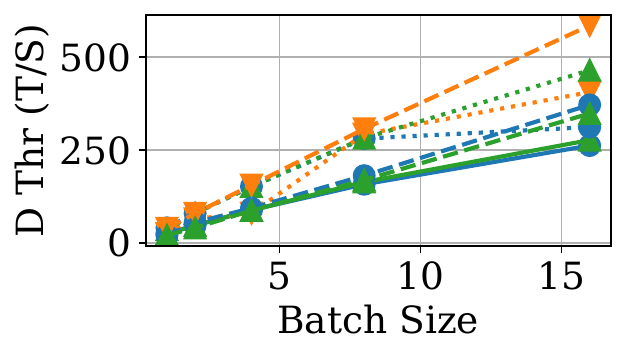}
        \label{fig:normal_phase_decoding_promptlen_1024_quanttp_L13}
    }
    \subfigure[Decode, Batch 1]{
        \includegraphics[width=0.22\textwidth]{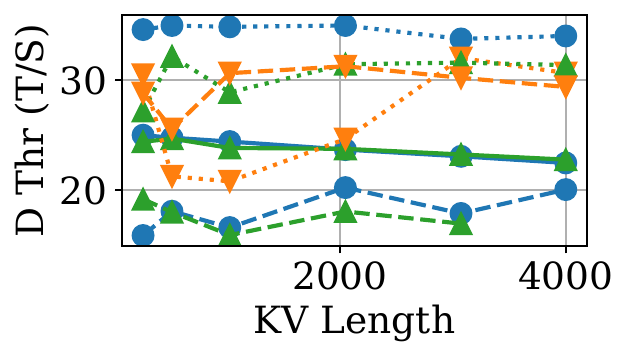}
        \label{fig:normal_phase_decoding_bsz_1_quanttp_L13}
    }

    \centering
    \includegraphics[width=.35\textwidth]{Figs/lmd/sparsetp_legend.pdf}

    \subfigure[Prefill, Prompt 1024]{
        \includegraphics[width=0.22\textwidth]{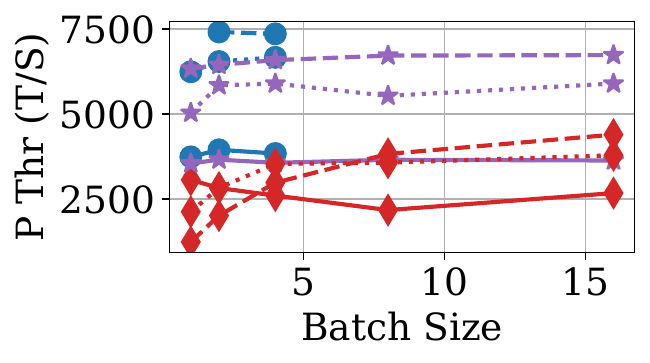}
        \label{fig:normal_phase_prefill_promptlen_1024_sparsetp_L13}
    }
    \subfigure[Prefill, Batch 1]{
        \includegraphics[width=0.22\textwidth]{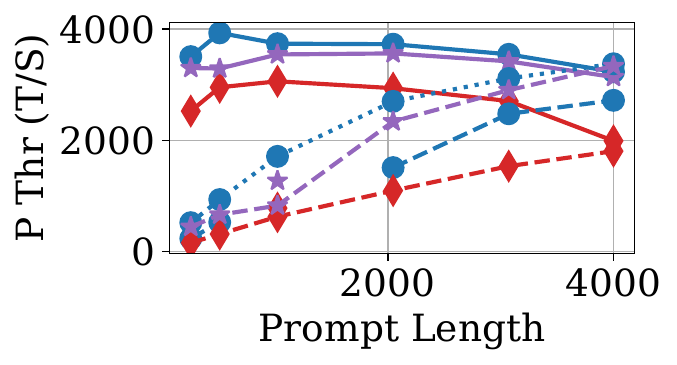}
        \label{fig:normal_phase_prefill_bsz_1_sparsetp_L13}
    }
    \subfigure[Decode, KV Length 1024]{
        \includegraphics[width=0.22\textwidth]{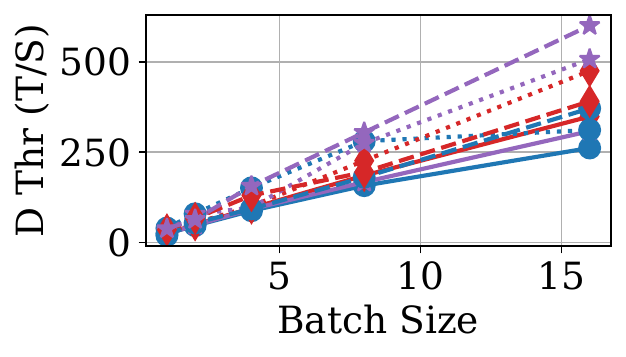}
        \label{fig:normal_phase_decoding_promptlen_1024_sparsetp_L13}
    }
    \subfigure[Decode, Batch 1]{
        \includegraphics[width=0.22\textwidth]{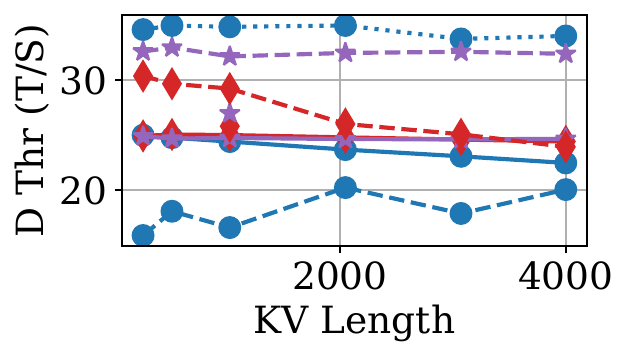}
        \label{fig:normal_phase_decoding_bsz_1_sparsetp_L13}
    }
\caption{
Tensor parallelism analysis of \textbf{LLaMA-13B}. (a-d) The throughput of quantization-based methods. (e-h) The throughput of sparsity-based methods.
}
\label{fig:tp-analysis-L13}
\vspace{-10pt}
\end{figure*}

\begin{figure*}[!h]    
    \centering
    \includegraphics[width=.32\textwidth]{Figs/lmd/quanttp_legend.pdf}

    \subfigure[Prefill, Prompt 1024]{
        \includegraphics[width=0.22\textwidth]{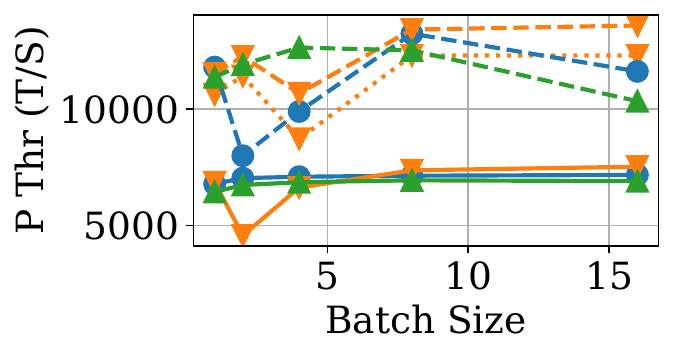}
        \label{fig:normal_phase_prefill_promptlen_1024_quanttp_M7}
    }
    \subfigure[Prefill, Batch 1]{
        \includegraphics[width=0.22\textwidth]{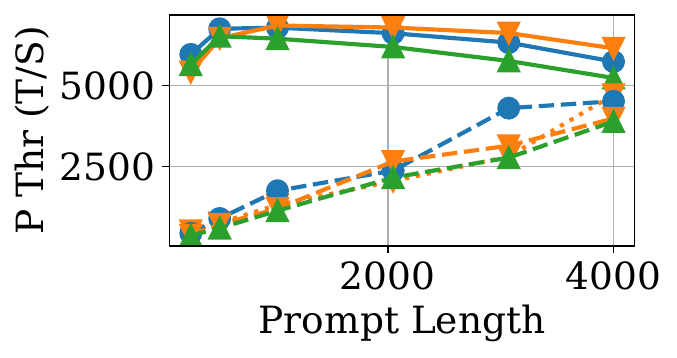}
        \label{fig:normal_phase_prefill_bsz_1_quanttp_M7}
    }
    \subfigure[Decode, KV Length 1024]{
        \includegraphics[width=0.22\textwidth]{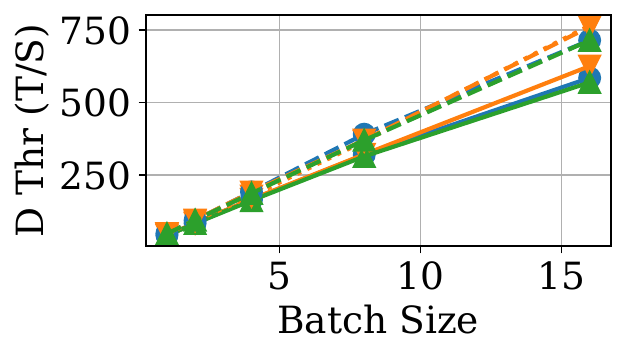}
        \label{fig:normal_phase_decoding_promptlen_1024_quanttp_M7}
    }
    \subfigure[Decode, Batch 1]{
        \includegraphics[width=0.22\textwidth]{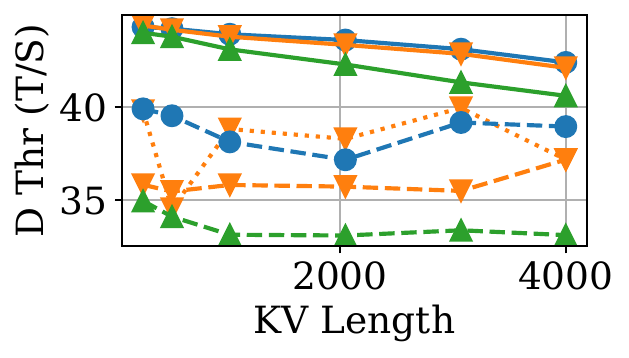}
        \label{fig:normal_phase_decoding_bsz_1_quanttp_M7}
    }

    \centering
    \includegraphics[width=.35\textwidth]{Figs/lmd/sparsetp_legend.pdf}

    \subfigure[Prefill, Prompt 1024]{
        \includegraphics[width=0.22\textwidth]{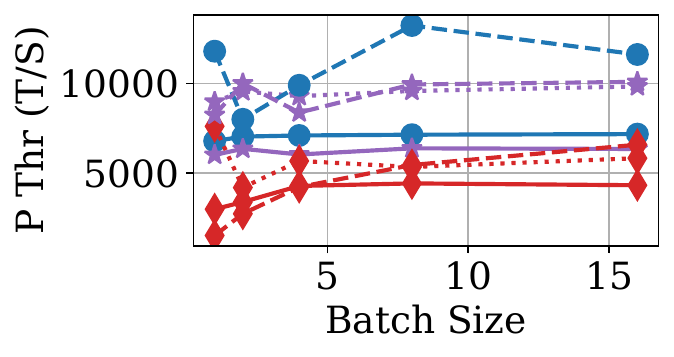}
        \label{fig:normal_phase_prefill_promptlen_1024_sparsetp_M7}
    }
    \subfigure[Prefill, Batch 1]{
        \includegraphics[width=0.22\textwidth]{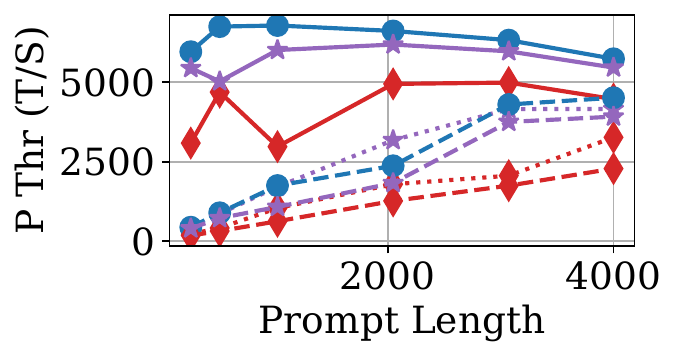}
        \label{fig:normal_phase_prefill_bsz_1_sparsetp_M7}
    }
    \subfigure[Decode, KV Length 1024]{
        \includegraphics[width=0.22\textwidth]{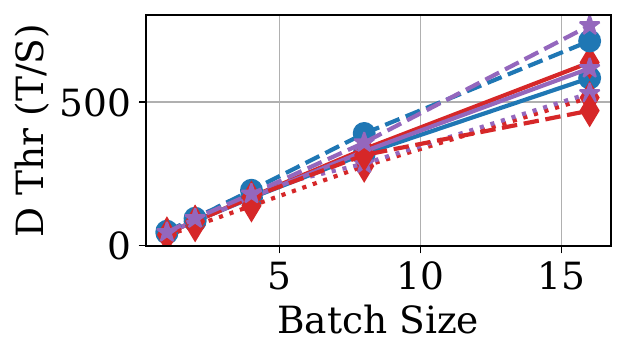}
        \label{fig:normal_phase_decoding_promptlen_1024_sparsetp_M7}
    }
    \subfigure[Decode, Batch 1]{
        \includegraphics[width=0.22\textwidth]{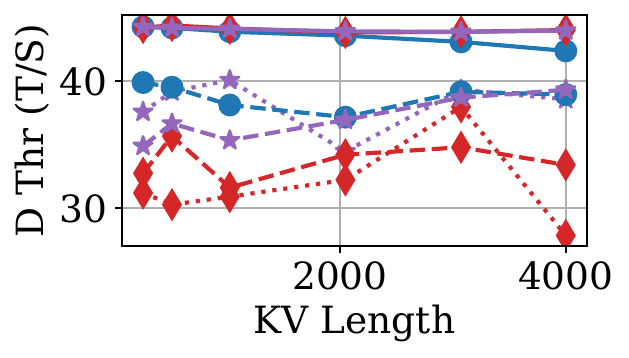}
        \label{fig:normal_phase_decoding_bsz_1_sparsetp_M7}
    }
\caption{
Tensor parallelism analysis of \textbf{Mistral-7B}. (a-d) The throughput of quantization-based methods. (e-h) The throughput of sparsity-based methods.
}
\label{fig:tp-analysis-M7}
\vspace{-10pt}
\end{figure*}

\begin{figure*}[!h]    
    \centering
    \includegraphics[width=.32\textwidth]{Figs/lmd/quanttp_legend.pdf}

    \subfigure[Prefill, Prompt 1024]{
        \includegraphics[width=0.22\textwidth]{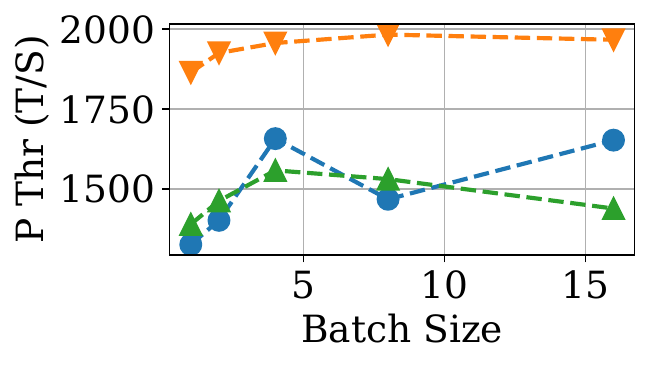}
        \label{fig:normal_phase_prefill_promptlen_1024_quanttp_L70}
    }
    \subfigure[Prefill, Batch 1]{
        \includegraphics[width=0.22\textwidth]{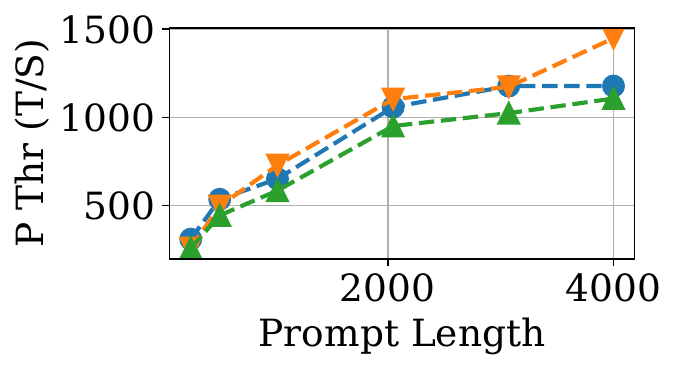}
        \label{fig:normal_phase_prefill_bsz_1_quanttp_L70}
    }
    \subfigure[Decode, KV Length 1024]{
        \includegraphics[width=0.22\textwidth]{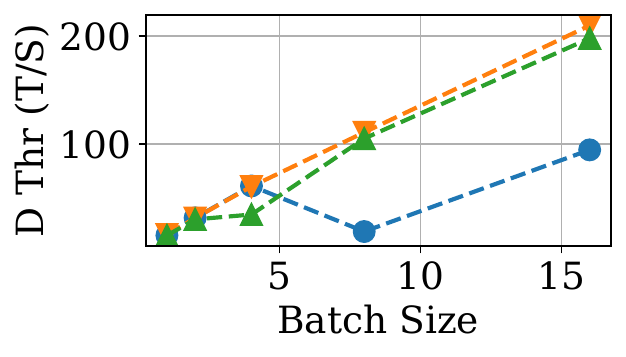}
        \label{fig:normal_phase_decoding_promptlen_1024_quanttp_L70}
    }
    \subfigure[Decode, Batch 1]{
        \includegraphics[width=0.22\textwidth]{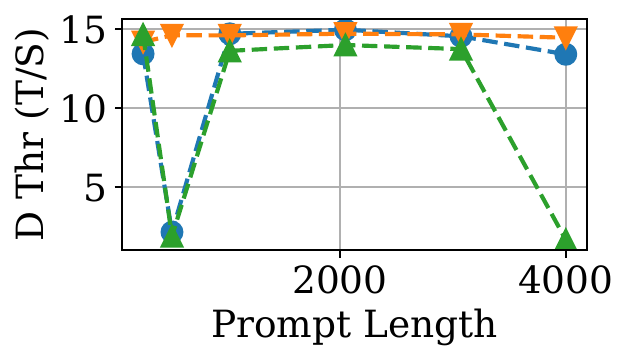}
        \label{fig:normal_phase_decoding_bsz_1_quanttp_L70}
    }

    \centering
    \includegraphics[width=.35\textwidth]{Figs/lmd/sparsetp_legend.pdf}

    \subfigure[Prefill, Prompt 1024]{
        \includegraphics[width=0.22\textwidth]{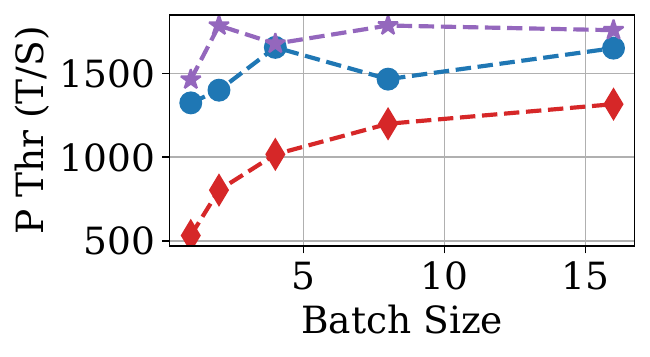}
        \label{fig:normal_phase_prefill_promptlen_1024_sparsetp_L70}
    }
    \subfigure[Prefill, Batch 1]{
        \includegraphics[width=0.22\textwidth]{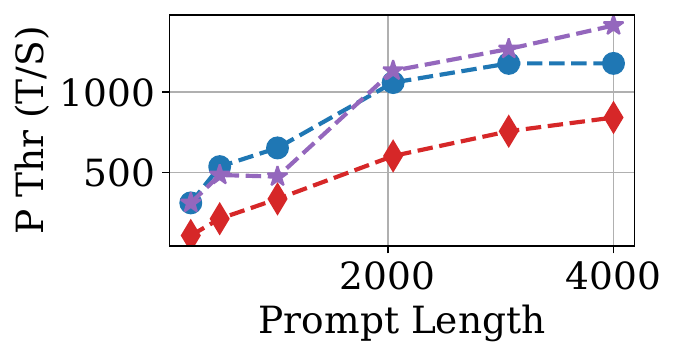}
        \label{fig:normal_phase_prefill_bsz_1_sparsetp_L70}
    }
    \subfigure[Decode, KV Length 1024]{
        \includegraphics[width=0.22\textwidth]{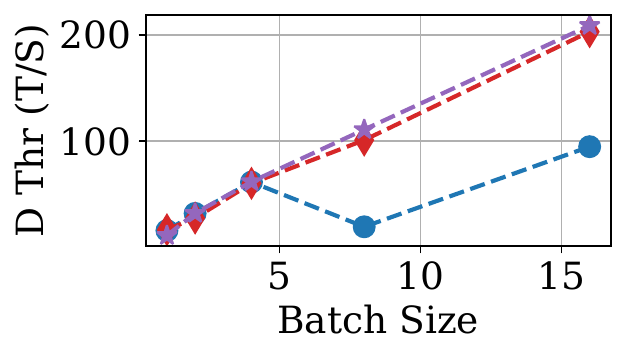}
        \label{fig:normal_phase_decoding_promptlen_1024_sparsetp_L70}
    }
    \subfigure[Decode, Batch 1]{
        \includegraphics[width=0.22\textwidth]{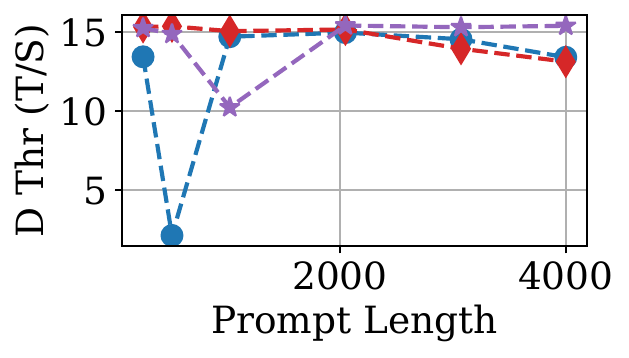}
        \label{fig:normal_phase_decoding_bsz_1_sparsetp_L70}
    }
\caption{
Tensor parallelism analysis of \textbf{LLaMA-70B}. (a-d) The throughput of quantization-based methods. (e-h) The throughput of sparsity-based methods.
}
\label{fig:tp-analysis-L70}
\vspace{-10pt}
\end{figure*}

\section{More results of Throughput Analysis}
\label{app:experimental-throughput-analysis}
We evaluate throughput performance on a GPU node with four NVIDIA A6000 GPUs interconnected via NVLink and powered by an Intel Xeon Gold 6326 CPU at 2.90 GHz. We exclude the initialization overhead and average the throughput performance over three times for fair comparison. We add more experiments to demonstrate the generality of our statement in the throughput analysis as follows.

First, we measure the prefill and decoding throughput on TRL, TRL+FA, and LMD using Mistral-7B and LLaMA-13B, depicted in Figure~\ref{fig:thr-analysis-M7} (a-b) and Figure~\ref{fig:thr-analysis-L13} (a-b), respectively. The relative speedup of the StreamingLLM algorithm in the decoding throughput varies across LLMs and serving techniques, as shown in Figure~\ref{fig:thr-analysis-M7} (c-d) and Figure~\ref{fig:thr-analysis-L13} (c-d).  The high speedup from TRL does not mean the significant speedup benefits.

Second, we measure the prefill and decoding throughput on LMD with various batch and prompt lengths in Figure~\ref{fig:thr-analysis-M7} (e-h) and Figure~\ref{fig:thr-analysis-L13} (e-l). We have observed that these Large Language Models (LLMs) show negative speedup in certain prompt lengths and batch sizes, which is consistent with the statement mentioned in Section~\ref{sec:evaluation}. However, the prompt lengths and batch sizes that lead to this disadvantage vary among different LLMs. Worth noticing that in Figure~\ref{fig:thr-analysis-L13}, we omit the throughput information for the KIVI-4 algorithm due to the out-of-memory issue when evaluating on LLaMA-13B with a single A6000 GPU.


Third, we present additional findings on tensor parallelism in Figures~\ref{fig:tp-analysis},~\ref{fig:tp-analysis-L13},~\ref{fig:tp-analysis-M7}, and~\ref{fig:tp-analysis-L70}. Performance improvements with larger tensor parallelism (TP) are notably evident during the prefill stage for various compression methods. However, larger TP does not confer significant benefits in the decoding stage when the batch size is small. Our observations indicate that the throughput advantages derived from \texttt{KV} \texttt{cache} compression typically become more pronounced under heavy evaluation settings (e.g., batch size, KV length, and model size).


\begin{figure*}[htbp]
    \centering
    \subfigure[KIVI]{
        \includegraphics[width=0.22\textwidth]{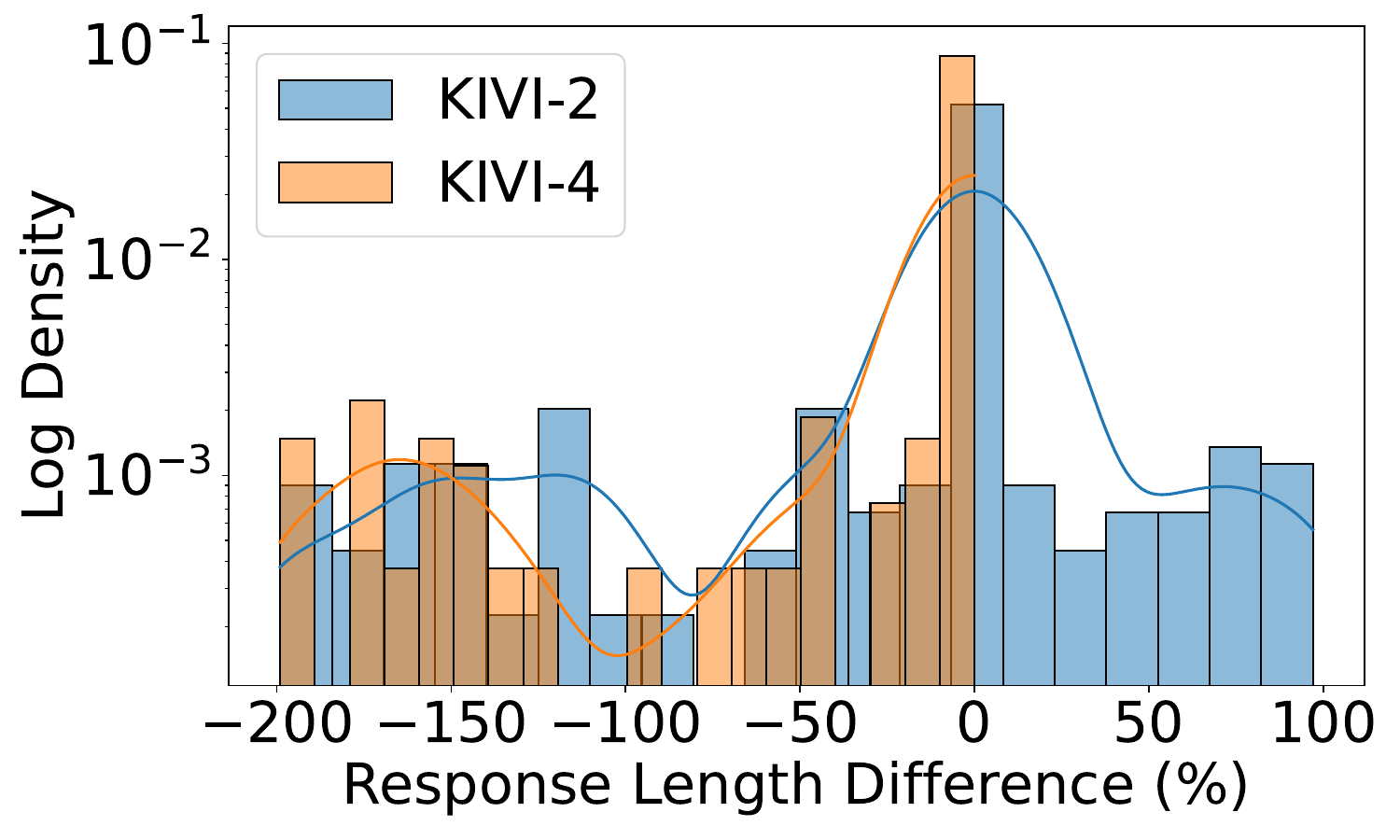}
        \label{fig:kivi-motivation_discrepancy-mistral}
        }
    \subfigure[GEAR]{
        \includegraphics[width=0.22\textwidth]{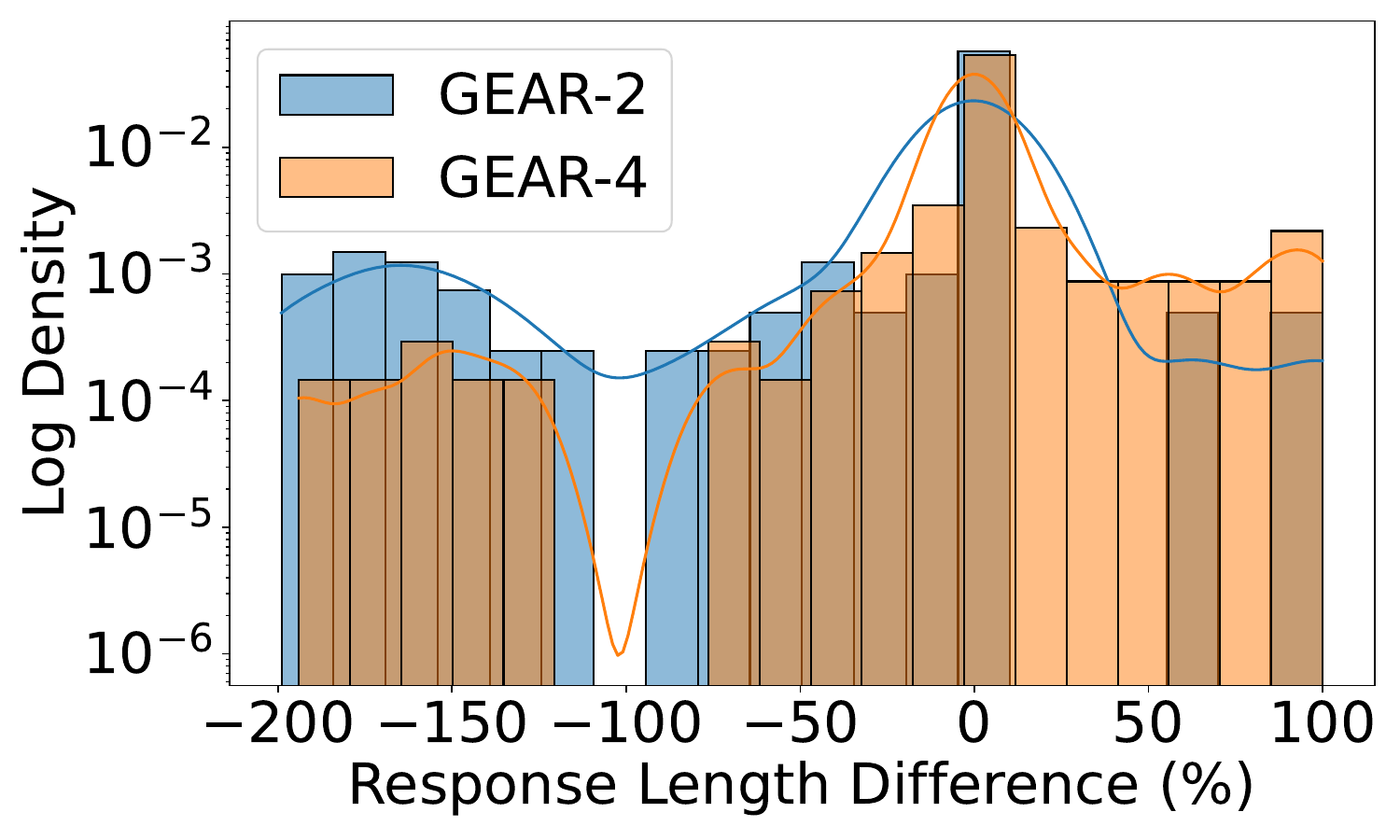}
        \label{fig:gear-motivation_discrepancy-mistral}
        }
    \subfigure[H2O]{\includegraphics[width=0.22\textwidth]{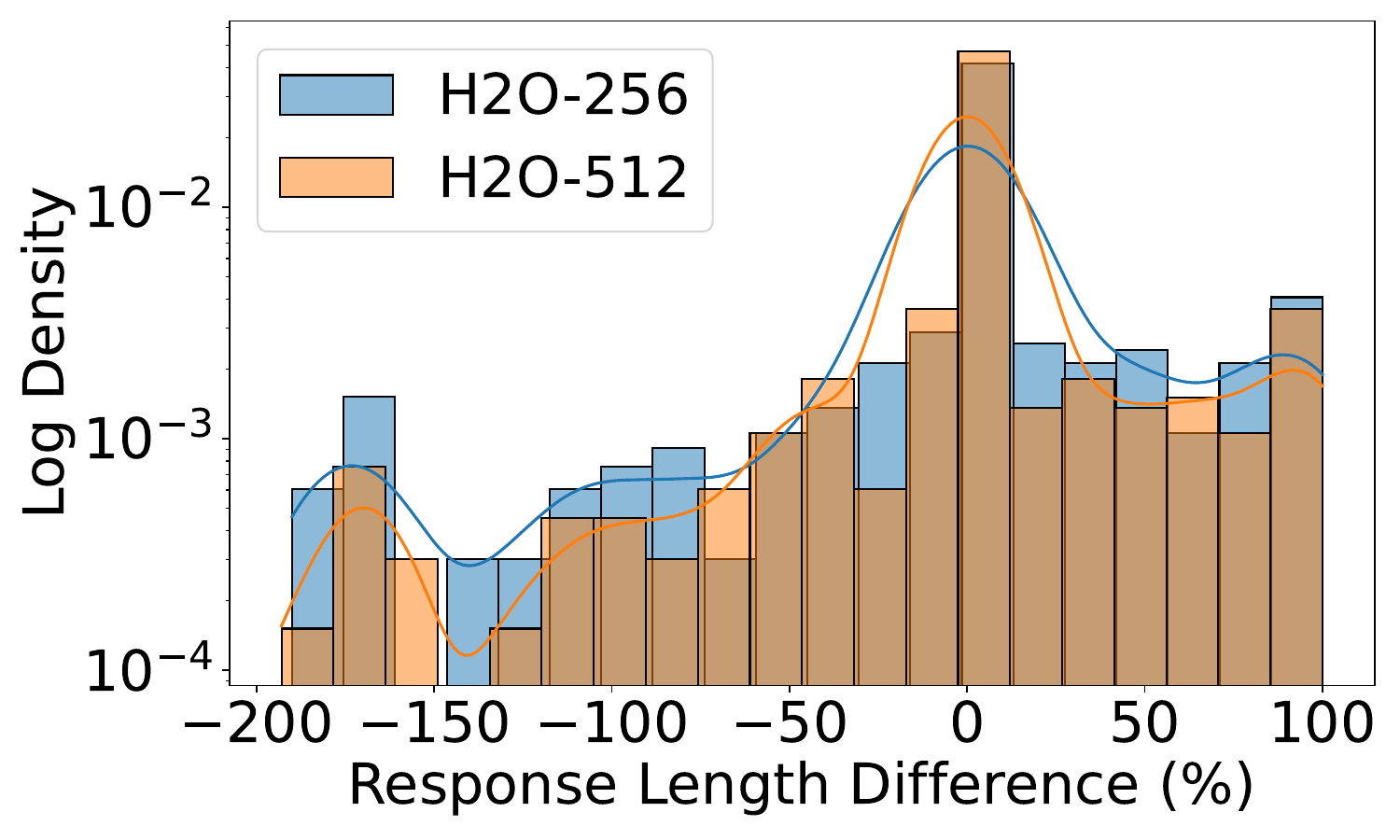}
        \label{fig:h2o-motivation_discrepancy-mistral}
    }
    \subfigure[StreamingLLM]{\includegraphics[width=0.22\textwidth]{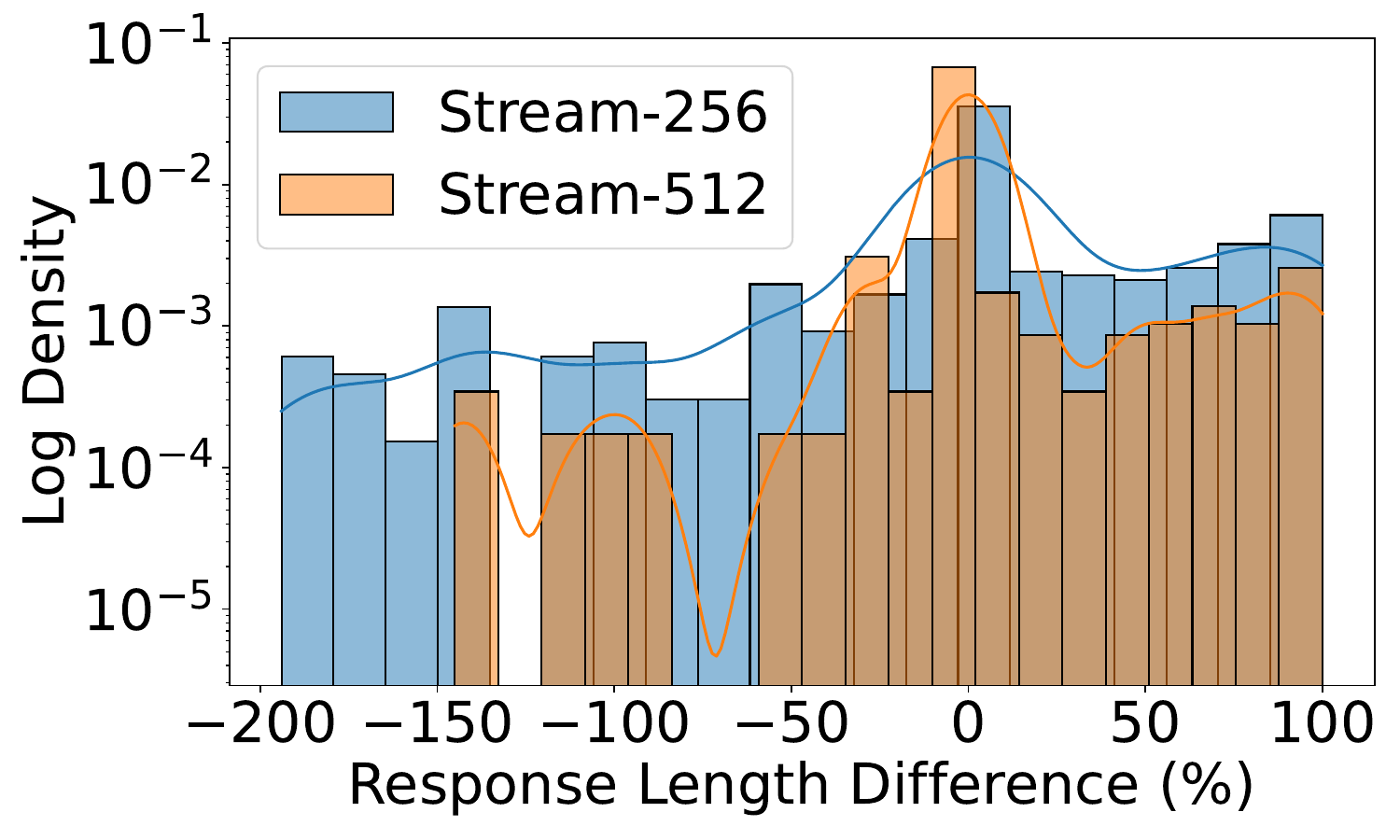}
        \label{fig:stream-motivation_discrepancy-mistral}
    }
    \vspace{-10pt}
    \caption{The Mistral-7B's distribution of response length difference across different compression algorithms.} 
    \label{fig:length-difference-distribution-mistral}
    \vspace{-10pt}
\end{figure*}


\begin{figure}[htbp]
    \centering
    \includegraphics[width=0.44\textwidth]{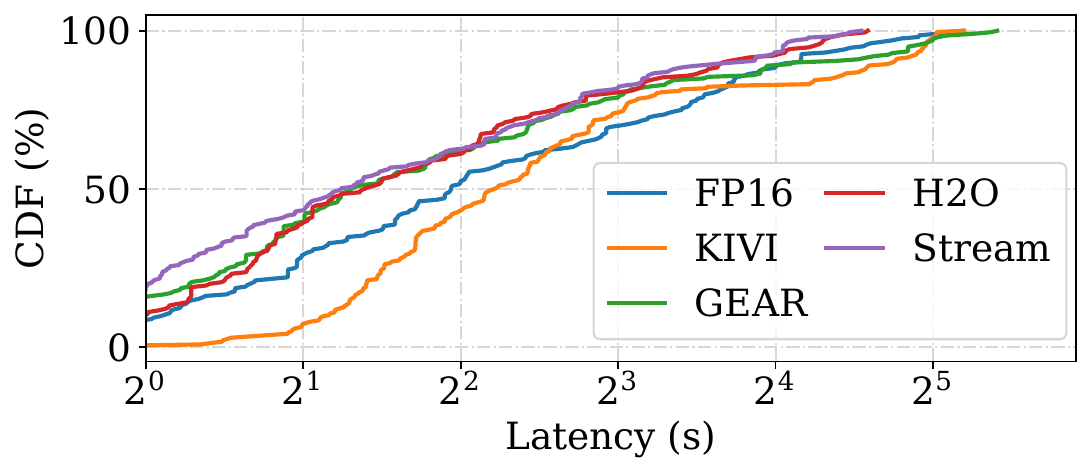}
    \vspace{-10pt}
    \caption{The Mistral-7B's CDF of the end-to-end latency (seconds) of various algorithms.} 
    \label{fig:length-latency-difference-distribution-mistral}
    \vspace{-10pt}
\end{figure}

\section{More Results of Length Analysis}
\label{app:experiment-response-length-analysis}
First, we outline the configurations of the compression algorithms in Section~\ref{subsec:length-distribution-analysis}. For text generation, we fix the temperature as $1$ for the FP16 baseline and compression methods. We also vary $T$ to assess the impact of length differences from the hyperparameter $T$ in Table~\ref{tab:response-length-variation}. For quantization-based methods, we only vary the quantization bits for KIVI and GEAR. For sparsity-based methods, we only vary the \texttt{KV} \texttt{cache} length for StreamingLLM and H2O. All other compression-related configurations remain consistent with those detailed in Appendix~\ref{app:eval-alg-details}.


Second, we supply more experimental results on Mistral-7B to demonstrate the generality of our statement in \textbf{Observations 3 and 4}, respectively. Particularly, we perform a similar experimental analysis as Table~\ref{tab:response-length-variation} on Mistral-7B and show the ratio (\%) of samples experiencing response length variations induced by temperature and \texttt{KV} \texttt{cache} compression in Table~\ref{tab:response-length-variation-mistral}. Similar to the LLaMA model, \texttt{KV} \texttt{cache} compression shows a clear tendency to produce verbose responses in Mistral-7B. We also repeat the experiments in Figure~\ref{fig:length-difference-distribution} and show the results of Mistral-7B in Figure~\ref{fig:length-difference-distribution-mistral}. The impact of the compression ratio on the response length remains consistent between the LLaMA and Mistral. We also measure the end-to-end latency for various compression algorithms on Mistral-7B, as shown in Figure~\ref{fig:length-latency-difference-distribution-mistral}. Our observation is that the latency benefits of \texttt{KV cache} compression are not prominent, and the verbose response length should be accounted for performance measurement of various \texttt{KV cache} compression.

Third,


\begin{table}[!t]\centering
\caption{The results of length analysis similar to Table~\ref{tab:response-length-variation}, but measured on Mistral-7B.}\label{tab:response-length-variation-mistral}

\begin{adjustbox}{width=0.48\textwidth} 
\begin{tabular}{lrrrrrrr}\toprule
\textbf{Metric} & \textbf{T=0.9} & \textbf{T=1.1} & \textbf{KIVI} &\textbf{GEAR} &\textbf{H2O} &\textbf{Stream} \\\midrule
\% of samples $D$ of which $ \geq 50\% $ & 45.1\% & 45.9 \&  &2.8\% & 0.8\% &11.0\% & 17.3\% \\ 
\% of samples $D$ of which $ \leq -50\% $ & 17.7\% & 20.0 \% &44.9\% &49.4\% &14.3\% & 16.3\% \\
\bottomrule
\end{tabular}
\end{adjustbox}
\end{table}

\begin{figure}[htbp]
    \centering
    \subfigure[Quantization]{
        \includegraphics[width=0.22\textwidth]{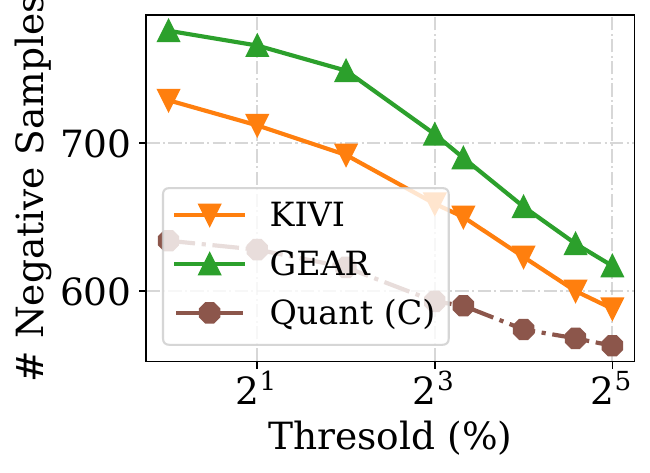}
        \label{fig:quant-failure-mistral}
        }
    \subfigure[Sparsity]{
        \includegraphics[width=0.22\textwidth]{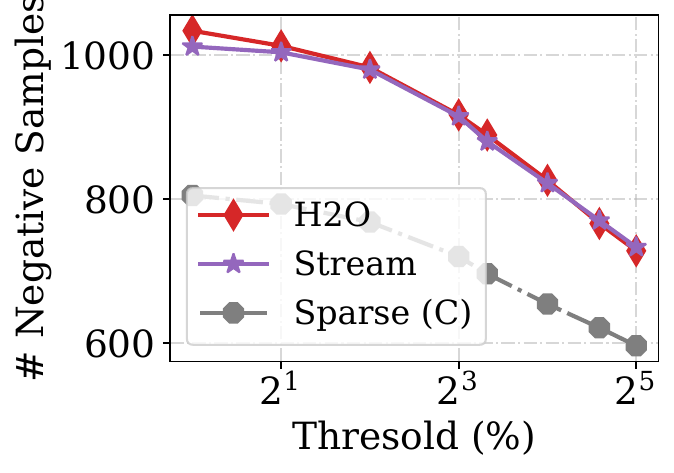}
        \label{fig:sparse-failure-mistral}
    }
    \vspace{-10pt}
    \caption{The threshold ($x$-axis) versus the number of negative samples ($y$-axis) for quantization-based (a) and sparsity-based (b) methods. The experimental results are measured on Mistral-7B.}
    \label{fig:compression-negative-analysis-mistral}
\end{figure}


\begin{figure}[htbp]
    \centering
        \includegraphics[width=0.48\textwidth]{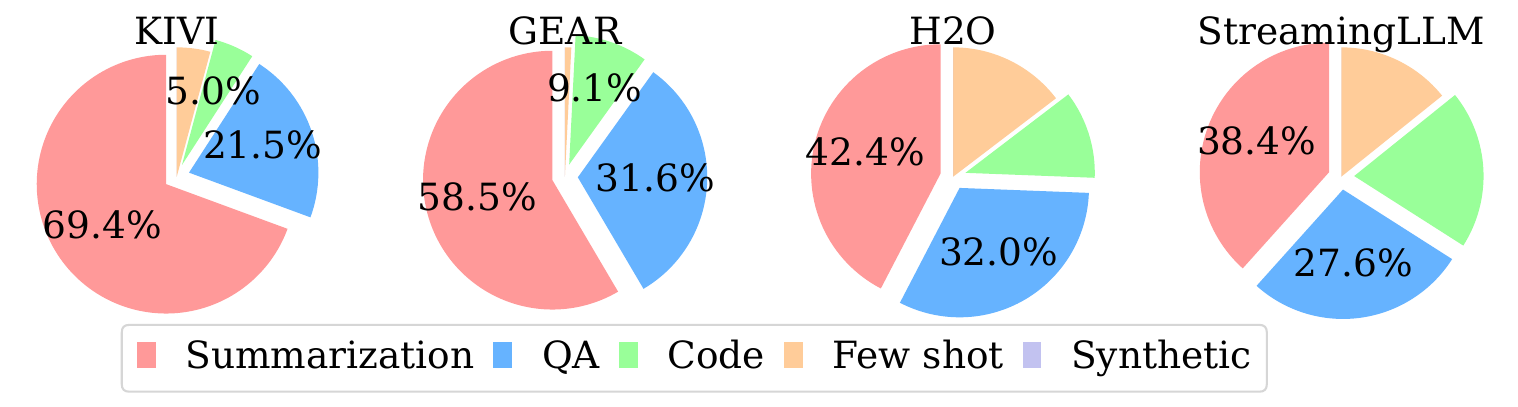}
    \vspace{-15pt}
    \caption{The pie chart details the proportion of negative samples over task types across varying compression algorithms on Mistral.}
    \label{fig:compression-task-type-mistral}
\end{figure}

\section{More Results of Negative Sample Analysis}
\label{app:experiment-neg-sample-analysis}

First, the detailed task description of the LongBench used in the negative sample analysis can be found in \url{https://huggingface.co/datasets/THUDM/LongBench#task-description}. It contains the detailed task description of the LongBench dataset, including the task name, task type, evaluation metric, and average length. We use the corresponding task type to collect the number of negative samples.

Section~\ref{subsec:survey-benchmarks} suggests that compression algorithms excel in proceeding short prompt lengths with no accuracy loss. Thus, we use LongBench and Llama-3.1-8B-instruct to conduct negative sample analysis. We assess the average scores of LLaMA-3.1-8B-instruct for baseline, KIVI, GEAR, H2O, and StreamingLLM on the LongBench test dataset are 41.2, 41.3, 40.9, 39.1, and 38.9, respectively. We also cover more experimental results about negative sample analysis on Mistral-7B in Section~\ref{subsec:survey-benchmarks}. In Mistral-7B, the average scores for baseline, KIVI, GEAR, H2O, and StreamingLLM on the LongBench test dataset are 33.3, 33.4, 33.4, 31.8, and 30.4, respectively. First, we vary the threshold in Algorithm~\ref{alg:failure-case-collection} to uncover the relationship between the threshold and the number of negative samples on Mistral-7B, as shown in Figure~\ref{fig:length-difference-distribution-mistral}. We conclude that the minor accuracy loss from compression algorithms does not indicate that each sample experiences a minor performance loss. It is not easy to eliminate the existence of negative samples. Second,  we present the sensitivity of task types to \texttt{KV} \texttt{cache} compression in Figure~\ref{fig:compression-task-type-mistral} on Mistral. Similar to LLaMA, performing \texttt{KV} \texttt{cache} compression on Mistral-7B considerably affects the accuracy performance on summarization and QA tasks. Overall, the experimental results further reinforce our statement in Section~\ref{subsec:failure-case-analysis}.



\begin{table}[!htp]\centering
\vspace{-10pt}
\caption{The accuracy of the length predictor for Mistral-7B.}\label{tab:throughput-accuracy-mistral}
\scriptsize
\begin{tabular}{lrrrrrr}\toprule
\textbf{Tools} & \textbf{FP16} & \textbf{KIVI} &\textbf{GEAR} &\textbf{H2O} &\textbf{Stream} \\\midrule
Length Predictor & 92.6\% & 92.3\% &88.8\% &92.8\% &89.5\% \\
\bottomrule
\end{tabular}
\vspace{-10pt}
\end{table}

\section{Throughput Predictor}
\label{app:thr-predictor}
We use Vidur's released code\footnote{\url{https://github.com/microsoft/vidur}} to realize the throughput predictor. The runtime time information of each operator in LLMs is profiled on A6000 NVIDIA RTX. The key difference between different compression algorithms and the FP16 baseline hinges upon the attention operation. Hence, apart from attention operators, all other operators are reused among different \texttt{KV cache} compression algorithms. We enumerate various combinations of batch sizes, sequence lengths, and stages to attain ample profiled runtime speed information for LLaMA-7B and Mistral-7B. Vidur provides the implementation code to construct and optimize the throughput predictor. We define the accuracy as $(1 - \frac{|T^{\text{pred}} - T^{\text{gt}}|}{T^{\text{gt}}}) \times 100\%$. 

\section{Length Predictor}
\label{app:length-ratio-predictor}
We collect the response length information from ShareGPT to synthesize the response length dataset. To account for the long-context prompt, we choose LongFormer with a maximum sequence size of 4096. We set the input of the length predictor as the input response and the target of the length predictor as the ratio between the response length and the prompt length. We define the accuracy as $(1 - \frac{|L^{\text{pred}} - L^{\text{gt}} |}{L^{\text{gt}}}) \times 100\%$. Table~\ref{tab:throughput-accuracy} has reported the prediction results on LLaMA3-8B. We include the prediction results on Mistral-7B in the second row of Table~\ref{tab:throughput-accuracy-mistral}. Overall, the bert-based length predictor can deliver accurate response length prediction for LLaMA and Mistral models.

\begin{table}[!htp]\centering
\vspace{-10pt}
\caption{The measured score of various algorithms evaluated on the negative sample benchmark dataset and Mistral-7B using LongBench's provided metric.}\label{tab:benchmark-accuracy-mistral}

\scriptsize
\begin{tabular}{lccccc}\toprule
\textbf{Task Type} & \textbf{Baseline} & \textbf{KIVI} &\textbf{GEAR} &\textbf{H2O} &\textbf{Stream} \\\midrule
\textbf{Summarization}  & 27.2 & 15.2 & 16.6 & 15 & 11.1 \\ 
\textbf{Question Answering} & 26.8 & 17.6 & 16.4 & 18.0 & 15.4 \\ 
\textbf{Code}           & 90.8 & 47.5 & 47.5 & 64.7 & 59.6 \\ 
\bottomrule
\end{tabular}
\vspace{-10pt}
\end{table}

\section{Performance on Negative Benchmark}
\label{app:perf-on-negative-bench}
We use the Mistral-7B and report the corresponding measured score on the negative sample benchmark dataset in Table~\ref{tab:benchmark-accuracy-mistral}.

\end{document}